\documentclass[10pt,journal,compsoc]{IEEEtran}
\usepackage{ifpdf}

\ifCLASSOPTIONcompsoc
  \usepackage[nocompress]{cite}
\else
  \usepackage{cite}
\fi
\ifCLASSINFOpdf
\else
\fi
\usepackage{amsmath}
\usepackage{algorithmic}

\usepackage{array}

\usepackage{stfloats}
\usepackage{url}

\usepackage{bbding}

\usepackage{algorithm}

\makeatletter
\let\NAT@parse\undefined
\makeatother
\usepackage{overpic}
\usepackage{amsfonts,amssymb}
\usepackage{subfig}
\usepackage{booktabs}
\usepackage{makecell}

\usepackage{graphicx}
\usepackage{multirow}
\usepackage{xurl}
\usepackage{wrapfig,lipsum}
\usepackage{float}
\usepackage[numbers]{natbib}
\usepackage{xcolor}
\usepackage{pifont}
\usepackage{verbatim}

\usepackage{hyperref}

\begin{document}
%
\title{A Survey on Deep Neural Network Pruning: Taxonomy, Comparison, Analysis, and Recommendations}
%
%
%
%

\author{Hongrong~Cheng*,~Miao Zhang*$\dagger$,~\IEEEmembership{Member,~IEEE},
~Javen~Qinfeng~Shi,~\IEEEmembership{Member,~IEEE}
\IEEEcompsocitemizethanks{
\IEEEcompsocthanksitem * Equal contribution. $\dagger$ Corresponding author.\\
H. Cheng and J. Q. Shi are with the University of Adelaide. M. Zhang is with Harbin Institute of Technology (Shenzhen). \\
E-mail:\{hongrong.cheng, javen.shi\}@adelaide.edu.au,\\
\{zhangmiao\}@hit.edu.cn.} \\
\thanks{Manuscript received May 01, 2023; revised June 11, 2024.}
}

%
%

\markboth{Journal of \LaTeX\ Class Files,~Vol.~14, No.~8, August~2024}
{Shell \MakeLowercase{\textit{et al.}}: Bare Demo of IEEEtran.cls for Computer Society Journals}
%



\IEEEtitleabstractindextext{%
\begin{abstract}
Modern deep neural networks, particularly recent large language models, come with massive model sizes that require significant computational and storage resources. To enable the deployment of modern models on resource-constrained environments and to accelerate inference time, researchers have increasingly explored pruning techniques as a popular research direction in neural network compression. More than three thousand pruning papers have been published from 2020 to 2024. However, there is a dearth of up-to-date comprehensive review papers on pruning. To address this issue, in this survey, we provide a comprehensive review of existing research works on deep neural network pruning in a taxonomy of 1) universal/specific speedup, 2) when to prune, 3) how to prune, and 4) fusion of pruning and other compression techniques. We then provide a thorough comparative analysis of eight pairs of contrast settings for pruning (e.g., unstructured/structured, one-shot/iterative, data-free/data-driven, initialized/pre-trained weights, etc.) and explore several emerging topics, including pruning for large language models, vision transformers, diffusion models, and large multimodal models, post-training pruning, and different levels of supervision for pruning to shed light on the commonalities and differences of existing methods and lay the foundation for further method development. Finally, we provide some valuable recommendations on selecting pruning methods and prospect several promising research directions for neural network pruning. To facilitate future research on deep neural network pruning, we summarize broad pruning applications (e.g., adversarial robustness, natural language understanding, etc.) and build a curated collection of datasets, networks, and evaluations on different applications. We maintain a repository on \url{https://github.com/hrcheng1066/awesome-pruning} that serves as a comprehensive resource for neural network pruning papers and corresponding open-source codes. We will keep updating this repository to include the latest advancements in the field. 
\end{abstract}
     
\begin{IEEEkeywords}
deep neural network pruning, model compression, model acceleration, large language models, vision transformers, large multimodal models, diffusion models, edge devices.
\end{IEEEkeywords}}

\maketitle

\IEEEdisplaynontitleabstractindextext

%
\IEEEpeerreviewmaketitle

\IEEEraisesectionheading{\section{Introduction}\label{sec:introduction}}

%
%
%
%

\label{introduction}
Over the past several years, Deep Neural Networks (\textbf{DNNs}) have achieved conspicuous progress in various domains and applications, such as Computer Vision (\textbf{CV}) \cite{simonyan2015very,dosovitskiy2021image}, Natural Language Processing (\textbf{NLP}) \cite{devlin2019bert,chowdhery2022palm}, Audio Signal Processing (\textbf{ASP}) \cite{bohnstingl2021towards,latif2023sparks}, and cross-modal applications \cite{lin2024vila,liu2024sora}. Although DNNs achieve remarkable success in various areas, their performance relies heavily on model parameters and computational cost. For example, the widely used ResNet-50 \cite{he2016deep} takes over 95 MB of memory, containing over 23 million parameters \cite{you2019drawing}. $\textrm{BERT}_{\textrm{BASE}}$ \cite{devlin2019bert} is around 440 MB with 110 million parameters, GPT-3 includes up to 175 billion parameters \cite{wu2023ai}, and GPT-4 has even more. The trend of enlarging neural network size is anticipated to persist. 

However, the more parameters of DNNs, the more time and memory space they typically require for processing the inputs \cite{han2015learning}. The high training and inference costs associated with these models present a significant challenge to their deployment on devices constrained by limited computational resources (such as CPU, GPU, and memory), energy, and bandwidth \cite{han2015deep, dong2017more, you2019gate}. For example, real-life applications such as autonomous driving, field rescue, and bushfire prevention necessitate high accuracy and efficient resource usage, including fast real-time response and compact memory footprint. Deep neural networks' computational complexity and memory footprint can make them impractical for deployment on edge devices \cite{luo2017thinet}. With the popularity of Large Language Models (\textbf{LLMs}) in recent years, there is growing interest in compressing neural networks for computers with flexible hardware requirements \cite{frantar2023sparsegpt}. In addition, deep neural networks that contain redundant features can undermine their robustness, elevating the risk of adversarial attacks \cite{sehwag2020hydra}. For instance, high-dimensional feature spaces created by these networks can provide more entry points for adversarial attacks, undermining the network's ability to generalize beyond its original training data.  

\begin{figure}[t]
\centering
\begin{minipage}{8cm}
  \includegraphics[width=8cm,height=4cm]{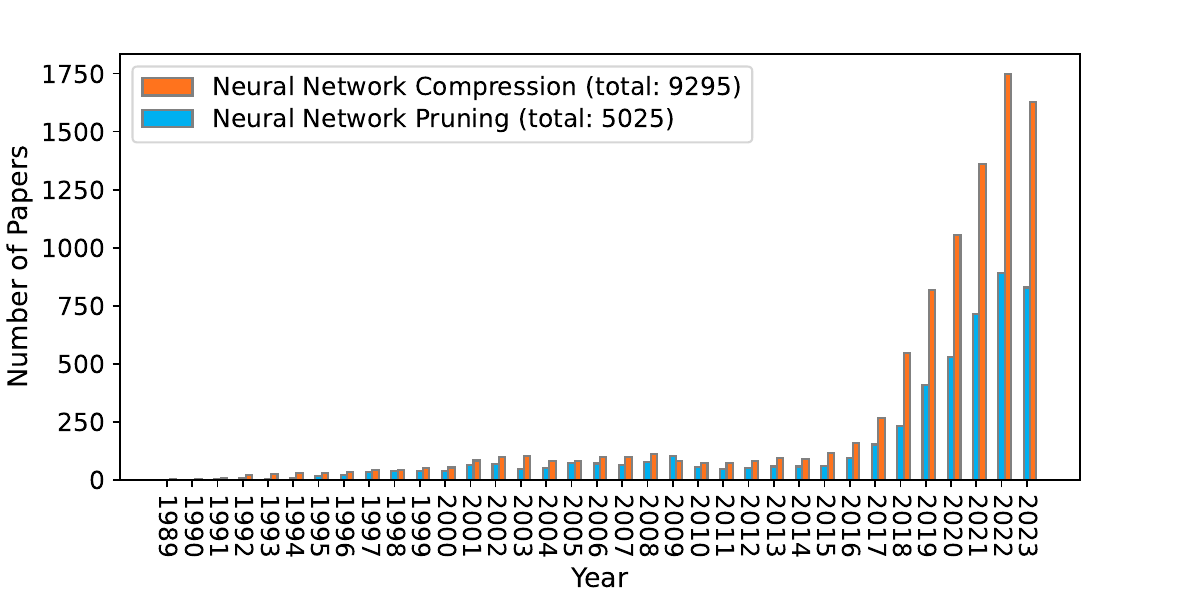}
\end{minipage}
\vspace{-0.3cm}
\caption{The number of papers on neural network pruning and compression from 1989-2023.\protect\footnotemark}
\label{Fig:pruning-papers-statistics}
\vspace{-0.5cm}
\end{figure}

To relieve this issue, researchers have proposed various neural network compression techniques to design lightweight models, including neural network pruning (\cite{ashkboos2024slicegpt, ma2023llmpruner}), low-rank factorizations of the weight matrices (\cite{denton2014exploiting,lin2018holistic}), quantization (\cite{dettmers2023qlora,shao2024omniquant}), knowledge distillation (\cite{gu2024minillm,xu2024survey}), neural architecture search (\cite{liu2019darts,zhang2021idarts}), and other techniques (\cite{ding2021repvgg,chu2024mobilevlm}). Among them, there is continuing interest in neural network pruning, which has been proven as a desirable and effective way to save memory space and computation time at inference while maintaining a comparable or even better performance compared to the original DNNs. As shown in Fig.~\ref{Fig:pruning-papers-statistics}, 
the number of papers on pruning has been markedly increasing since 2015. It presents more than half of the papers on neural network compression.

Research on pruning can be traced back to literature as early as 1988 \cite{hanson1988comparing}. However, it was only until the emergence of \cite{han2015deep} that the research community realized the potential of pruning in removing significant redundancy in deep neural networks, and pruning began to gain widespread attention. Several pieces of literature review prior work on deep neural network pruning, as shown in Table~\ref{Table-survey-works-this-paper}. Although these works overview several aspects of pruning and provide helpful guidance for researchers, many of them (e.g., \cite{mishra2020survey,ghimire2022survey,park2024comprehensive,xu2024a}) focus on multiple compression techniques, such as pruning, quantization, and knowledge distillation, with only brief examination of each technique. For example, \citet{mishra2020survey} summarize compression techniques, including pruning, quantization, low-rank factorization, and knowledge distillation, where pruning is primarily introduced from channel/filter pruning, and many essential pruning techniques (such as lottery ticket hypothesis) are not included. Some reviews concentrate on one specific aspect. For example, \citet{wang2022recent} provide only an overview of pruning at initialization and do not include studies on pruning during training, pruning after training, etc. \citet{he2023structured} focuses only on structured pruning and does not discuss the other two common types of pruning: unstructured and semi-structured pruning. \citet{ghimire2022survey}) focus on reviewing Convolutional Neural Network (\textbf{CNN}) pruning and lack descriptions of pruning for other deep neural networks, such as Recurrent Neural Networks (\textbf{RNNs}) \cite{fang2023depgraph}, Transformer based models \cite{frantar2023sparsegpt}, and diffusion models \cite{fang2023structural}. Some recent review works (such as \cite{zhu2023survey,xu2024a}) survey compression techniques for large models, encompassing a broad array of topics from pruning, quantization, knowledge distillation, etc. Among these, pruning is discussed concisely.

\footnotetext{The data is from https://www.webofknowledge.com.}

This survey aims to provide a comprehensive overview of deep neural network pruning for diverse readers. We review representative pruning methods, propose a new taxonomy, conduct a comprehensive analysis of how different pruning techniques perform in practice, and give practitioners who wish to utilize pruning recommendations on choosing a suitable pruning method for various requirements. Our contributions are as follows:

\textbf{(1) Comprehensive review.} To our best knowledge, this survey is the most comprehensive overview of modern deep neural network pruning techniques. It distills ideas from over 300 related academic papers, covering pruning methods for small to medium models to cutting-edge large models. In addition, we establish a new taxonomy, as shown in Fig.~\ref{Fig:pruning-taxonomy}, and provide detailed descriptions of the representative methods for each class of pruning methods.

\begin{table}[t]
  \caption{Representative surveys and descriptions.}
  \label{Table-survey-works-this-paper}
  \vspace{-0.1cm}
  \centering
    \begin{tabular}{l|c|c}
    \Xhline{0.3ex}
    Survey & Main work description &  Year\\
    \hline
    \cite{reed1993pruning} & survey of pruning methods & 1993 \\
    \cite{blalock2020what} & overview 81 papers and propose ShrinkBench, & 2020 \\
    & an open-source framework for evaluation & \\
   \cite{neill2020survey} & pruning+weight sharing+low-rank matrix+ & 2020\\
    & KD+quantization & \\
    \cite{liu2020pruning} & pruning criteria+procedure & 2020 \\
    \cite{choudhary2020comprehensive} & pruning+quantization+KD+low-rank factor. & 2020\\
    \cite{mishra2020survey} & pruning+KD+NAS+Tensor-Decompose & 2020 \\
    & +quantization+hardware acceleration & \\
    \cite{hoefler2021sparsity} & work on sparsity in deep learning up to 2020 & 2021\\
    \cite{wang2022recent} & overview pruning at initialization & 2022 \\
    \cite{ghimire2022survey} & pruning+KD+NAS+quantization & 2022 \\
    & +Tensor-Decompose+hardware accel. \\
    \cite{he2023structured} & structured pruning & 2023 \\
    \cite{zhu2023survey} & pruning+KD+quantization+others for LLM & 2023 \\
    \cite{wang2024model} & pruning+KD+quantization+others for LLM & 2024\\
    \cite{park2024comprehensive} & pruning+KD+quantization+others for LLM & 2024 \\
    \cite{xu2024a} & model compression+others for LLMs & 2024 \\
    \Xhline{0.3ex}
  \end{tabular}
  \vspace{-0.4cm}
\end{table}

\textbf{(2) Comparative experiments and analysis.} We conduct a comparative analysis of eight pairs of contrast settings for pruning and emerging advances, such as pruning for LLMs and different supervision levels for pruning. Unlike existing surveys, we conduct experiments and provide discussions. 

\textbf{(3) Collection of abundant resources.} We summarize miscellaneous pruning applications and provide benchmark datasets, networks, and evaluations for different applications. Our collected resources in Appendix D can guide researchers and practitioners in understanding, utilizing, and developing different network pruning methods for various requirements. Ongoing updates of representative pruning efforts are available at \url{https://github.com/hrcheng1066/awesome-pruning}.

\textbf{(4) Recommendations and future directions.} This survey provides valuable recommendations for choosing appropriate pruning methods for different application requirements and highlights promising future research directions. 

The remainder of this survey is organized as follows. In Section \ref{taxonomy}, we establish a clear taxonomy of pruning. Section \ref{granularity} - \ref{learntoprune} offer an overview of speedup, when to prune, and how to prune, followed by a comprehensive comparative analysis of different kinds of pruning methods in Section \ref{comparative}. Section \ref{fusion} discusses integrating pruning with other compression methods. Practical recommendations for choosing pruning methods and future directions are provided in Section \ref{recommendations}. We conclude this paper in Section \ref{conclusion}. 

\section{Taxonomy}
\label{taxonomy}
\begin{figure*}[t]
\centering
\begin{minipage}{16cm}
\includegraphics[width=16cm,height=6.5cm]{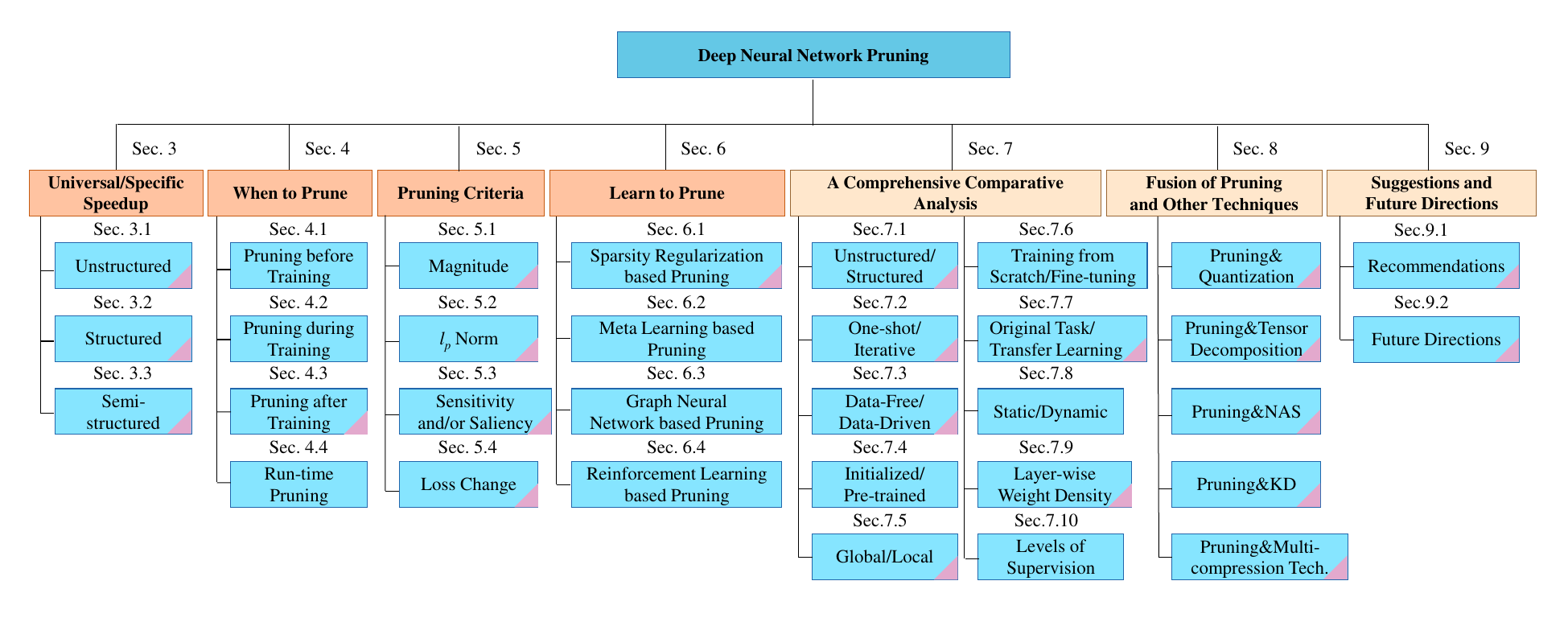}
\end{minipage}
\vspace{-0.4cm}
\caption{\small An overview of the hierarchical structure of the survey, with sections involving large models highlighted by a pink triangle.}
\label{Fig:pruning-taxonomy}
\vspace{-0.4cm}
\end{figure*}

There are three critical questions when pruning a deep neural network. \textbf{(1) Whether to achieve universal or specific acceleration through neural network pruning?} \textbf{Universal acceleration operates independently of specialized hardware/software, while specific acceleration relies on it.} \textbf{(2) When to prune the neural network? Specifically, is the neural network pruned before, during, or after training the network for static pruning or dynamic (i.e., run-time) pruning?} \textbf{(3) Whether to prune based on specific criteria or learn to prune?} The answers to the three questions correspond to the three primary aspects of deep neural network pruning, as depicted in the orange sections of Fig.~\ref{Fig:pruning-taxonomy}.

\textbf{The first question} is whether speedup depends on specific hardware/software. It is usually divided into three types: unstructured (\cite{frankle2019lottery,sun2024simple,tanaka2020pruning}), semi-structured (also called pattern-based) (\cite{frantar2023sparsegpt,meng2020pruning,ma2020image}) and structured (\cite{ma2023llmpruner,wang2023trainability,fang2023structural}). Only structured pruning can achieve universal neural network acceleration without requiring special hardware or software. Conversely, both unstructured and semi-structured pruning need the support of special hardware or software. Given that the primary objective of pruning is acceleration, the first question addresses 
the most fundamental and user-concerned aspect of this process.

\textbf{The second question} particularly concerns the arrangement between pruning weights and training weights of the neural network for static pruning. Based on whether pruning is performed before, during, or after training the network, static pruning can be divided into three categories: pruning before training (\textbf{PBT}) (\cite{lee2019snip,tanaka2020pruning,su2020sanity,wang2020picking}), pruning during training (\textbf{PDT}) (\cite{liu2017learning,wen2016learning,huang2018data}), and pruning after training (\textbf{PAT}) (\cite{frankle2019lottery,yang2023global,ma2023llmpruner}). In dynamic pruning, subnetworks are generated at run-time for each input data.

\textbf{The third question} considers whether to prune neural networks using specific criteria or by learning. Criteria rely on a scoring formula to measure the importance of each weight (or filter, neuron, and so on). Commonly used pruning criteria include magnitude, norm, loss change, etc. In addition, it is possible to prune neural networks by learning, such as through sparsity regularization training \cite{huang2018data} or dynamic sparse training \cite{liu2020dynamic}. Whether through criteria or learning, pruning aims to determine the weights of a network that should be pruned.

The above three aspects determine the main characteristics of a pruning algorithm. Different combinations of these three aspects form various pruning methods. We provide a new taxonomy of deep neural network pruning in Section~\ref{granularity} - \ref{learntoprune}, and Section~\ref{fusion}, as shown in Fig.~\ref{Fig:pruning-taxonomy}.

\section{Specific or Universal Speedup}
\label{granularity}
This section categorizes pruning into unstructured, semi-structured, and structured. The first two types correspond to specific speedup, while the third corresponds to universal speedup. 

\subsection{Unstructured Pruning}
Unstructured pruning, also called non-structured pruning or weight-wise pruning, is the finest-grained case.

\begin{figure*}[t]
\centering
  \subfloat[From the view of neurons and connections]{
  \begin{minipage}{6.5cm}
      \includegraphics[width=6.5cm,height=3.5cm]{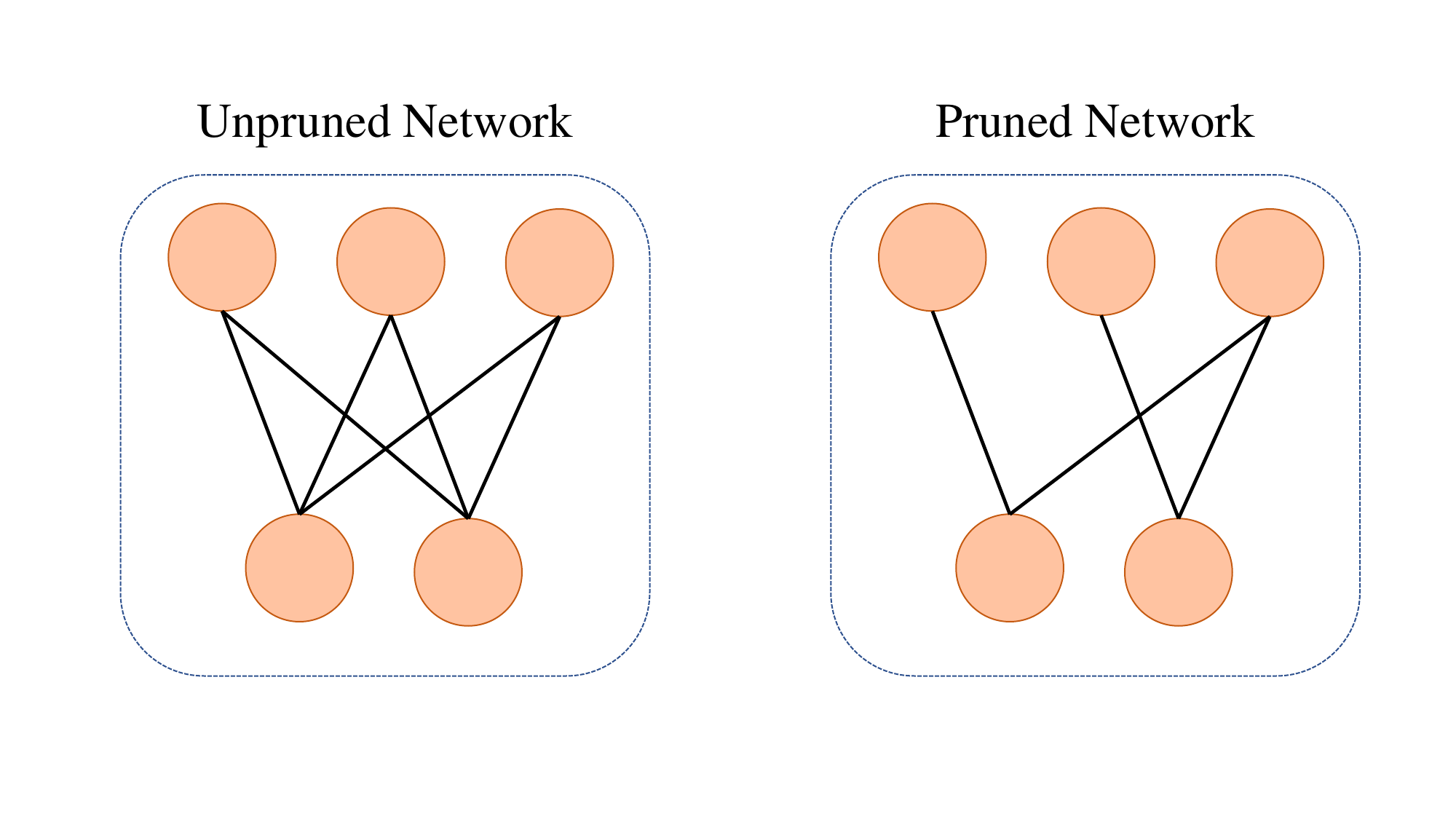}
      \vspace{-0.3cm}
        \label{Fig:visualization-unstructured-pruning-a}
  \end{minipage}
  }
  \subfloat[From the view of weights and masks]{  
  \begin{minipage}{6.5cm}
       \includegraphics[width=6.5cm,height=3.5cm]{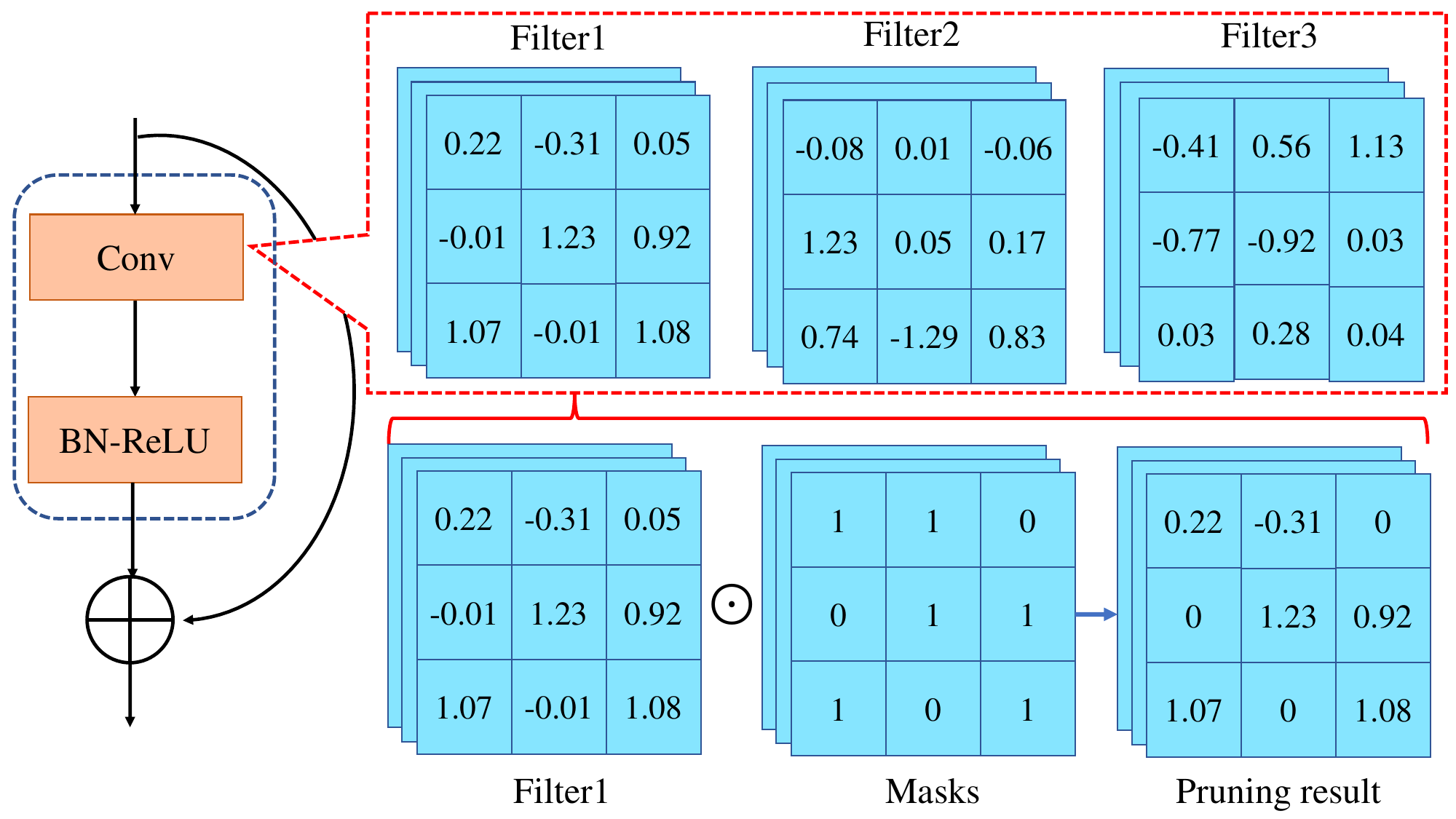}
       \vspace{-0.3cm}
     \label{Fig:visualization-unstructured-pruning-b}
  \end{minipage} 
  }
 \caption{The visualization of unstructured pruning. The light orange circles denote neurons. }
 \label{Fig:visualization-unstructured-pruning}
\vspace{-0.4cm}
\end{figure*}

\textbf{Definition 1} (Unstructured Pruning). Given neural network weights $W=\{w_0,w_1,...,w_K\}$, a dataset $\mathcal{D} = \{(\mathbf{x}_i, \mathbf{y}_i)\}_{i=1}^{N}$ composed of input ($\mathbf{x}_i$), output ($\mathbf{y}_i$) pairs, and a desired total number of non-zero weights $k$ (less than $K$), unstructured pruning can be written as the following constrained optimization problem \cite{lee2019snip}:
\begin{equation}
\begin{split}
\mathop{\textrm{min}}\limits_{W} \ & \mathcal{L}(W;\mathcal{D}) = \mathop{\textrm{min}}\limits_{W} \frac{1}{N}\sum_{i=1}^N\ell(W;(\mathbf{x}_i,\mathbf{y}_i)), \\
& \textrm{s.t.} \ \lVert W\rVert_0 \leq k.
\label{eq:unstructured-pruning-1}
\end{split}
\end{equation}
In practice, for small or medium models, unstructured pruning usually does not directly set the weights to 0 but sets their corresponding masks (or indicators) $M$ to 0 \cite{liu2021group, wang2020picking}. In this case, unstructured pruning is regarded as applying a binary mask to each weight. Consequently, Eq.(\ref{eq:unstructured-pruning-1}) is correspondingly changed as:
\begin{equation}
\begin{split}
\mathop{\textrm{min}}\limits_{\mathbf{w,m}} \ & \mathcal{L}(W\odot M;\mathcal{D}) = \mathop{\textrm{min}}\limits_{W,M} \frac{1}{N}\sum_{i=1}^N\ell(W \odot M;(\mathbf{x}_i.\mathbf{y}_i)), \\
& \textrm{s.t.} \ \lVert M\rVert_0 \leq k.
\label{eq:unstructured-pruning-2}
\end{split}
\end{equation}
Generally, the network is retrained (i.e., fine-tuning or training-from-scratch) with fixed masks $M$, and the masked-out weights are not involved in retraining. In the case of large models, such as LLMs, assigning a mask to each weight poses a challenge due to the immense number of weights. It is common to directly set the weights that need to be pruned to zero, as seen in \cite{sun2024simple,frantar2023sparsegpt}. Fig.~\ref{Fig:visualization-unstructured-pruning} is an example of weight-wise pruning by removing the weights directly
(as shown in Fig.~\ref{Fig:visualization-unstructured-pruning} (a)) or masking the weights with their corresponding masks (as shown in Fig.~\ref{Fig:visualization-unstructured-pruning} (b)), respectively. Since it can remove weights anywhere, the irregular replacement of non-zero weights leads to actual acceleration requiring the support of special software and/or hardware \cite{wen2016learning,li2017pruning,tang2021manifold,han2015deep}. Therefore, we classify unstructured pruning as a specific speedup technique. 

\subsection{Structured Pruning}
\textbf{Definition 2} (Structured Pruning). 
Given a specific prune ratio and a neural network with $S=\{s_1, s_2, ..., s_L\}$, where $s_i$ can be the set of channels, filters, neurons, or transformer attention heads in layer $i$. Structured pruning aims to search for $S^{'}=\{s_1^{'}, s_2^{'}, ..., s_L^{'}\}$ to minimize performance degeneration and maximize speed improvement under the given prune ratio, where $s_i^{'}\subseteq s_i$, $i\in \{1,.., L\}$. 

\begin{figure*}[t]
\centering
  \subfloat[Structured pruning for CNNs]{
  \begin{minipage}{5cm}
      \includegraphics[width=5cm,height=3.5cm]{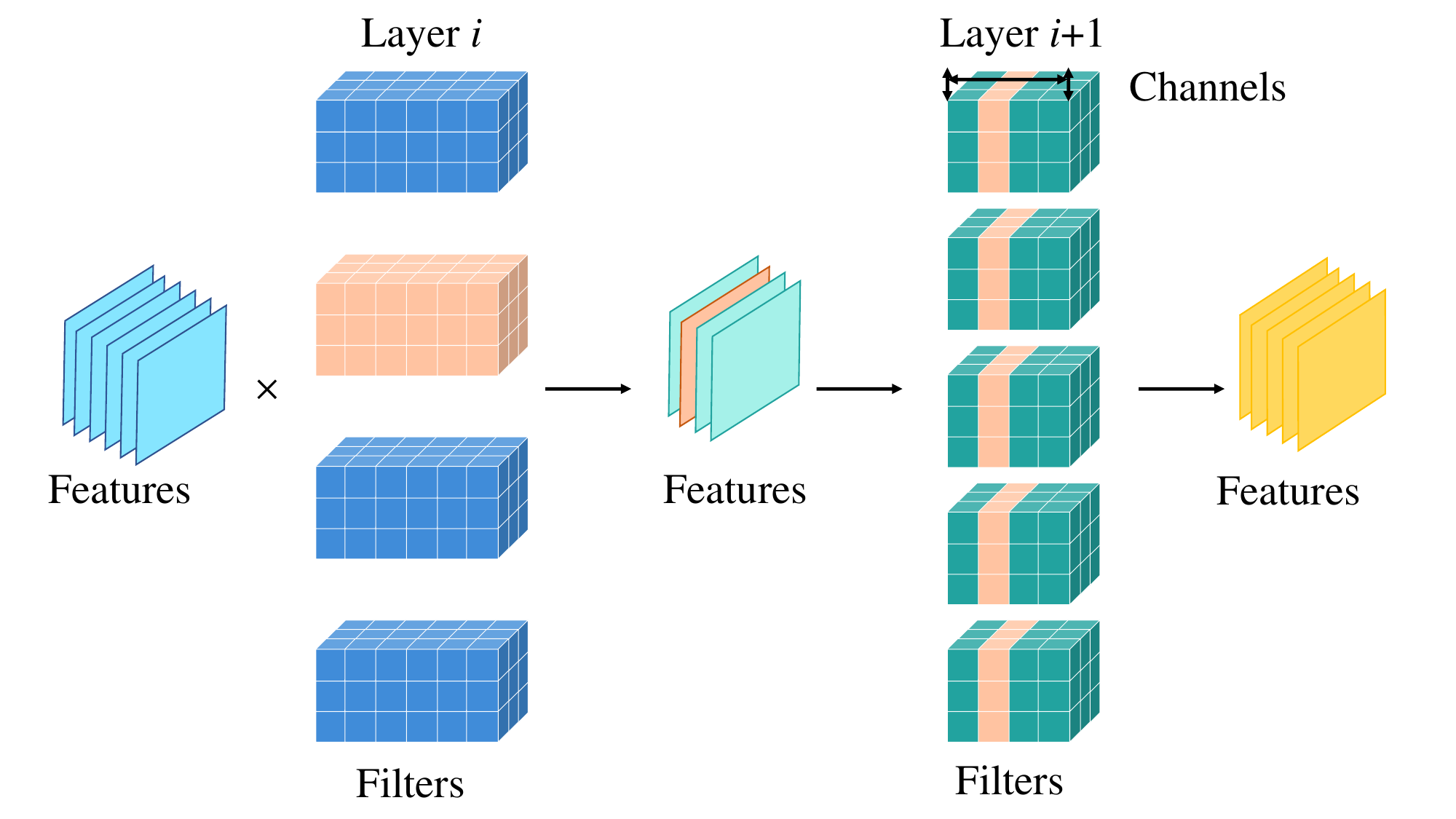}
      \vspace{-0.3cm}
        \label{Fig:filter-channel-pruning}
  \end{minipage}
  }
  \subfloat[Structured pruning for Transformers]{  
  \begin{minipage}{5cm}
       \includegraphics[width=5cm,height=3.5cm]{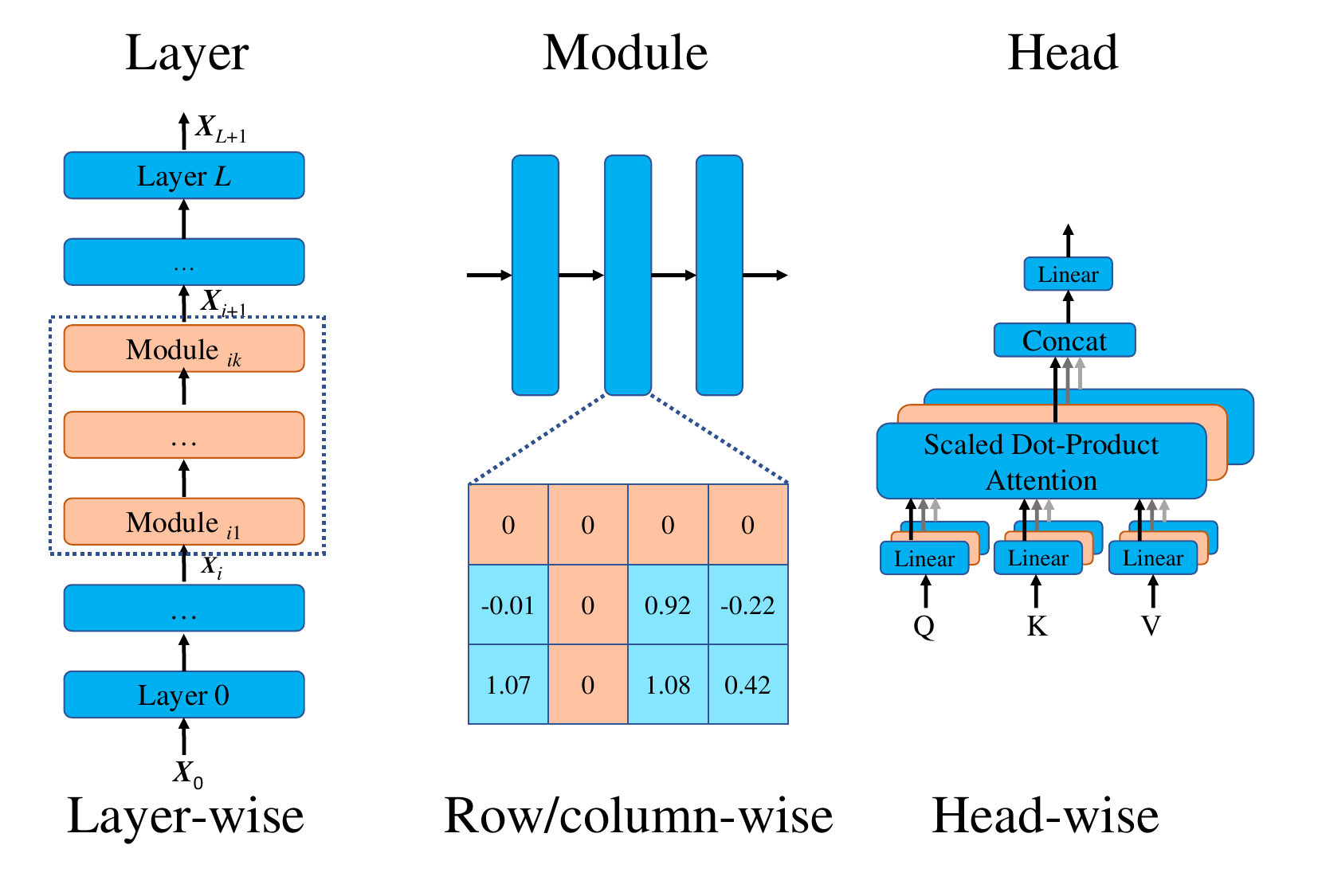}
       \vspace{-0.3cm}
     \label{Fig:structured-pruning}
  \end{minipage} 
  }
  \subfloat[Examples of semi-structured pruning]{  
  \begin{minipage}{5cm}
  \centering\includegraphics[width=5cm,height=3.5cm]{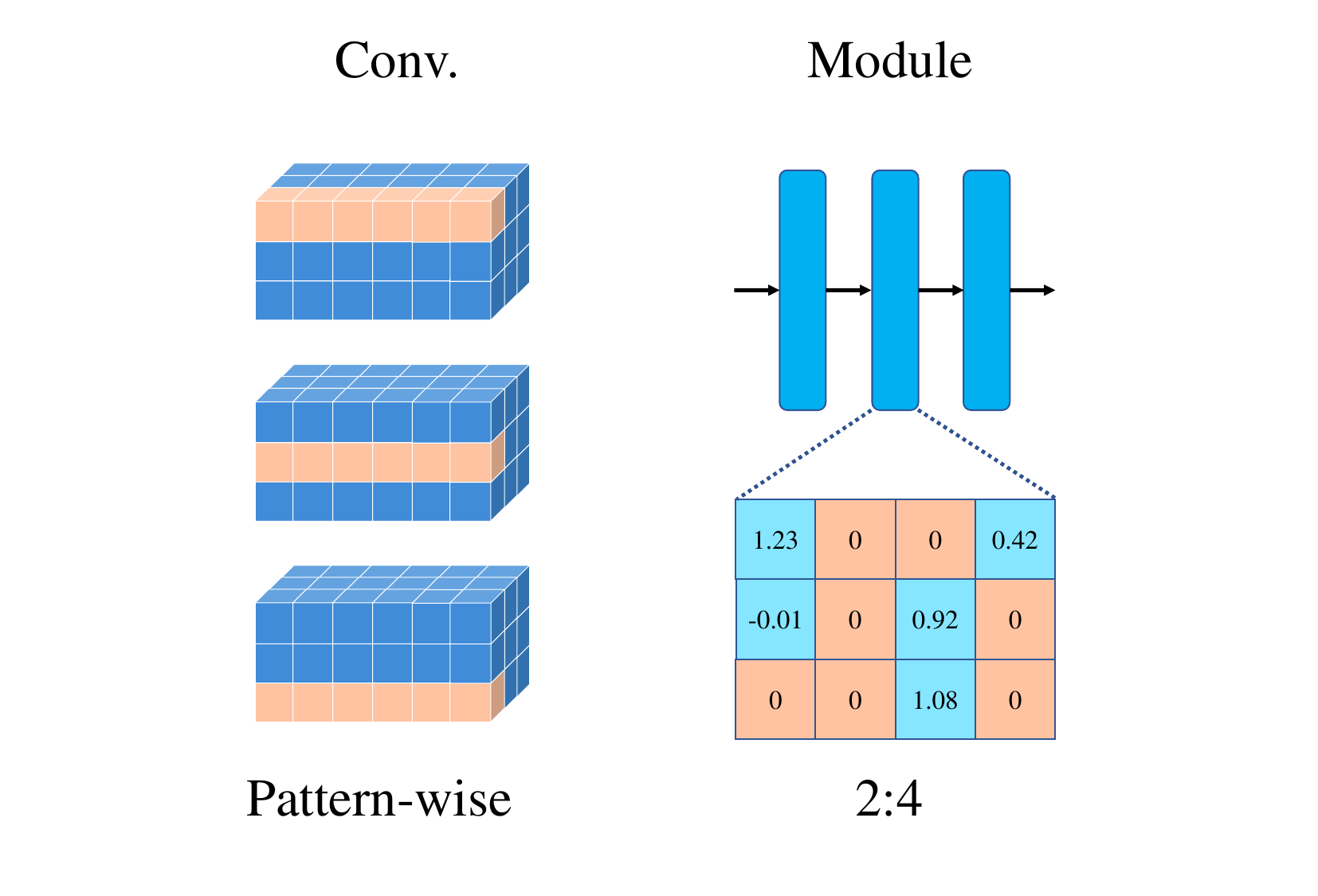}
  \vspace{-0.3cm}
     \label{Fig:semi-strucutred}
  \end{minipage} 
  }  
 \caption{The visualization of typical structured and semi-structured pruning: each Conv. consists of cubes representing weights. Orange cube sets/blocks indicate the pruned structures. Best view in color and zoom in.}
 \label{Fig:visualization-structured-pruning}
\vspace{-0.4cm}
\end{figure*}

Structured pruning removes entire filters (\cite{you2019gate}), channels (\cite{liu2021group,ashkboos2024slicegpt}), transformer attention heads (\cite{ma2023llmpruner}), or even layers (\cite{men2024shortgpt}), as shown in Fig.~\ref{Fig:visualization-structured-pruning} (a) and Fig.~\ref{Fig:visualization-structured-pruning} (b), and can rebuild a narrow model with a regular structure~\footnote{For Transformers \cite{vaswani2017attention}, deleting rows (columns) of weight matrices is similar to pruning filters (channels) for CNNs.}. It does not require the support of special hardware and software (such as sparse convolution libraries) and can directly speed up networks and reduce the size of the neural networks \cite{liu2021group,nonnenmacher2022sosp, luo2017thinet,li2017pruning}. 

\subsection{Semi-structured Pruning}
To improve the flexibility of structured pruning and achieve a lower accuracy drop when the pruning rate is high, some works (\cite{meng2020pruning,ma2020image}) introduce semi-structured pruning, also called pattern-based pruning in \cite{ma2020image}, to achieve high accuracy and structural regularity simultaneously. 
Some examples are shown in Fig.~\ref{Fig:visualization-structured-pruning} (c). For example, \citet{meng2020pruning} treat a filter as several stripes and propose to prune stripes in each filter. SparseGPT~\cite{frantar2023sparsegpt} introduces a 2:4 or 4:8 sparsity pattern to reduce a LLM's parameters by half. In this configuration, at least two zeros are mandatory in every set of four consecutive values. The 2:4 pattern can utilize NVIDIA Ampere GPU’s sparse tensor cores to accelerate matrix multiplication~\cite{yang2023global}. In contrast, structured pruning is classified as coarse-grained structured pruning (\cite{ma2020image,li2021npas}), while semi-structured pruning is classified as fine-grained.

\section{When to Prune}
\label{pipelines}
\begin{figure*}[t]
\centering
  \subfloat[SNIP \cite{lee2019snip} as an example of PBT]{
  \begin{minipage}{5cm}
      \includegraphics[width=5cm,height=3.5cm]{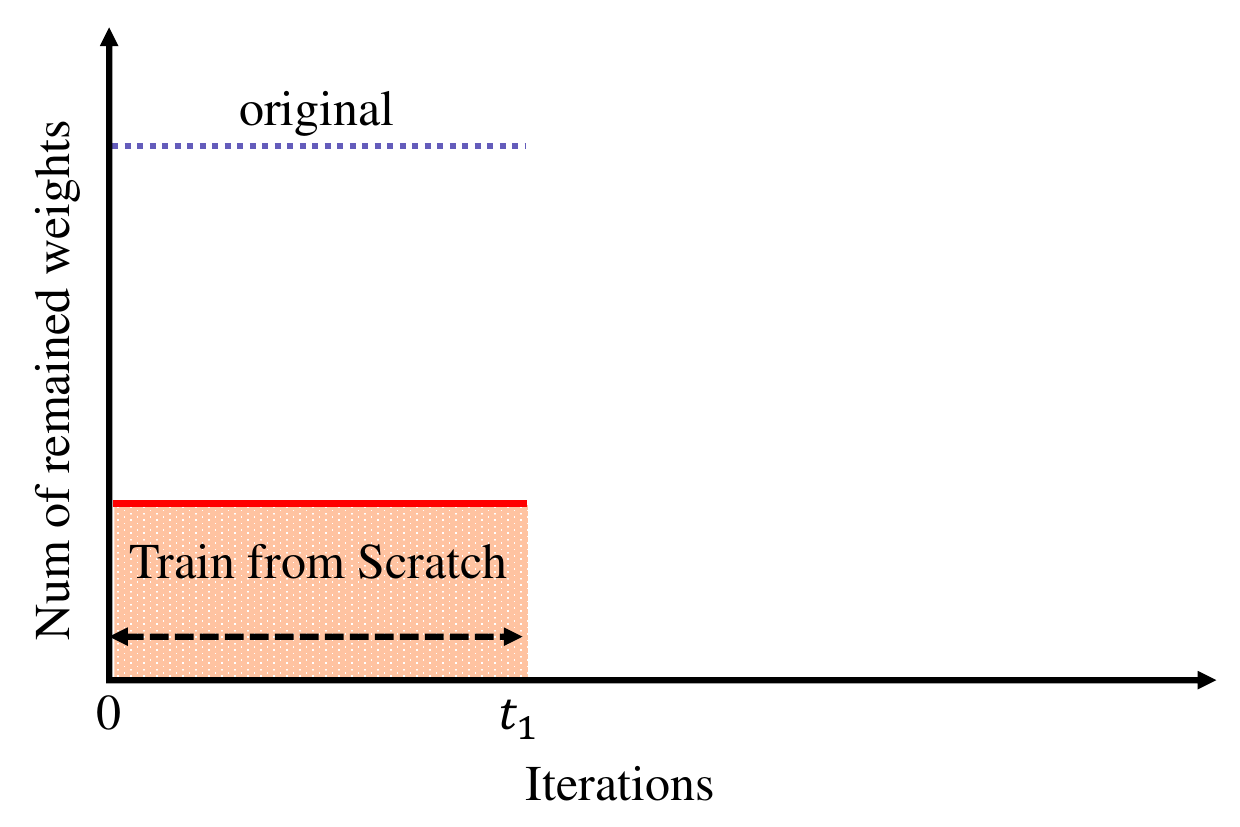}
      \vspace{-0.2cm} 
        \label{Fig:typical-sparsity-changes-a}
        \vspace{-0.2cm} 
  \end{minipage}
  }
  \subfloat[FCF \cite{li2019compressing} as an example of PDT]{  
  \begin{minipage}{5cm}
       \includegraphics[width=5cm,height=3.5cm]{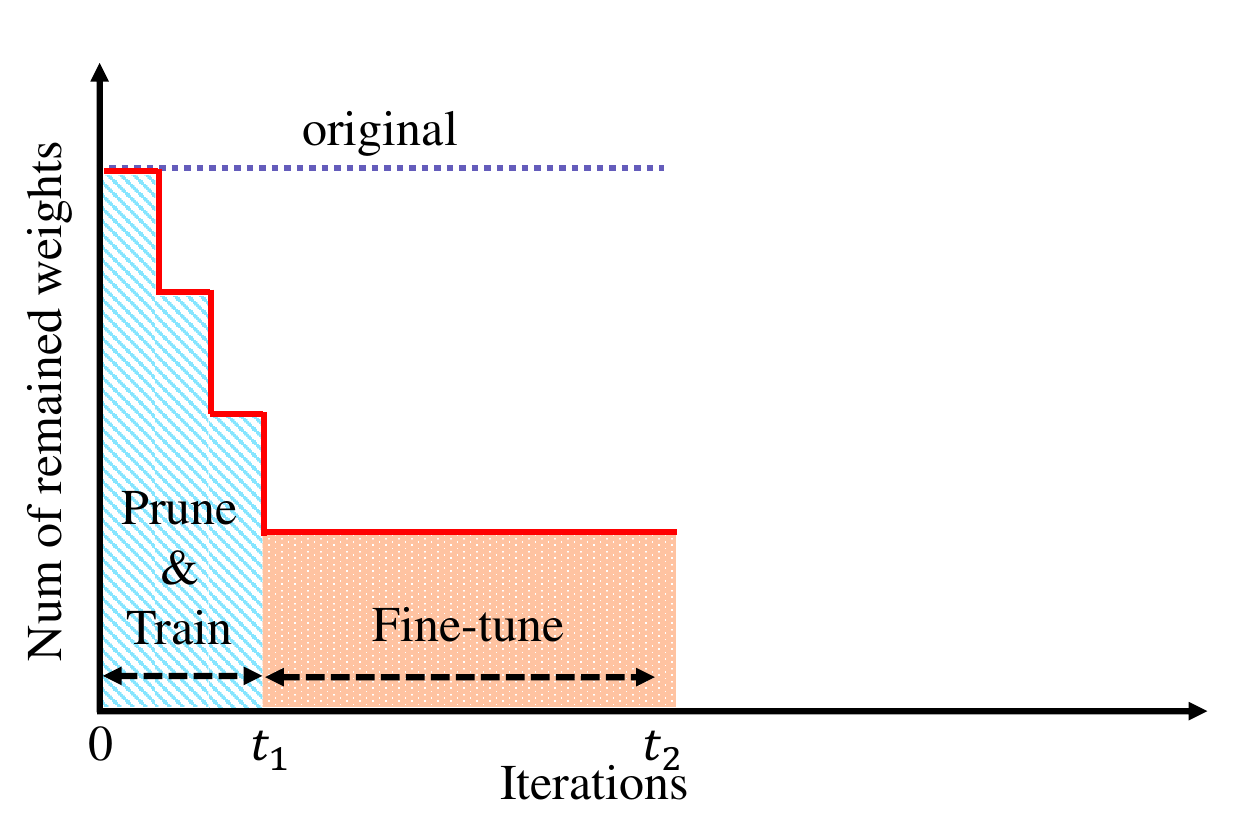}
     \vspace{-0.2cm}   
     \label{Fig:typical-sparsity-changes-b}
     \vspace{-0.2cm} 
  \end{minipage} 
  }
  \subfloat[GFP \cite{liu2021group} as an example of PAT]{  
  \begin{minipage}{5cm}
       \includegraphics[width=5cm,height=3.5cm]{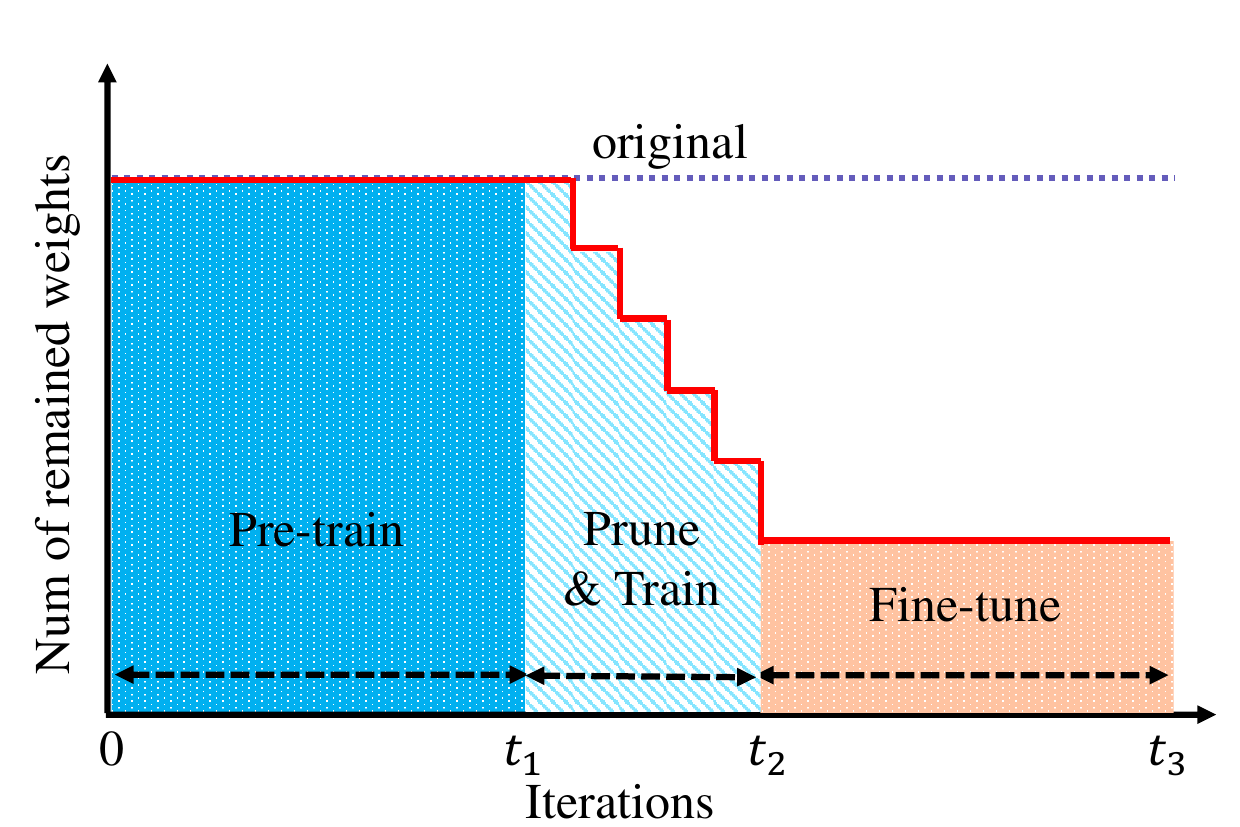}
     \vspace{-0.2cm}   
     \label{Fig:typical-sparsity-changes-c}
     \vspace{-0.2cm} 
  \end{minipage} 
  }  
 \caption{The illustration of typical sparsity changes for PBT, PDT, and PAT (best view in color and zoom in). PBT is usually the cheapest, while PAT is the most expensive.}
 \label{Fig:typical-sparsity-changes}
\vspace{-0.4cm}
\end{figure*}

This section distinguishes three pipelines for static pruning, as illustrated in Fig.~\ref{fig:pruning-pipelines-red}, and run-time pruning. The examples of the three pipelines of static pruning are shown in Fig.~\ref{Fig:typical-sparsity-changes}. Some notations and statistics on three static pruning pipeline types are shown in Appendix A and D, respectively.

\subsection{Pruning Before Training}
Pruning Before Training (\textbf{PBT}) (as shown in Table~\ref{tab:puning-at-initialization}), also called foresight pruning \cite{wang2020picking} or pruning at initialization \cite{lee2019snip,tanaka2020pruning}, represents a class of pruning methods that use randomly initialized weights for pruning the network. The principal motivation of PBT methods is to eliminate the cost of pre-training. Without loss of generality, we define a neural network as a function $f(\mathbf{x};W\odot M)$. The mask $M$ is used for pruning initialized weights $W_{0}$ sampled from a specific initialization distribution. After pruning, the network $f(\mathbf{x};W_{0} \odot M^{'})$ is trained to converge to $f(\mathbf{x};W_{t} \odot M^{'})$ after $t$ epochs, where $M^{'}$ indicates the sparsity after pruning. 

\begin{figure*}[t]
\centering
\begin{minipage}{12.5cm}
  \includegraphics[width=12.5cm,height=6cm]{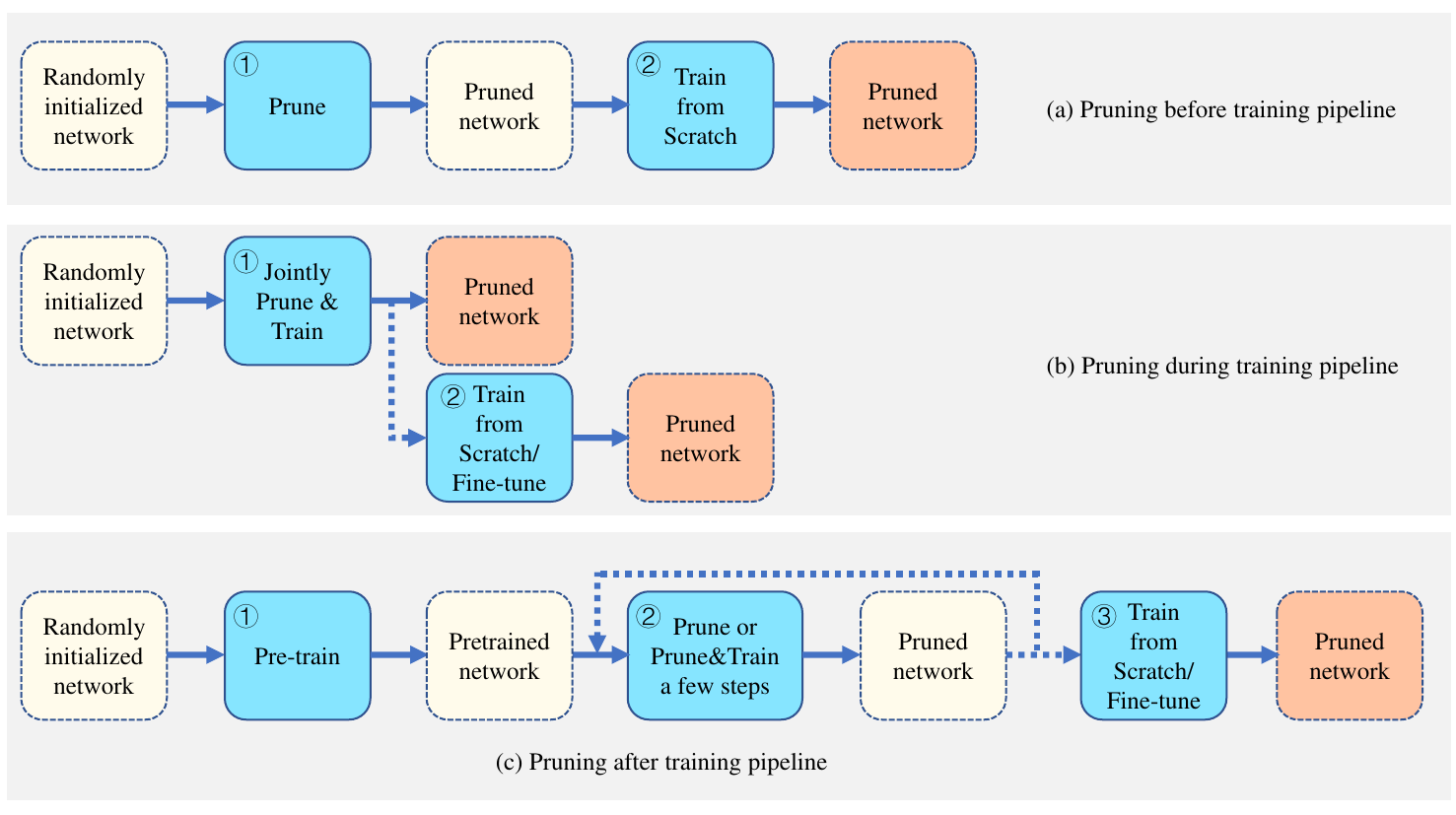}
\end{minipage}
\caption{The typical pipelines of static pruning. Dashed boxes represent models and solid boxes denote actions. Dashed arrows indicate optional steps.}
\label{fig:pruning-pipelines-red}
\vspace{-0.4cm}
\end{figure*}

PBT usually follows two steps: directly prunes untrained dense network based on a specific criterion and then trains the sparse network to convergence for high performance, as illustrated in Fig.~\ref{fig:pruning-pipelines-red} (a). The second step is related to static sparse training \cite{liu2022unreasonable} that aims to train a sparse network with a fixed sparse pattern. By avoiding the time-consuming pre-training process, PBT methods achieve the gains at training and inference time. 

\begin{table*}[t]
  \caption{Representative methods of pruning before training. ``U/S'' denotes unstructured or structured pruning.\protect\footnotemark}
  \vspace{-0.1cm}
  \centering
  \label{tab:puning-at-initialization}
  \begin{tabular}{l|ccccc}
    \Xhline{0.3ex}    
    Method & Criterion & U/S & Data-Driven (Y/N) &  One-shot (Y/N) \\
    \hline
    SNIP (2019) \cite{lee2019snip} & $\left |\nabla_{W}\mathcal{L}(W) \odot W \right |$ & U & Y & Y\\
    GraSP (2020) \cite{wang2020picking} & $-(H \nabla_{W}\mathcal{L}(W)) \odot W$ & U & Y & Y\\
    Smart-Ratio (2020) \cite{su2020sanity} & keep-ratios & U & N & Y\\
    SynFlow (2020) \cite{tanaka2020pruning} & $\nabla_{W} \mathcal{R}_{SF}(W)\odot W$ & U & N & N\\ 
    PFS (2020) \cite{wang2020pruning} & $l_1$-norm based sparsity regularization & S & Y & N\\
    RST (2022) \cite{bai2022dual} & $l_2$-norm based sparsity regularization & U & Y & N\\
    \Xhline{0.3ex}
\end{tabular}
\vspace{-0.3cm}
\end{table*}
\footnotetext{See Appendix~A for the notations in Table~\ref{tab:puning-at-initialization} \~{} Table~\ref{tab:pruning-after-training}.}

Currently, PBT primarily targets CNNs. \citet{lee2019snip} pioneer the research of PBT and propose a Single-shot Network Pruning (\textbf{SNIP}) method to remove the weights whose absence leads to the slightest loss change. \citet{lee2020signal} explain the feasibility of SNIP through signal propagation, empirically find pruning damages the dynamical isometry \cite{Saxe2014exact} of neural networks, and propose a data-free orthogonal initialization, an approximation of exact isometry, to prune random networks. Different from the signal propagation perspective in \cite{lee2020signal}, which focuses on initialization scheme, \citet{wang2020picking} propose Gradient Signal Preservation (\textbf{GraSP}) to exploit gradient norm after pruning to remove weights that have the least effect on the gradient flow. \citet{tanaka2020pruning} go a step further and point out that an effective subnetwork can be identified without training and looking at the data. They propose a data-free pruning algorithm called Iterative Synaptic Flow Pruning (\textbf{SynFlow}) that uses the concept of synaptic flow to consider the interaction of different layers and avoid layer collapse through gradual pruning. \citet{gebhart2021unified} exploit the Path Kernel (i.e., the covariance of path values in a network) based on Neural Tangent Kernel \cite{jacot2018neural} to unify several initialization pruning methods, such as SNIP \cite{lee2019snip}, CraSP \cite{wang2020picking}, SynFlow \cite{tanaka2020pruning}, under a single framework. Some works (\cite{su2020sanity,frankle2021pruning}) suggest that the delicate pruning criteria may not be crucial for performance, but the layerwise sparsity. For example, \citet{su2020sanity} propose randomly pruning each layer by using a series of layerwise keep-ratios (smart-ratios) without using any data. Inspired by Weight Agnostic Neural Networks (\textbf{WANNs}) \cite{gaier2019weight}, \citet{ramanujan2020what} empirically find that with an untrained network grows wider and deeper, it will contain a subnetwork that performs as well as a trained network with the same number of parameters. Then they propose the edge-popup method to identify such randomly initialized subnetworks. Unlike this method, which maintains randomly weighted values, \citet{bai2022dual} propose Dual Lottery Ticket Hypothesis (\textbf{DLTH}) where both the subnetwork and weights are randomly selected at initialization and propose Random Sparse Network Transformation (\textbf{RST}) which fixes the sparse architecture but gradually train the remaining weights.

Some recent works (\cite{wang2020pruning,liu2022unreasonable,hoang2023revisiting}) explore why an effective subnetwork can be identified at initialization. For example, \citet{liu2022unreasonable} empirically demonstrate that the network size and appropriate layer-wise prune ratios are two vital factors in training a randomly pruned network at initialization from scratch to match the performance of the dense models. In contrast, \citet{kumar2024free} find that at the same prune ratio, subnetworks derived by pruning during or after training exhibit higher effective parameter count and greater expressiveness compared to those pruned at initialization.

\subsection{Pruning During Training}
Pruning During Training (\textbf{PDT}) (as shown in Table~\ref{tab:pruning-during-training}) generally 
takes a randomly initialized dense network $f(\mathbf{x};W_{0})$ as the input model and jointly trains and prunes the neural network by updating weights $W$ and masks of weights (or filters, channels, etc.) $M$ during training. These dynamic schemes change the masks and produce the subnetwork $f(\mathbf{x};W_{t} \odot M_{t})$ after $t$ iterations/epochs. After pruning, many PDT methods (\cite{huang2018data,evci2020rigging,zhao2019variational,he2018soft,liu2019metapruning,ning2020dsa,cho2023pdp}) directly obtain the subnetworks without the need for training-from-scratch or fine-tuning. The typical pipeline of PDT is illustrated in Fig.~\ref{fig:pruning-pipelines-red} (b). PDT methods have been less explored due to the more complicated dynamic process than PBT and PAT methods. We summarize the main prior solutions into four paradigms: (1) sparsity regularization based, (2) dynamic spare training based, (3) score-based, and (4) differentiable pruning based. Methods related to (2) conduct sparse-to-sparse training, while the other three take dense-to-sparse training.

\subsubsection{Sparsity Regularization based Methods}
\label{sparse-regularization-technique}
Sparsity regularization technique \cite{wen2016learning} is commonly used in PDT methods. This category of methods starts with dense networks, imposes sparse constraints on loss functions, and usually zeros out some weights or their masks during training. The main effort is to design the effective target loss function $\mathcal{L}$ with an advanced penalty scheme and efficient optimization algorithms. For example, \citet{wen2016learning} propose Structured Sparsity Learning (\textbf{SSL}) to learn a sparse structure by group LASSO \cite{yuan2006model} regularization during training. However, SSL requires computing the gradients of the regularization term w.r.t. all the weights, which is non-trivial.   \citet{gordon2018morphnet} propose \textbf{MorphNet} that reuses the parameters of Batch Normalization (\textbf{BN}) and conducts sparsity regularization on these parameters. However, some networks (e.g., certain VGGNets \cite{simonyan2015very}) have no BN layers. Instead of reusing BN parameters, some works associate scaling factors with channels, layers, etc. For example, \citet{huang2018data} propose Sparse Structure Selection (\textbf{SSS}) that associates scaling factors for CNN micro-structures (e.g., channels, residual blocks) and exploit sparsity regularization to force the output of the micro-structures to zero, rather than pushing the weights in the same group to zero as in \cite{wen2016learning}. In addition, SSS does not require extra fine-tuning in \cite{wen2016learning}. 
\citet{li2019compressing} propose Factorized Convolutional Filter (\textbf{FCF}) which introduces a binary scalar to each filter and proposes a back-propagation algorithm with Alternating Direction Method of Multipliers (\textbf{ADMM}) \cite{boyd2010distributed} to train the weights and the scalars during training jointly.

\begin{table*}[t]
  \caption{Representative methods of pruning during training. ``Retrain'' refers to training from scratch or fine-tuning, ``U/S'' denotes unstructured or structured pruning.}
  \vspace{-0.1cm}
  \centering
  \label{tab:pruning-during-training}
  \begin{tabular}{l|ccc}
    \Xhline{0.3ex}    
    Method & Object Function/Criterion & Retrain (Y/N) & U/S\\
    \hline
    Network Slimming (2017) \cite{liu2017learning} & $\textrm{min}_{W,\boldsymbol{\gamma}}\ell(\mathbf{y},f(\mathbf{x};W))+\lambda \lVert \boldsymbol{\gamma} \rVert_{1}$ & Y & S \\ 
    SSS (2018) \cite{huang2018data} & $\textrm{min}_{W,\boldsymbol{\gamma}}\ell(\mathbf{y},f(\mathbf{x};W,\boldsymbol{\gamma}))+\mathcal{R}(W)+\lambda \lVert \boldsymbol{\gamma} \rVert_{1}$ & N & S\\
    SET (2018) \cite{mocanu2018scalable} & $\lVert W \rVert$ for drop and random for grow & N & U \\
    DSA (2020) \cite{ning2020dsa} & $\textrm{min}_{\mathcal{A}} \mathcal{L}_{v}(W(\mathcal{A}),\mathcal{A}))$ & N & S \\
    GraNet (2021) \cite{liu2021sparse} & $\lVert W \rVert$ for drop and gradient for grow & N & U\&S \\
    FreeTickets (2022) \cite{liu2022deep} & $\lVert W \rVert$ for drop and gradient for grow & N & U \\
  \Xhline{0.3ex}
\end{tabular}
\vspace{-0.5cm}
\end{table*}

\subsubsection{Dynamic Sparse Training based Methods}
\label{dynamic-sparse-neural-network-training}
A class of the PDT methods (\cite{mocanu2018scalable,dai2019nest,mostafa2019parameter,evci2020rigging,liu2021sparse,lin2020dynamic,dettmers2019sparse}) take randomly initialized sparse network rather than dense network as the input model. Subsequently, a common method involves pruning a fraction of unimportant weights and then regrowing the same number of new weights to adjust the sparse architecture. By repeating the prune-and-grow cycle during training, these method keeps searching for better sparse architecture, a process classified as dynamic sparse training in \cite{evci2020rigging}. 

For example, \citet{mocanu2018scalable} propose Sparse Evolutionary Training (\textbf{SET}) that removes the smallest positive and most negative weights and grows new weights in random locations. 
Instead of pruning a fixed fraction of weights at each redistribution step, such as in SET \cite{mocanu2018scalable}, \citet{mostafa2019parameter} propose Dynamic Sparse Reparameterization (\textbf{DSR}) which uses an adaptive threshold for pruning. In addition, DSR \cite{mostafa2019parameter} reallocates weights across layers and does not restrict to the inner-layer weight redistribution in SET \cite{mocanu2018scalable}. \citet{liu2022deep} propose an ensemble method \textbf{FreeTickets} that ensemble sparse subnetworks created by sparse-to-sparse methods. Rather than using a random regeneration scheme, \citet{liu2021sparse} propose Gradual Pruning with zero-cost Neuroregeneration (\textbf{GraNet}) to remove connections of networks based on their weight magnitudes and regrow connections of networks based on their gradient. They argue that even the weights with zero gradient values indicate the connection importance. \citet{sokar2022dynamic} pioneer to explore dynamic sparse training in Reinforcement Learning (\textbf{RL}). 

\citet{evci2022gradeint} analyze the reasonability of dynamic sparse training and find that sparse networks have poor gradient flow at initialization, but dynamic sparse training significantly improves gradient flow. 

\subsubsection{Score-based Methods}
\label{hand-craft-pruning-score-criteria}
Some PDT methods exploit scoring criteria to prune networks during training. \citet{he2018soft} propose Soft Filter Pruning (\textbf{SFP}) filter pruning method which can train networks from scratch and prune them simultaneously by using the $l_2$ norm of each filter as its importance measure. Instead of associating masks with filters, they directly set the pruned filter weights to zero, which can be updated from zero through the forward-backward process. Therefore, pruned filters in one epoch can be recovered in the next. However, SFP requires manually preset a prune ratio for each layer. \citet{he2019filter} propose Filter Pruning via Geometric Median (\textbf{FPGM}) to prune redundant filters that are nearest to the geometric median \cite{fletcher2008robust} of the filters within the same layer. However, FPGM also requires a pre-defined prune ratio for each layer. 
\citet{liu2017learning} propose a method called \textbf{Network Slimming} that introduces a scaling factor for each channel and jointly trains the weights and the scaling factors by adding the regular loss $\ell$ with sparsity regularization on the factors, and the magnitudes of these scaling factors are used as the filter scores. In practice, they reuse the $\gamma$ parameters in BN layers as the scaling factors. 

\subsubsection{Differentiable Pruning based methods}
Advances in differentiable network compression (\cite{liu2019darts,zhang2021idarts}) popularize differentiable techniques for pruning, such as \cite{ning2020dsa,guoO2020dmcp,ma2022differentiable,cho2023pdp,yu2023xpruner}. For example, \citet{ning2020dsa} propose Differentiable Sparsity Allocation (\textbf{DSA}) to set layer-wise prune ratios. They soften the hard pruning with a differentiable method using masks drawn from a distribution governed by prune ratios, allowing for gradient calculation of the target loss w.r.t. these ratios. These gradients then measure layer sensitivity. \citet{guoO2020dmcp} model channel pruning as a differentiable Markov process with architecture parameters, representing each channel as a state and using state transitions to determine the probability of retaining a channel based on the retention of its predecessor. Unlike \cite{ning2020dsa} and \cite{guoO2020dmcp}, \citet{cho2023pdp} generate soft pruning masks for weights without extra trainable parameters to accomplish differentiable pruning for CNNs and Transformers.

\begin{table*}[th]
  \caption{Representative methods of pruning after training. ``U/S'' denotes unstructured or structured pruning.}
  \vspace{-0.1cm}
  \centering
  \label{tab:pruning-after-training}
  \begin{tabular}{l|cc}
    \Xhline{0.3ex}    
    Method & Object Function/Criterion & U/S\\
    \hline
    Auto-Balance (2018) \cite{ding2018auto} & score($\mathcal{F}_{i,j}$)=$\lVert \mathcal{F}_{i,j} \rVert_{1}$ & S \\
    LTH (2019) \cite{frankle2019lottery} & score($w_{i}$)=$\lVert w_{i} \rVert_{1}$ & U\\
    Taylor-FO-BN (2019) \cite{molchanov2019importance} &   score($w_{i}$)=$(\nabla_{w_{i}}\ell \odot w_{i})^2$ & S\\
    SSR (2019) \cite{lin2019towardscompact} & $\mathop{\textrm{min}}\limits_{W}\frac{1}{N}\sum_{i=1}^{N}\ell(\mathbf{y}_i,f(\mathbf{x}_i,W))+\lambda \mathcal{R}(W)$ & S \\    
    GReg (2021) \cite{wang2021neural} & score($\mathcal{F}_{i,j}$)=$\lVert \mathcal{F}_{i,j}\rVert_{1}$ & U\&S\\
    GFP (2021) \cite{liu2021group} &  score($c_{i}$)=$\sum_{a=1}^{N}(\nabla_{m_{i}}\ell_{a})^2$ or score($c_{i}$)=$\sum_{a=1}^{N}(\sum_{g\in Group}\nabla_{m_{i}^{g}}\ell_{a})^2$& S \\
    SparseGPT (2023) \cite{frantar2023sparsegpt} &  score($w_{ij}$)=$[|W|^{2}/\textrm{diag}((\mathbf{x}^{T}\mathbf{x}+\lambda I)^{-1})]_{ij}$ & U \\    
    Wanda (2023) \cite{sun2024simple} &  score($w_{ij}$)=$|w_{ij}|\cdot \lVert \mathbf{x}_{j} \rVert_{2}$ & U \\
    LLM-Pruner (2023) \cite{ma2023llmpruner} &  score($w_{i}$)=$|\nabla_{w_{i}}\mathcal{L} \odot w_{i}-\frac{1}{2}\sum_{j=1}^{N}(\nabla_{w_{i}}\mathcal{L} \odot w_{i})^2|$ & S \\
    ShortGPT (2024) \cite{men2024shortgpt} & $1-\frac{\mathbf{x}_{i}^{T}\mathbf{x}_{i+1}}{\lVert \mathbf{x}_{i} \rVert_{2} \lVert \mathbf{x}_{i+1} \rVert_{2}}$ & S \\
    LLM-Streamline (2024) \cite{chen2024compressing} & $\textrm{max}_\mathbf{layer_i}cos(\mathbf{x}_{i},\mathbf{x}_{i+n})$ & S \\
    \Xhline{0.3ex}    
\end{tabular}
\vspace{-0.4cm}
\end{table*}

\subsection{Pruning After Training}
\label{pruning-after-training}
Pruning After Training (\textbf{PAT}) (as shown in Table~\ref{tab:pruning-after-training}) is the most popular type of pruning pipeline because it is commonly believed that pre-training the dense network is necessary to obtain an efficient subnetwork \cite{liu2019rethinking}. Especially for large models, such as LLMs and diffusion models, pruning is specifically applied to pre-trained models. This class of pruning methods typically follows a Pretrain-Prune-Retrain (\cite{liu2021group}, \cite{ma2023llmpruner}) or Pretrain-Prune (\cite{frantar2023sparsegpt,ashkboos2024slicegpt}) process, as shown in Fig.~\ref{fig:pruning-pipelines-red} (c). (1) Pre-train a randomly initialized dense network $f(\mathbf{x};W_{0})$ to converge to $f(\mathbf{x};W_{t})$. (2) Prune the weights (or filters, neurons, etc.) that have the least influence on performance and fine-tune the pruned network $f(\mathbf{x};W_{t}^{'} \odot M^{'})$ for several iterations, where $W_{t}^{'}$ and $M^{'}$ are the weights and masks after pruning, respectively. This step can be done at least once (i.e., one-shot pruning) or multiple times (i.e., iterative pruning). (3) Train the remaining weights from scratch $f(\mathbf{x};W_{0} \odot M^{''})$ or fine-tune   $f(\mathbf{x};W_{t}^{''} \odot M^{''})$ to recover the performance \cite{renda2020comparing}, where $W_{t}^{''}$ and $M^{''}$ are the final results of weights and masks after the pruning process, respectively. Sparsity may gradually increase during the pruning process until it achieves the target. 

\subsubsection{LTH and its Variants}
Lottery Ticket Hypothesis (\textbf{LTH})  \cite{frankle2019lottery,liu2024survey} is one of the most influential hypotheses in the neural network pruning domain. Given a pre-trained network, LTH iteratively removes a percentage of the weights based on their magnitudes. After pruning, the remaining weights are retrained from scratch with the original initialization, rather than random reinitialization, to match the original networks' accuracy. It challenges the commonly held belief that pre-trained weights must be used for retraining and conjectures the existence of an independently trainable sparse subnetwork within a dense network. Inspired by LTH, there are various follow-up works to identify wider tickets across multiple kinds of neural networks (such as CNNs \cite{diffenderfer2021multi}, Generative Adversarial Networks (\textbf{GANs}) \cite{chen2021gans}, Variational AutoEncoders (\textbf{VAEs}) \cite{kalibhat2021winning}, Graph Neural Networks (\textbf{GNNs}) \cite{hui2023rethinking}, Transformer based models \cite{prasanna2020bert}) and understand LTH better, which can be classified into five main classes: (1) proposing a stronger lottery ticket hypothesis, (2) exploring the transferability of LTH, (3) generalizing LTH to other contexts, (4) theoretical justification, and (5) revisiting and questioning LTH. 

(1)~Some recent works (\cite{malach2020proving,orseau2020logarithmic,diffenderfer2021multi}) prove stronger hypothesis than LTH \cite{frankle2019lottery}. For example, \citet{diffenderfer2021multi} propose a stronger Multi-Prize LTH, which claims that winning tickets can be robust to extreme forms of quantization (i.e., binary weights and/or activations). Based on this, they introduce the Multi-Prize Tickets (\textbf{MPTs}) algorithm to find MPTs on binary neural networks for the first time. 

(2)~Some literature (\cite{morcos2019one,mehta2019sparse,gan2022playing,chen2021unified}) studies the transferability of a winning ticket found in a source dataset to another dataset, which provides insights into the transferability of LTH. For example, 
\citet{morcos2019one} find OneTicket that can generalize across a variety of datasets and optimizers within the natural image domain. \citet{mehta2019sparse} propose the ticket transfer hypothesis and transfer winning tickets for different image classification datasets. 

(3)~In addition to image classification, LTH has been extended to other contexts, such as node classification and link prediction (\cite{chen2021unified}), vision-and-language (\cite{gan2022playing}), and NLP (\cite{prasanna2020bert}). For example, \citet{chen2021unified} pioneer to generalize LTH to GNNs and propose Graph Lottery Ticket (\textbf{GLT}). \citet{prasanna2020bert} explore the existence of winning tickets for fine-tuned BERT \cite{devlin2019bert} and identify subnetworks that match the model's performance.

(4)~On one hand, some literature (\cite{zhang2021validating,zhang2021why,paul2023unmasking}) analyzes the reasons why LTH \cite{frankle2019lottery} is able to win. For example, \citet{zhang2021validating} exploit dynamical systems theory and inertial manifold to theoretically verify the validity of LTH. \citet{evci2022gradeint} observe that the success of LTH lies in effectively re-learning the original pruning solution they are derived. \citet{zhang2021why} pioneer to provide a formal justification for the improved generalization of winning tickets observed in experimental results from LTH.

(5)~On the other hand, some recent works (\cite{ma2021sanity,liu2021lottery,liu2019rethinking}) revisit and challenge the existence of LTH. For example, 
\citet{ma2021sanity} provide a more rigorous definition of LTH for precisely identifying winning tickets and find that whether and when the winning tickets can be identified highly replies on the training settings, such as learning rate, training epochs, and architecture characteristics like network capacities and residual connections. It is more likely to find winning tickets by using a small learning rate or an insufficient number of training epochs. 
 
It is worth pointing out that in some works \cite{wang2022recent,sehwag2020hydra}, LTH \cite{frankle2019lottery} is classified as a PBT method. However, LTH selects masks based on a pre-trained network, which does not conform to the definition of PBT that attempts pruning the initialized network before training. Therefore, it is more reasonable to classify LTH as a PAT method. 

\subsubsection{Other score-based Methods}
The most straightforward and intuitive way to select pruning candidates is to evaluate them based on their norms. For example, \citet{han2015deep} propose to measure weight importance by its absolute value. In addition to norm-based criteria, evaluating loss change with and without the weights is also popular. For example, \citet{nonnenmacher2022sosp} propose Second-order Structured Pruning (\textbf{SOSP}) to selectively zero out filter masks to minimize the effects of the loss change from removing some filters. \citet{ma2023llmpruner} propose LLM-Pruner to pioneer the removal of unimportant coupled channels and multi-attention heads in LLMs using the first-order Taylor expansion to estimate loss changes and fine-tune the pruned models using LoRA \cite{hu2022lora}. \citet{fang2023structural} present Diff-Pruning which scores weights in diffusion models using the first-order Taylor expansion over pruned timesteps. \citet{shi2023upop} introduce Unified and Progressive Pruning (\textbf{UPop}), which prunes large multimodal models by leveraging accumulated trainable mask gradients during each iteration of the search phase. In addition to using loss change, some works (\cite{men2024shortgpt,nova2023gradient,dery2024everybody,kim2024shortened}) design new metrics. For example, \citet{men2024shortgpt} introduce a metric called Block Influence (\textbf{BI}), which assesses the significance of a layer by measuring the extent to which it alters the hidden states and removes redundant layers of LLMs. \citet{kim2024shortened} assess the significance of a layer in LLMs by measuring its impact on perplexity (\textbf{PPL}) upon its removal.

\subsubsection{Sparsity Regularization based Methods} 
Some works (\cite{he2017channel,lin2019towards,lin2019towardscompact,xia2022structured,xia2024sheared}) exploit sparsity regularization technique. For example, \citet{he2017channel} propose an alternative two-step algorithm that introduces a scalar mask to each channel of a pre-trained CNN model, selects redundant channels based on LASSO regression \cite{tibshirani1996regression}, and reconstructs the outputs of unpruned channels using linear least squares. Energy-Constrained Compression (\textbf{ECC}) \cite{yang2019ecc} builds an energy consumption model via a bilinear regression function. Network Pruning via Performance Maximization (\textbf{NPPM}) \cite{gao2021network} trains a performance prediction network and uses it as a proxy of accuracy to guide searching for subnetworks based on regularization penalty. \citet{fang2023depgraph} develop a general method called \textbf{DepGraph} to analyze dependencies in various network structures (e.g., CNNs, RNNs, GNNs, Transformers) and propose a structured pruning based on sparsity regularization. Using the $l_{0}$ regularization approach, \citet{xia2024sheared} efficiently learn pruning masks that match the target architecture and maximize performance by jointly optimizing weights and pruning masks with a min-max objective for LLMs. Some methods (\cite{wang2021neural,wang2023trainability,ding2018auto,fang2023depgraph}) select important weights (or filters, neurons, etc.) by combining norm-based criteria and sparsity regularization.

\subsubsection{Pruning in Early Training}
Instead of fully training a network from $f(\mathbf{x};W_{0})$ to $f(\mathbf{x};W_{T})$, this class of methods explore 
the network architecture by training a network only for a few iterations or epochs, i.e., $f(\mathbf{x};W_{t})$, where $t<<T$. For example, \citet{you2020drawing} propose Early-Bird (\textbf{EB}) tickets which indicate that winning tickets can be identified at the early training stage via inexpensive training schemes (e.g., early stopping and low-precision training) at large learning rates and achieve similar performance to the dense network. Inspired by \cite{you2020drawing}, \citet{chen2021earlybert} propose \textbf{EarlyBERT} that identifies structured winning tickets in the early stage of BERT \cite{michel2019sixteen} training. \citet{frankle2020linear} find that, in large-scale settings (such as ResNet-50 and Inception-v3 on ImageNet), the subnetworks that exhibit stability to SGD noise are able to reach full accuracy early in training.

\subsubsection{Post-Training Pruning}
In contrast to the general PAT methods that follow the Pretrain-Prune-Retrain procedure, recently proposed post-training pruning methods simplify the three-step process Pretrain-Prune. It involves pruning a pre-trained model $f(\mathbf{x};W_{t})$ without retraining, typically achieving negligible accuracy loss by using compensation mechanisms to mitigate performance degradation. This class of pruning methods is particularly attractive for billion-parameter models because retraining such pruned models is still very expensive. For example, \citet{kwon2022fast} propose a structured post-training pruning framework for Transformers that features Fisher-based mask search, rearrangement, and tuning, achieving pruning on a single GPU in three minutes without retraining. SparseGPT \cite{frantar2023sparsegpt}, a groundbreaking unstructured post-training pruning method for LLMs, tackles the pruning problem as an approximate sparsity reconstruction problem and prunes LLMs at least 50\% sparsity with minor accuracy loss without retraining. To address SparseGPT's reconstruction cost, Wanda~\cite{sun2024simple} uses weight magnitudes and input norms to facilitate unstructured post-training pruning on LLMs without updating weights. Unlike SparseGPT and Wanda, SliceGPT \cite{ashkboos2024slicegpt} implements structured post-training pruning using orthogonal matrix transformations and principal component analysis (\textbf{PCA}) to remove columns and rows of the weight matrices in LLMs. FLAP \cite{an2024fluctuationbased} introduces a fluctuation metric and uses a bias compensation mechanism for LLMs' performance recovery.

\subsection{Run-time Pruning}
The prior works on pruning usually focus on static pruning methods where the pruned model is reused for different inputs. In contrast, some methods prune neural networks according to individual inputs dynamically, known as run-time pruning \cite{rao2019runtime}. This line of work is based on the premise that for a given task, the difficulty of producing accurate output can vary, implying that necessary model capacities for different inputs are different \cite{tang2021manifold}. For example, \citet{rao2019runtime} propose a Runtime Network Routing (\textbf{RNR}) framework to conduct dynamic routing based on the input image and current feature maps and select an optimal path subset for compression. \citet{tang2021manifold} point out that the importance of channels highly depends on the input data and propose to generate different subnetworks for each instance. At inference, only channels with saliencies larger than the threshold need to be computed, and the redundant features are skipped. \citet{hua2019channel} exploit input-specific characteristics and propose \textbf{CGNets} to predict unimportant regions by the partial sum of the output activation by performing convolution on a subset of input channels. \citet{gao2019dynamic} propose Feature Boosting and Suppression (\textbf{FBS}) to predict the saliency of channels and skip those with less contribution to the classification results at run-time. \citet{elkerdawy2022fire} pose dynamic model pruning as a self-supervised binary classification problem. \citet{meng2022contrastive} propose Contrastive Dual Gating (\textbf{CDG}), another self-supervised dynamic pruning method that uses contrastive learning \cite{hadsell2006dimensionality}. \citet{tuli2023acceltran} introduce DynaTran, which prunes activations at runtime based on the magnitude of the input matrix to enhance transformer inference throughput.

\section{Pruning Criteria}
\label{criteria}
In this section, we summarize some commonly used pruning criteria for evaluating the importance of weights (or filters, neurons, etc.) from different perspectives, including magnitude (\cite{han2015learning,see2016compression,lubana2021gradient,sehwag2019towards}), norm (\cite{he2018soft,li2017pruning}), saliency and/or sensitivity (\cite{zhao2019variational,molchanov2017pruning}), and loss change (\cite{molchanov2017pruning,li2021towards,you2019gate,nonnenmacher2022sosp,liu2021group}). There is no rigid boundary between these criteria, but a different emphasis.

\subsection{Magnitude-based Pruning}
\cite{hanson1988comparing} is one of the earliest works that propose magnitude-based pruning to reduce hidden units. \citet{han2015deep} popularize magnitude-based pruning for deep neural networks, which prunes the lowest-magnitude weights. It is based on the assumption that weights with smaller absolute values tend to have the least influence on the network's output \cite{li2017pruning}. The formulation is defined as 
\begin{equation}
    m_i = \begin{cases} 
    1: if \ \lVert w_i \rVert_{1} \ge a \\
    0: if \ \lVert w_i \rVert_{1} < a 
    \end{cases},
\end{equation}
where $a$ is a threshold. Some criteria combine weight magnitude with activation magnitude. For example, the score for a weight $w_{ij}$ in Wanda \cite{sun2024simple} is defined by:
\begin{equation}
    s_{ij} = \left | w_{ij} \right |\cdot \lVert \mathbf{x}_{j}\rVert_{2}.
\end{equation}

In addition to weight and activation, magnitude-based pruning can be applied to other values (\cite{yu2022width,dery2024everybody}). For example, \citet{dery2024everybody} propose to prune LLMs by selecting modules with the highest magnitude of module relevance until the pruning constraint is met.

Magnitude-based criteria can be applied to either unstructured (\cite{frankle2019lottery,see2016compression,sun2024simple}) or structured pruning (\cite{li2017pruning,dai2019nest,dery2024everybody}). For example, \citet{li2017pruning} score the filters by calculating the sum of the absolute magnitude of their weights. In addition, magnitude-based criteria can be combined with global/local and one-shot/iterative schedules. For example, the works in  \cite{rosenfeld2021predictability} and \cite{lee2021layeradaptive} propose magnitude-based iterative global pruning methods. \citet{lubana2021gradient} argue that magnitude-based pruning results in faster model convergence than magnitude-agnostic methods. 

\subsection{$l_p$ Norm}
Some methods (\cite{he2018soft,li2017pruning,sun2024simple}) use the $l_p$ norm to evaluate the importance of weights (or filters, neurons, etc.). For example, \citet{he2018soft} exploit the $l_p$ norm to evaluate the importance of the filter $\mathcal{F}_{i,j}$, as shown in Eq.(\ref{eq.4}).
\begin{equation}
    \lVert \mathcal{F}_{i,j} \rVert_{p}=\left(\sum_{n=1}^{c_{in}^{(i)}}\sum_{k_1=1}^{k^{(i)}}\sum_{k_2=1}^{k^{(i)}}\left | \mathcal{F}_{i,j}(n,k_1,k_2)\right |^p \right)^{\frac{1}{p}}
    \label{eq.4},
\end{equation}
where $k^{(i)}$ is the kernel size of layer $i$ in a network and $c_{in}^{(i)}$ is the number of channels at layer $i$. Weights with smaller $l_p$ norms are more likely to be pruned than those with higher $l_p$ norms. In addition, the importance value is often optimized with norm-based sparsity regularization (\cite{ding2018auto}), which is discussed in \ref{regularization}. 

\subsection{Sensitivity and/or Saliency}
Some works (\cite{lecun1989optimal,santacroce2023matters,zhao2019variational}) utilize sensitivity and/or saliency to evaluate the importance of weights (or filters, neurons, etc.). For example, \citet{lecun1989optimal} define weight saliency as the loss change induced by pruning that weight. \citet{lee2019snip} propose a saliency criterion called the connection sensitivity criterion as the normalized magnitude of the derivatives $g_{j}$:
\begin{equation}
    s_{j}(W;\mathcal{D})=\frac{\left| g_{j}(W;\mathcal{D})\right|}{\sum_{k=1}^{M}\left| g_{k}(W;\mathcal{D})\right|},
\end{equation}
where $s_j$ is the sensitivity of $w_j$, $W$ represents the network's weights, $g_{j}$ is the derivative of the loss $\mathcal L(W\odot M)$ w.r.t. the mask $m_{j}$. The higher the sensitivity, the more important the weight is. 
\citet{zhao2019variational} reformulate the BN layer by extending the scale factor $\gamma$ on the shift term $\beta$, which is treated as channel saliency. They reformulate BN as follows:
\begin{equation}
   \mathbf{x}_{out} = \gamma \cdot BN(\mathbf{x})+\tilde{\beta},    
\end{equation}
where $\tilde{\beta}=\gamma \cdot \beta$.
Rather than relying on the value of $\gamma$, unimportant channels are pruned based on $\gamma$'s distributions.  

\subsection{Loss Change} 
\label{losschange}
It is a widely used criterion to measure the importance of a weight (or filter, neuron, etc.) by evaluating the loss change of the network with and without it. The loss change is usually approximated in a Taylor expansion-based way, such as \cite{you2019gate,nonnenmacher2022sosp,liu2021group,ma2023llmpruner,fang2023structural,ouderaa2024llm}.

\textbf{The first-order Taylor expansion} is the most commonly used method for measuring the loss change. The loss change with a small perturbation at $W$ is defined as follows:
\begin{equation}
    \Delta \mathcal{L} = \mathcal{L}(W+\Delta W) - \mathcal{L}(W) = \nabla_{W} \mathcal{L}\ \Delta W.
\end{equation}

For example, \citet{you2019gate} introduce scaling factors $\boldsymbol{\lambda}$ to the BN and exploits the first-order Taylor expansion to estimate the loss change $\Delta \mathcal{L}$ caused by setting some scaling factors to zero as follows:
\begin{equation}
    \Delta \mathcal{L}(\boldsymbol{\lambda}) = \left | \boldsymbol{\lambda} \nabla_{\boldsymbol{\lambda}} \mathcal{L} - R_{1}(\boldsymbol{\lambda})\right | \approx \left | \boldsymbol{\lambda} \nabla_{\boldsymbol{\lambda}} \mathcal{L}\right |=\left| \frac{\partial \mathcal{L}}{\partial \boldsymbol{\lambda}} \boldsymbol{\lambda} \right|,
\end{equation}
where $R_{1}(\boldsymbol{\lambda})$ is the Lagrange remainder. The importance score of the $i$-th filter is defined as
\begin{equation}
    \textrm{Score}(\mathcal{F}_i) = \mathop{\sum}\limits_{(\mathbf{x},\mathbf{y})\in \mathcal{D}}\left | \frac{\partial \mathcal{L}(\mathbf{y},f(\mathbf{x};W)}{\partial \lambda_i} \lambda_i\right |,
\end{equation}
where $\lambda_{i}$ is the scalar factor of the $i$-th filter. 

\textbf{The second-order Taylor expansion} of the loss function is early used in \cite{lecun1989optimal,hassibi1992second} for removing unimportant weights and gradually exploited in many subsequent methods (\cite{wang2019eigendamage,kurtic2022optimal,nonnenmacher2022sosp,liu2021group,ouderaa2024llm}), which includes the first-order (gradient) term, the second-order (Hessian) term, and the higher-order terms are neglected. Without loss of generality, the approximation of the loss change leads to 
\begin{equation}
    \mathcal{L}(W+\Delta W) - \mathcal{L}(W) = \nabla_{W} \mathcal{L}\ \Delta W + \frac{1}{2}\Delta W^{T}H\Delta W,
\end{equation}
where $H=\nabla_{W}^2 \mathcal{L}(W)$. 

For example, \citet{liu2021group} apply the second-order Taylor expansion to approximate the loss change when removing a channel (setting its mask to 0):
\begin{equation}
\begin{aligned}
    s_i &= \Delta \mathcal{L} = \mathcal{L}(M-\textbf{e}_\textbf{i}) - \mathcal{L}(M) \approx -\textbf{e}_\textbf{i}^{T}\nabla_{M} \mathcal{L}+\frac{1}{2}\textbf{e}_\textbf{i}^{T}(\nabla_{M}^{2}\mathcal{L})\textbf{e}_\textbf{i} \\
    &= -\textbf{e}_\textbf{i}^{T}g + \frac{1}{2}\textbf{e}_\textbf{i}^{T}H\textbf{e}_\textbf{i}=-g_i+\frac{1}{2}H_{ii},
\end{aligned}    
\end{equation}
where $\textbf{e}_\textbf{i}$ is the one-hot vector with the $i$-th entry equals one, and $g$ is the gradient of the loss function $\mathcal{L}$ w.r.t. $M$.

\section{Learn to Prune}
\label{learntoprune}
In this section, we present some methods for learning to prune networks, including sparsity regularization (\cite{he2017channel,liu2017learning,huang2018data,li2020dhp, Voita2019analyzing}) and pruning methods based on meta-learning (\cite{liu2019metapruning,li2020dhp}), graph neural networks (\cite{zhang2022graph}), and reinforcement-learning (\cite{he2018amc,rao2019runtime}). 

\subsection{Sparsity Regularization based Pruning}
\label{regularization}
Sparsity regularization based pruning \cite{he2017channel,dery2024everybody} learns the weights and their masks by solving the following problem:
\begin{equation}
\mathop{\textrm{min}}\limits_{W,M} \mathcal{L}(W,M) \\,  
\label{sparsity-regularization}
\end{equation}
where $\mathcal{L}=\ell(W,M) + \lambda \mathcal{R}(\cdot)$.
One common way is to introduce a scaling factor vector $\boldsymbol{\gamma}$ for weights (or channels, filters, etc.). The network weights and the scaling factors $\boldsymbol{\gamma}$ are trained jointly with sparsity regularization imposed on the latter. The magnitude of the scaling factors is treated as the important scores. Specifically, $\mathcal{L}$ in Eq.~\ref{sparsity-regularization} can be exemplified as follows:
\begin{equation}
    \mathcal{L} = \frac{1}{N}\sum_{i=1}^{N}\ell(\mathbf{y}_i,f(\mathbf{x}_i;W, \boldsymbol{\gamma}))+\lambda \sum_{\gamma_{i} \in \boldsymbol{\gamma}}\mathcal{R}(\gamma_{i}).
\end{equation}

For example, \citet{he2017channel} cast channel selection as the minimization of reconstruction error in feature maps and formulate the channel pruning problem as follows:
\begin{equation}
\begin{aligned}
    & \mathop{\textrm{min}}\limits_{\beta, W} \frac{1}{2N}\lVert \mathbf{y} - \sum_{i=1}^{c} \beta_i \mathbf{x}_{c_{i}} w_{c_{i}}^T\rVert_F^2 + \lambda \lVert \beta \rVert_1, \\
    & \textrm{subject to} \ \lVert \beta\rVert_0 \leq c', \forall i \ \lVert w_{c_{i}} \rVert_{F} = 1. \\
\end{aligned}
\end{equation}
where $\lVert \cdot \rVert_{F}$ is the Frobenius norm, $\mathbf{x}_{c_{i}}$ is an $N \times k_{h}k_{w}$ matrix from $i$-th channel of input $\mathbf{x}$, $w_{c_{i}}$ is an $n \times k_{h}k_{w}$ weight matrix from $i$-th channel of $W$, $k_h$ and $k_w$ are the kernel height and width, respectively. $N$, $c$, $c'$, and $n$ represent the number of samples, channels, retained channels, and output channels. To solve this problem, \citet{he2017channel} use LASSO regression \cite{tibshirani1996regression} and a greedy strategy to select the unimportant channels. 

\subsection{Meta-Learning based Pruning}
Some works (\cite{liu2019metapruning,li2020dhp}) adopt meta-learning to prune models. For example, \citet{liu2019metapruning} train a meta network, PruningNet, to predict weights for different pruned networks. The PruningNet takes a network encoding vector $(v_1, v_2, ..., v_L)$ as input and outputs the weights $W$ of the pruned network:
\begin{equation}
    W = \textrm{PruningNet}(v_1,v_2,...,v_L),
\end{equation}
where $v_i$ is the number of the channels for the $i$-th layer. The weights and the corresponding accuracy of each pruned network are obtained by inputting the network encoding into the fully trained PruningNet. Considering the huge search space of network encoding vectors, the pruned network is found by evolutionary search under the constraints. 

\subsection{Graph Neural Network based Pruning}
Any network can be viewed as a graph. \citet{zhang2022graph} propose a method called GraphPruning for model compression. Specifically, GraphPruning designs a graph aggregator $G$ with weights $\mathbf{\theta}_{G}$, combined with the Fully Connected (FC) layers, to generate the weights $W = (w^{(1)},w^{(2)},...,w^{(L)})$ of the ``Pruned Network'' as follows:
\begin{equation}
\begin{aligned}
    & (n_1,n_2,...,n_{L}) = G(b_1,b_2,...,b_{L}|\mathbf{\theta}_{G}), \\
    & w^{(i)} = FC_{i}(n_{i}|\mathbf{\theta}_{i}),   
\end{aligned}
\end{equation}
where $b_{i}\in R^{1\times 7}$ denotes the embedding features of the $i$-th node, $n_i$ is the $i$-th column of the output with the graph aggregator, $\theta_{i}$ are the weights of the $i$th FC layer, and $w^{(i)}$ are the weights of the $i$-th pruned layer of the pruned network. Then, the ``Pruned Network'' is fully trained. The graph aggregator is responsible for extracting high-level features for each node, while each FC layer is used to generate reasonable weights for the ``Pruned Network''. Afterward, the best configuration of the ``Pruned Network'' under computational constraints is searched by RL methods, during which the weights of the graph aggregator and FCs are not updated.  

\subsection{Reinforcement Learning based Pruning}
Rather than using RL to search for the best configurations of the pruned networks as in \cite{zhang2022graph}, some AutoML pruning methods (\cite{he2018amc,rao2019runtime}) adopt RL to compress models automatically. For example, \citet{he2018amc} propose AutoML for Model Compression (\textbf{AMC}), which is based on Q-learning, a type of RL that focuses on how an agent should take actions to maximize the cumulative reward. Specifically, \citet{he2018amc} design the Deep Deterministic Policy Gradient (\textbf{DDPG}) agent to receive an embedding state $s_{i}$ of layer $l_{i}$ from the environment and output a sparsity ratio as action $a_{i}$. Then, layer $l_{i}$ is compressed with $a_{i}$ using a specific compression method (such as a channel pruning method). After that, the agent moves to layer $l_{i+1}$ and repeats the same process until the final layer $l_{L}$. The update process is as follows: 
\begin{equation}
\begin{aligned}
    Loss = \frac{1}{N}\sum_{i=1}^{N}(\mathbf{y}_{i}-Q(s_{i},a_{i}|W^{Q}))^{2}, \\
    \mathbf{y}_{i} = r_{i}-b+\gamma Q(s_{i+1},\mu \ (s_{i+1})|W^{Q}),
\end{aligned}
\end{equation}
where $b$ is the baseline reward, $\gamma$ is a discount factor used to avoid over-prioritizing short-term rewards, $W^{Q}$ are the weights of the network $Q$ following Block-QNN \cite{blockqnn}, and $r_{i}$ is the reward of the whole trajectory for the $i$-th sample.

\citet{he2018amc} observe that $\textrm{Error}$ is inversely-proportional to $log(\textrm{FLOPs})$ or $log(\#\textrm{Param})$. Based on this observation, the reward function is defined as:
\begin{equation}
\begin{aligned}
    &\textrm{R}_{\textrm{FLOPs}} = -\textrm{Error} \cdot log(\textrm{FLOPs}), \\
    &\textrm{R}_{\textrm{Param}} = -\textrm{Error} \cdot log(\# \textrm{Param}).  \\  
\end{aligned}
\end{equation}
This reward function provides an incentive for reducing FLOPs or the number of network parameters. 

\section{A Comprehensive Comparative Analysis}
\label{comparative}
In this section, we compare some pruning methods on commonly used models, including eight pairs of contrast settings for pruning, different layer-wise densities, and various supervision levels for pruning. To avoid the influence of specific functions on pruning results, we mainly use the same functions under contrast settings (experimental details in Appendix~B). Appendix~C provides a more extensive comparison across different methods.

\subsection{Unstructured vs. Structured Pruning}
Unstructured pruning methods (\cite{frantar2023sparsegpt, han2015deep}) remove weights anywhere and can achieve high prune ratios with little impact on accuracy. In contrast, structured pruning (\cite{liu2021group,nonnenmacher2022sosp,ma2023llmpruner}) conducts pruning on entire filters (or channels, neurons, layers, etc.), resulting in really compressed network and accelerated inference. However, the accuracy is often lower than that of unstructured pruning under the same prune ratio, weight-level scoring, pipeline, and learning schemes. The possible reason is that unstructured pruning only focuses on the importance of individual weights, while structured pruning forces structural coupling, which demands simultaneous pruning across multiple layers and expects all removed weights to be consistently unimportant. However, achieving consistency in identifying unimportant weights under the structural coupling constraints is challenging. \citet{amersfoort2020single} argue that SNIP-structured and GraSP-structured methods incur more noise than their vanilla unstructured counterparts. 

\begin{table}[t]
 \caption{Top-1 accuracy (\%) of unstructured and structured pruning on VGG-16. ``Ratio'' refers to the percentage of parameters removed from the original count. \textbf{Bold}/\underline{Underline} marks the best/second best performance, respectively, among the compared entities. Unless otherwise specified, ``Ratio'' and \textbf{Bold}/\underline{Underline} in other tables have the same meaning.}
 \vspace{-0.1cm}
  \centering
  \label{Table-unstructured-and-structured}
  \begin{tabular}{l|c|c|c|c}
    \Xhline{0.3ex}      
     Dataset & \multicolumn{2}{c|}{CIFAR-10} & \multicolumn{2}{c}{CIFAR-100} \\
     \hline
     Method \textbackslash Ratio (\%) & 10.00 & 40.00 & 10.00 & 50.00 \\
     \hline 
     SNIP-unstructured \cite{lee2019snip} & \textbf{93.84} & \textbf{93.73} & \textbf{73.09} & \textbf{72.67} \\
     SNIP-structured & \underline{93.74} & \underline{93.71} & \underline{72.82} & \underline{70.68} \\
     \hline 
     GraSP-unstructured \cite{wang2020picking} & \underline{93.58} & \textbf{93.18} & \textbf{72.46} &  \textbf{71.42} \\
     GraSP-structured & \textbf{93.67} & \underline{93.04} & \underline{72.43} & \underline{71.30} \\
     \Xhline{0.3ex}      
\end{tabular}
\vspace{-0.1cm}
\end{table}

\begin{table}[t]
  \centering
  \caption{Perplexity of unstructured, semi-structured, and structured pruning on LLMs with WikiText2~\cite{merity2016pointer}, where lower is better.  ``O'' denotes OPT~\cite{zhang2022opt}.}
  \vspace{-0.1cm}
  \label{tab:unstructured-and-structured-llm}
  \scalebox{0.85}{%
  \begin{tabular}{l|c|c|c|c|c}
  \Xhline{0.3ex}      
  \multirow{2}{*}{Method} & Ratio & \multicolumn{4}{c}{Model} \\
  \cline{3-6}
  & (\%) & O-125M & O-1.3B & O-2.7B & O-6.7B \\
  \hline 
  Unpruned & 0 & 27.65 & 14.62 & 12.47 & 10.86 \\
  \hline
  SparseGPT-unstruct~\cite{frantar2023sparsegpt} & \multirow{2}{*}{50.00}& \textbf{33.17} & \textbf{26.77} & \textbf{12.88} &  \textbf{11.92} \\
  SparseGPT-2:4~\cite{frantar2023sparsegpt} & & \underline{45.51} & \underline{29.44} & \underline{14.92} & \underline{13.01} \\
  \hline
  K-OBD-2:4~\cite{ouderaa2024llm} & \multirow{2}{*}{50.00}& \textbf{68.74} & \textbf{27.22} & \textbf{20.23} &  \textbf{15.55} \\
  K-OBD-struct~\cite{ouderaa2024llm} & & \underline{75.95} & \underline{37.68} & \underline{26.88} & \underline{25.54} \\
  \hline
  LLM-Surgeon-unstruct~\cite{ouderaa2024llm} & \multirow{3}{*}{50.00}& \textbf{30.30} & \textbf{15.47} & \textbf{12.68} &  \textbf{10.97} \\
  LLM-Surgeon-2:4~\cite{ouderaa2024llm} & & \underline{44.64} & 25.10 & \underline{14.64} & \underline{12.10} \\
  LLM-Surgeon-struct~\cite{ouderaa2024llm} & & 49.78 & \underline{22.95} & 17.15 & 14.90 \\
  \Xhline{0.3ex}      
\end{tabular}
}
\vspace{-0.3cm}
\end{table}

We compare unstructured and structured pruning methods on VGG-16 \cite{simonyan2015very} and report the best results in Table~\ref{Table-unstructured-and-structured} from three random runs. Additionally, we compare unstructured, semi-structured, and structured pruning methods on OPTs \cite{zhang2022opt} with data sourced from \cite{ouderaa2024llm} (Table~\ref{tab:unstructured-and-structured-llm}). As shown in Table~\ref{Table-unstructured-and-structured} and Table~\ref{tab:unstructured-and-structured-llm}, at the same prune ratio, unstructured pruning generally outperforms semi-structured (if any), which performs better than structured pruning. 

\subsection{One-shot vs. Iterative Pruning}
\label{oneshotiterative}
One-shot pruning methods score once and then prune the network to a target prune ratio. Conversely, the iterative pruning methods alternately process the score-prune-update cycle until achieving the target prune ratio. As a result, the pruning cost in one-shot methods is usually negligible, greatly saving pruning efforts. However, these methods are not beneficial to those significant weights whose importance is not immediately apparent at the beginning \cite{morcos2019one}. Therefore, one-shot pruning generally requires more carefully designed scoring criteria to match the performance of the original network. In addition, the results in \cite{tanaka2020pruning} show that one-shot pruning may more easily suffer from layer collapse, resulting in a sharp accuracy drop. In contrast, iterative methods require more pruning cost but generally yield better accuracy \cite{you2019gate,li2017pruning,liu2021lottery,mehta2019sparse,huang2004learning}. 

\begin{figure}[t]
\centering
  \subfloat[VGG-16 on CIFAR-10]{
  \begin{minipage}{4cm}
  \includegraphics[width=4cm,height=2.5cm]{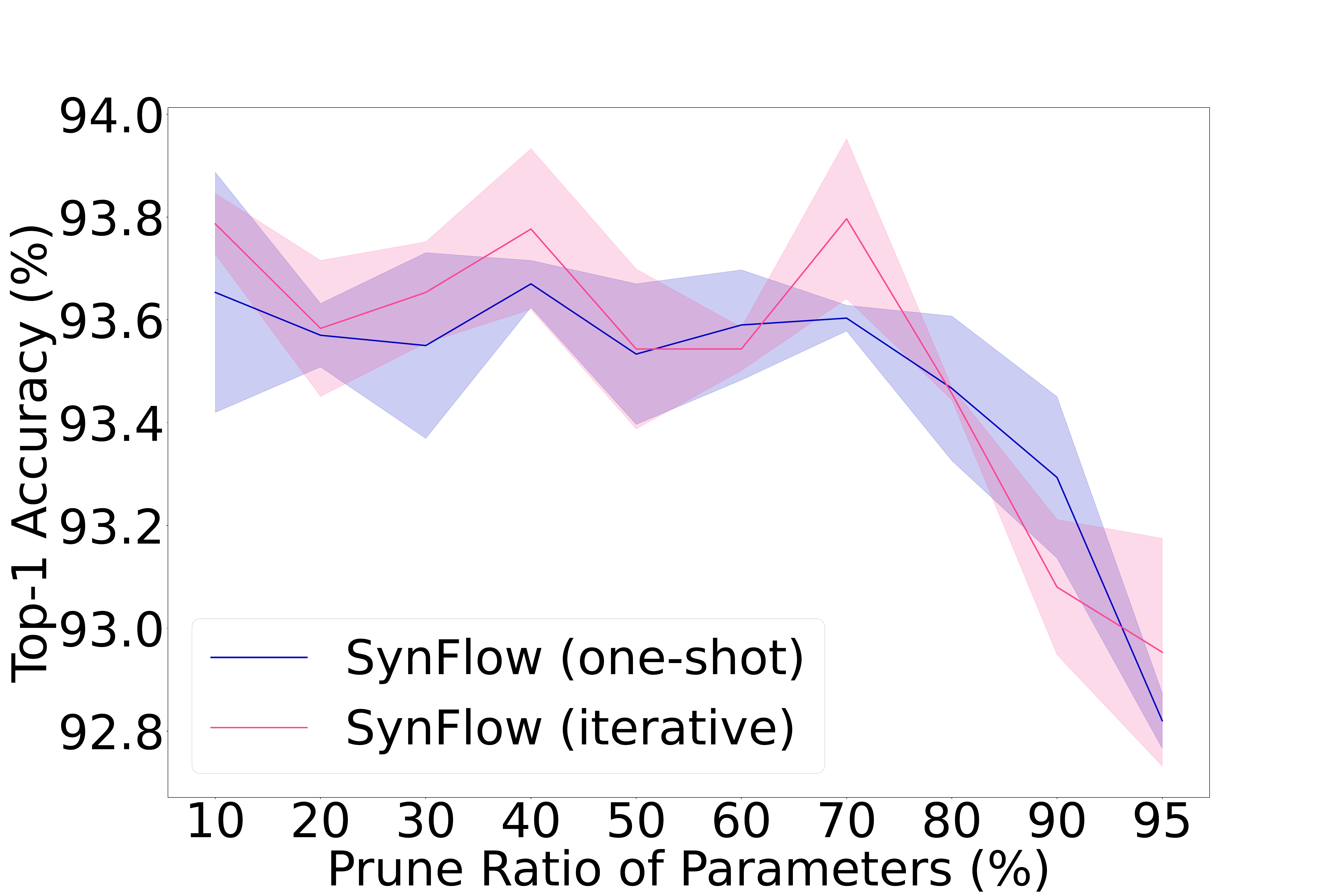}
     \vspace{-0.2cm} 
     \label{Fig:one-shot-iterative-pruning-a}
  \end{minipage}
  }
  \subfloat[ResNet-32 on CIFAR-10]{  
  \begin{minipage}{4cm}
  \includegraphics[width=4cm,height=2.5cm]{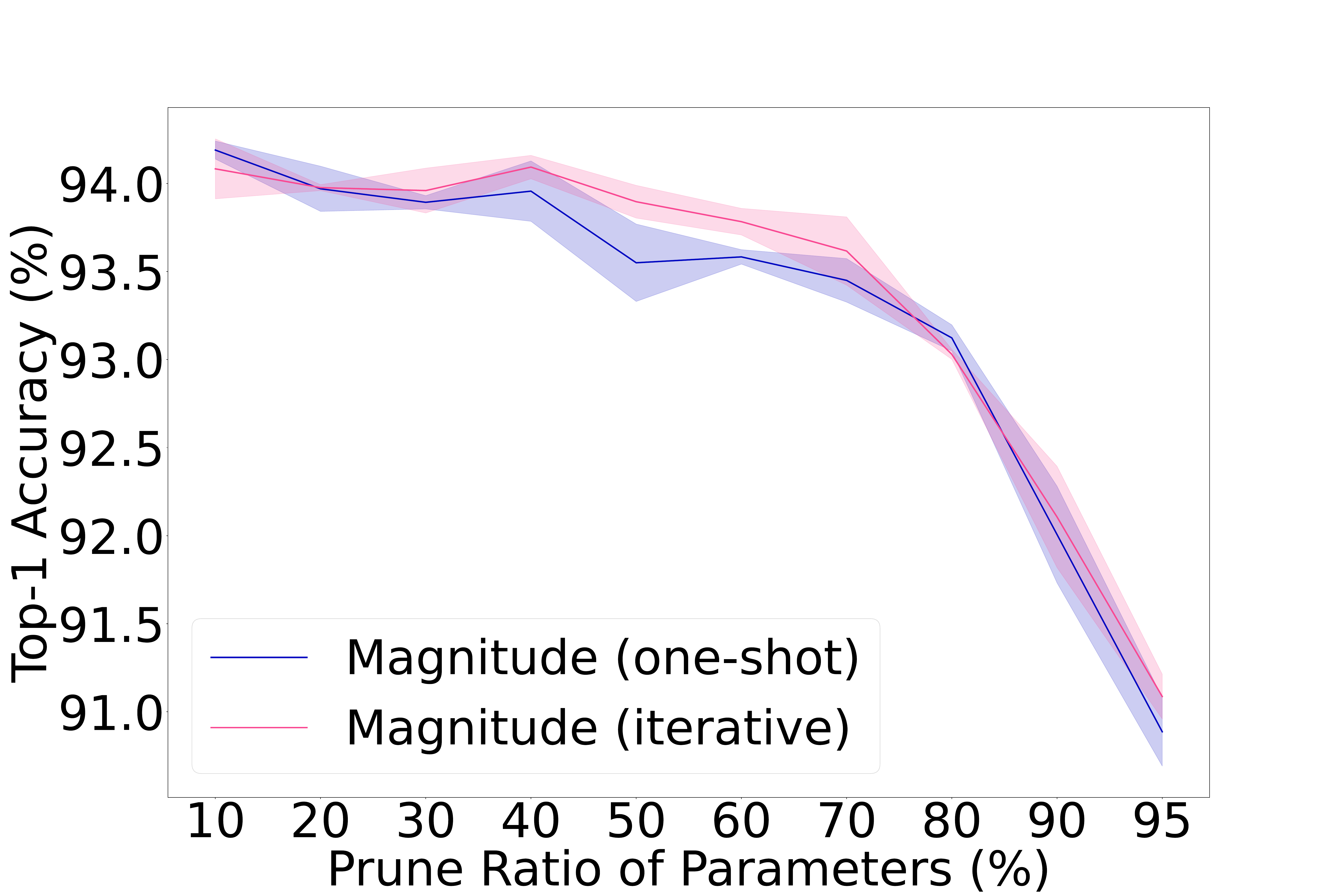}
     \vspace{-0.2cm} 
     \label{Fig:one-shot-iterative-pruning-b}
  \end{minipage} 
  } \\
  \vspace{-0.3cm}
  \subfloat[LLaMA-7B on PTB]{  
  \begin{minipage}{4cm}
  \includegraphics[width=4cm,height=2.5cm]{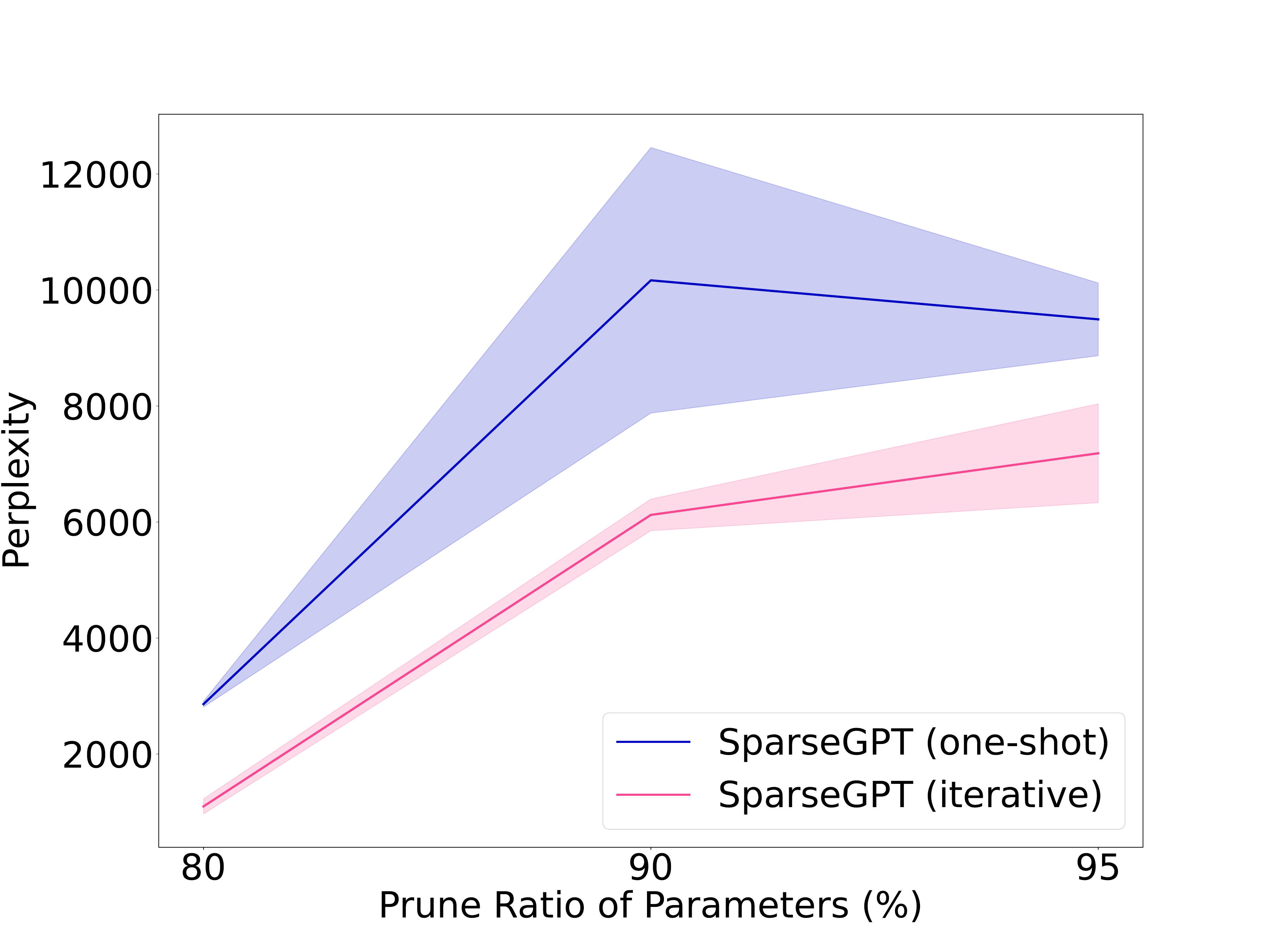}
     \vspace{-0.2cm} 
     \label{Fig:one-shot-iterative-pruning-c}
  \end{minipage} 
  }
 \subfloat[OPT-1.3B on WikiText2]{  
  \begin{minipage}{4cm}
  \includegraphics[width=4cm,height=2.5cm]{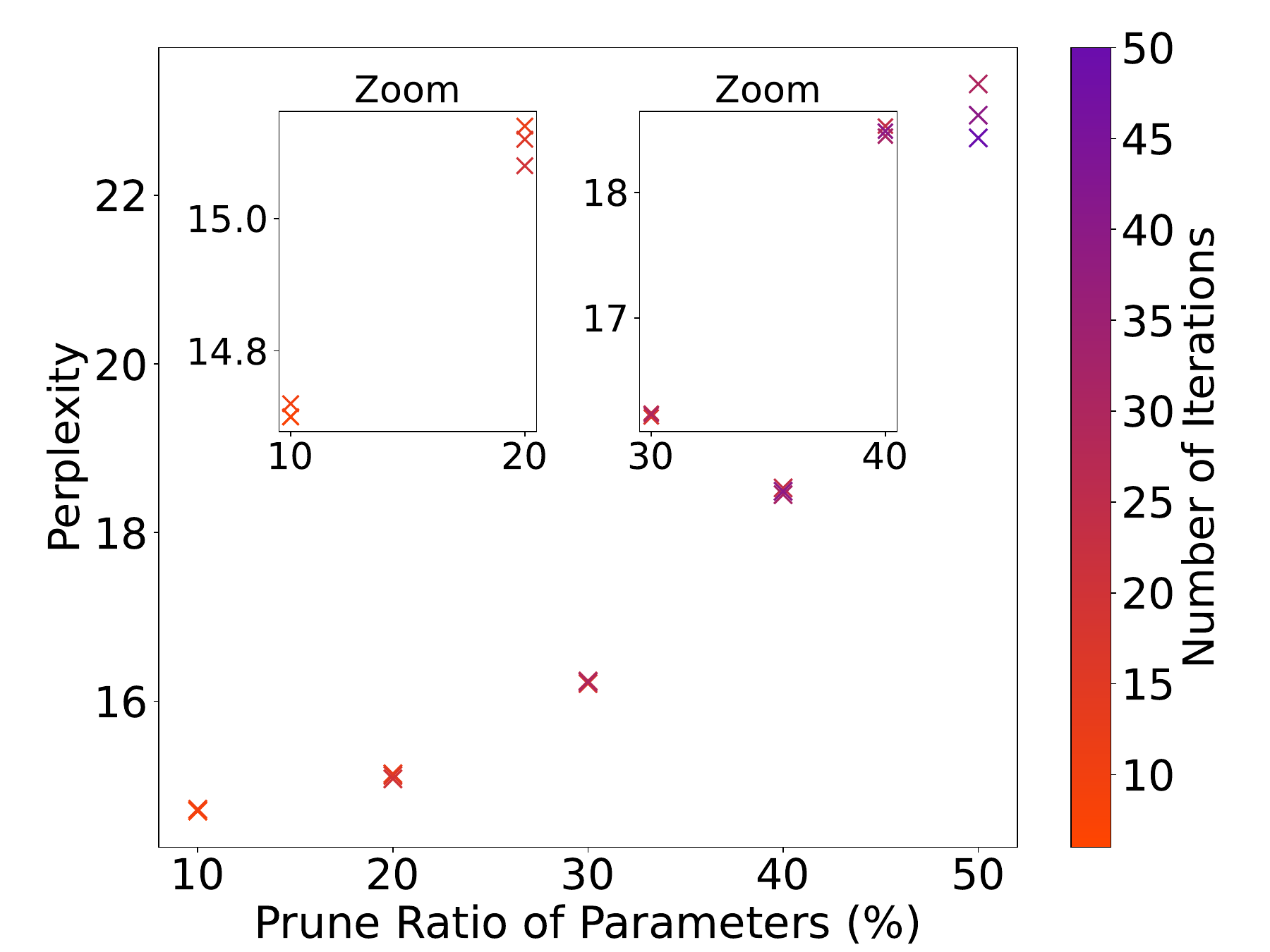}
     \vspace{-0.2cm} 
     \label{Fig:one-shot-iterative-pruning-d}
  \end{minipage} 
  } 
 \caption{One-shot vs. iterative pruning. Shaded regions indicate standard deviation based on three independent runs. Best view in color and zoom in.}
 \label{Fig:one-shot-iterative-pruning}
\vspace{-0.4cm}
\end{figure}

\citet{lin2020dynamic} analyze the difference between one-shot and iterative pruning methods from the perspective of stochastic gradient. Their results show that the iterative pruning method computes a stochastic gradient at the pruned model and takes a step that best suits the compressed model. In contrast, one-shot pruning computes a stochastic gradient at the original weights and moves towards the best dense model. The work in \cite{ouderaa2024llm} theoretically supports iterative pruning by arguing that the surrogate loss landscape, based on a Taylor expansion, only holds locally, making it unreliable for larger weight changes.

We prune VGG-16 \cite{simonyan2015very} and ResNet-32 \cite{he2016deep} on CIFAR-10 \cite{krizhevsky2009learning} using SynFlow \cite{tanaka2020pruning} and Magnitude-based pruning, and LLaMA-7B \cite{touvron2023llama} on PTB \cite{marcus1993building} using SparseGPT \cite{frantar2023sparsegpt}, applying one-shot and iterative pruning, respectively. Additionally, we compare the pruning results of OPT-1.3B \cite{zhang2022opt} on WikiText2 \cite{merity2016pointer} under different iteration settings, using experimental results from \cite{ouderaa2024llm}. 
As illustrated in Fig.~\ref{Fig:one-shot-iterative-pruning} (a) - (c), iterative pruning generally performs better than that of the corresponding one-shot pruning and Fig.~\ref{Fig:one-shot-iterative-pruning} (d) indicates that more iterations tend to yield better performance. 

\subsection{Data-free vs. Data-driven Pruning}
\label{datafreedriven}
Pruning methods can be categorized into two types based on whether data is used during the pruning phase: data-free and data-driven. Data is generally believed to be essential for finding good subnetworks. Most existing pruning works (\cite{liu2021group,nonnenmacher2022sosp,lee2019snip,ma2023llmpruner,shi2023upop}) belong to data-driven methods, and only a few methods (\cite{tanaka2020pruning,su2020sanity,li2019exploiting}) are data-free. 
We apply PBT methods, including three data-free (Random, Magnitude, and SynFlow \cite{tanaka2020pruning}) and two data-driven (SNIP \cite{lee2019snip} and GraSP \cite{wang2020picking}) methods, to prune VGG-16 and ResNet-32/50 on CIFAR-10/100 and ImageNet \cite{russakovsky2015imagenet}, respectively. Table~\ref{Tab:data-free-data-driven-cnn} shows that SynFlow and SNIP are similarly effective, with SynFlow significantly outperforming GraSP, indicating that the effectiveness of PBT methods is not strictly dependent on data usage. In addition, we utilize PAT methods, selecting Random and L1/L2-norm as data-free methods, and data-driven approaches such as Wanda \cite{sun2024simple}, LLM-Pruner \cite{ma2023llmpruner}, and LoRAPruner \cite{zhang2023loraprune} to prune LLaMA-7B. In contrast, Table~\ref{Tab:data-free-data-driven-llm} consistently demonstrates that data-driven PAT methods typically outperform data-free PAT methods. 

\begin{table}[t]
  \caption{Top-1 accuracy (\%) of data-free (first three) and data-driven (last two) pruning methods. Original Top-1 accuracy on CIFAR-10, CIFAR-100, and ImageNet are 93.76\%, 73.18\%, and 76.20\%, respectively. An asterisk (*) denotes a result from our reproduction; others are from \cite{frankle2021pruning}.}
  \vspace{-0.1cm}
  \centering
  \label{Tab:data-free-data-driven-cnn}
  \scalebox{0.95}{%
  \begin{tabular}{l|c|c|c|c|c|c}
    \Xhline{0.3ex}      
    Model & \multicolumn{2}{c|}{VGG-16*} & \multicolumn{2}{c|}{ResNet-32*} & \multicolumn{2}{c}{ResNet-50} \\
     \cline{2-7}
    Dataset & \multicolumn{2}{c|}{CIFAR-10} & \multicolumn{2}{c|}{CIFAR-100} & \multicolumn{2}{c}{ImageNet} \\
     \hline
     Method \textbackslash Ratio (\%)& 50.00 & 90.00 & 50.00 & 90.00 & 73.80 & 89.30 \\
    \hline 
     Random  & 93.12 & 90.68 & 72.10 & 67.29 & 71.20 & 65.20\\
     Magnitude  & 93.32 & 92.78 & 71.86 & \textbf{68.77} & 72.50 & 66.50 \\
     SynFlow \cite{tanaka2020pruning} & \textbf{93.74} & \underline{93.26} & \textbf{72.15} & 68.44 & \underline{72.60} & \textbf{68.00} \\ 
     \hline
     SNIP \cite{lee2019snip} & \underline{93.70} & \textbf{93.40} & 70.87 & 68.00 & \textbf{72.70} & 66.60 \\ 
     GraSP \cite{wang2020picking} & 92.97 & 92.51 & \underline{72.13} & \underline{68.56} & 72.10 & \underline{67.20}  \\
    \Xhline{0.3ex}      
  \end{tabular}
  }
  \vspace{-0.1cm}
\end{table}

\begin{table}[t]
   \caption{Zero-shot accuracy (\%) of data-free (first three) and data-driven (last three) pruning methods on LLaMA-7B~\cite{touvron2023llama} with common sense datasets. ``Avg.'' is calculated among four datasets. An asterisk (*) denotes a result obtained from our reproduction; others are from \cite{zhang2023loraprune}.}
    \label{Tab:data-free-data-driven-llm}
    \vspace{-0.1cm}
    \centering
    \small
    \scalebox{0.70}{%
    \begin{tabular}{l|c|ccccc}
    \Xhline{0.3ex} 
    Ratio (\%) & Method & BoolQ & PIQA & HellaSwag & WinoGrande & Avg. $\uparrow$ \\ 
    \hline
    0 & Unpruned & 73.18 & 78.35 & 72.99 & 67.01 & 72.88 \\
    \hline
    20.00 & 
    Random* & 55.57 & 73.39 & 64.49 & 60.38 & 63.46 \\
    w/ tune & L1-norm* & 58.47 & 75.35 & 65.40 & 60.93 & 65.04 \\
     & L2-norm* & 65.02 & 75.14 & 65.07 & 62.12 & 66.84\\
    \cline{2-7}
    & Wanda~\cite{sun2024simple} & \textbf{65.75} & 74.70 & 64.52 & 59.35 & 66.08 \\
    & LLM-Pruner~\cite{ma2023llmpruner} & 64.62 & \underline{77.20} & \underline{68.80} & \textbf{63.14} & \underline{68.44}\\
    & LoRAPruner~\cite{zhang2023loraprune} & \underline{65.62} & \textbf{79.31} & \textbf{70.00} & \underline{62.76} & \textbf{69.42}\\
    \hline
    50.00 & 
    Random* & \underline{61.04} & 62.30 & 40.37 & 53.51 & 54.31 \\
    w/ tune & L1-norm* & 40.73 & 66.32 & 42.66 & 51.85 & 50.39 \\
    & L2-norm* & 38.50 & 67.08 & 43.47 & 52.80 & 50.46 \\
    \cline{2-7} 
    & Wanda~\cite{sun2024simple} & 50.90 & 57.38 & 38.12 & \textbf{55.98} & 50.60 \\
    & LLM-Pruner~\cite{ma2023llmpruner} & 60.28 & \underline{69.31} & \underline{47.06} & 53.43 & \underline{57.52}\\
    & LoRAPruner~\cite{zhang2023loraprune} & \textbf{61.88} & \textbf{71.53} & \textbf{47.86} & \underline{55.01} & \textbf{59.07}\\
    \Xhline{0.3ex} 
    \end{tabular}
    }
 \vspace{-0.4cm}   
\end{table}

\subsection{Pruning on Initialized vs. Pre-trained Weights}
\begin{figure}[t]
\centering
 \subfloat[ResNet-152 on CIFAR-100]{  
  \begin{minipage}{4cm}
       \includegraphics[width=4cm,height=2.5cm]{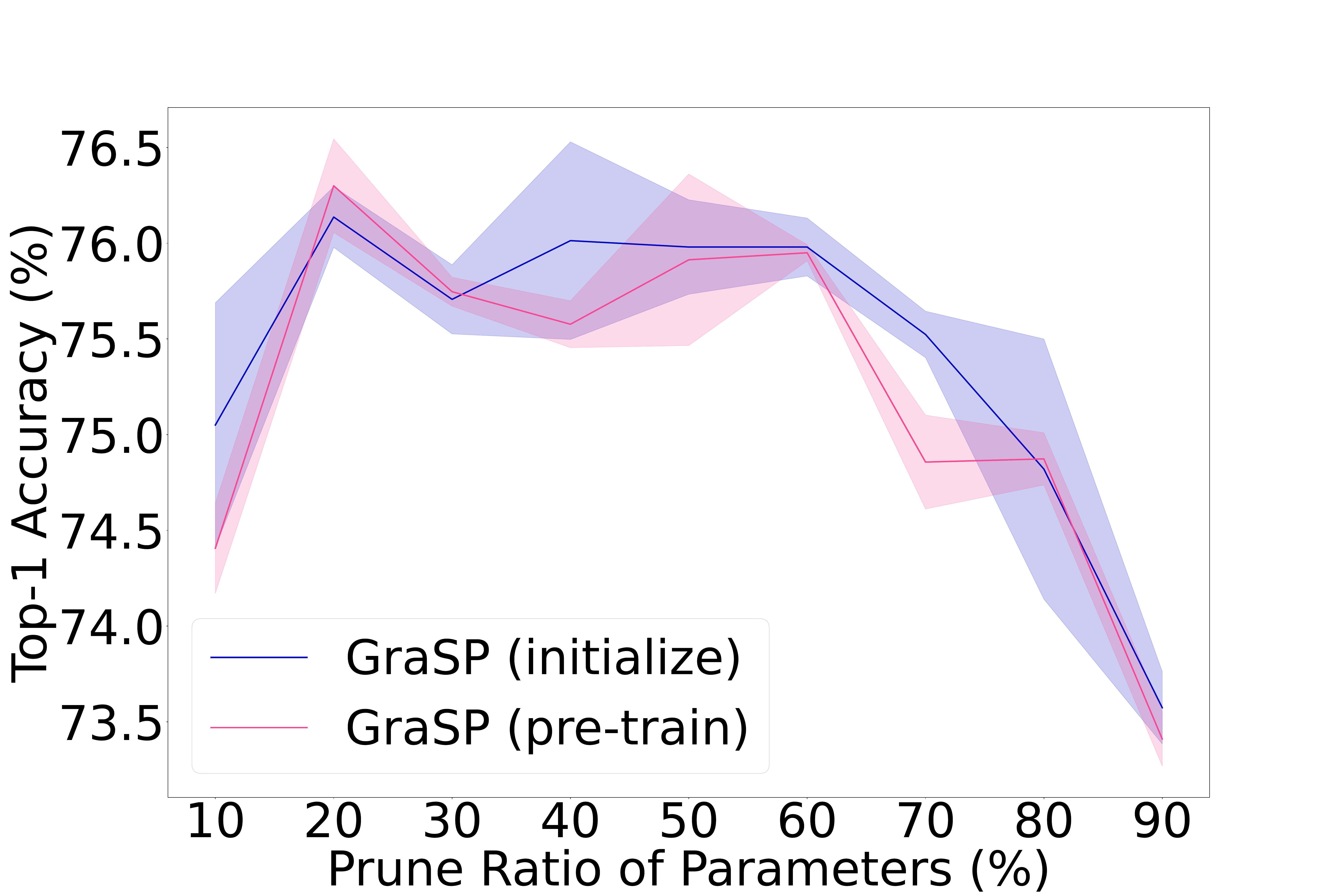}
       \vspace{-0.2cm}
     \label{Fig:initialized-pretrained-weights-a}
  \end{minipage} 
  }
  \subfloat[DeiT-Tiny on CIFAR-10]{  
  \begin{minipage}{4cm}
       \includegraphics[width=4cm,height=2.5cm]{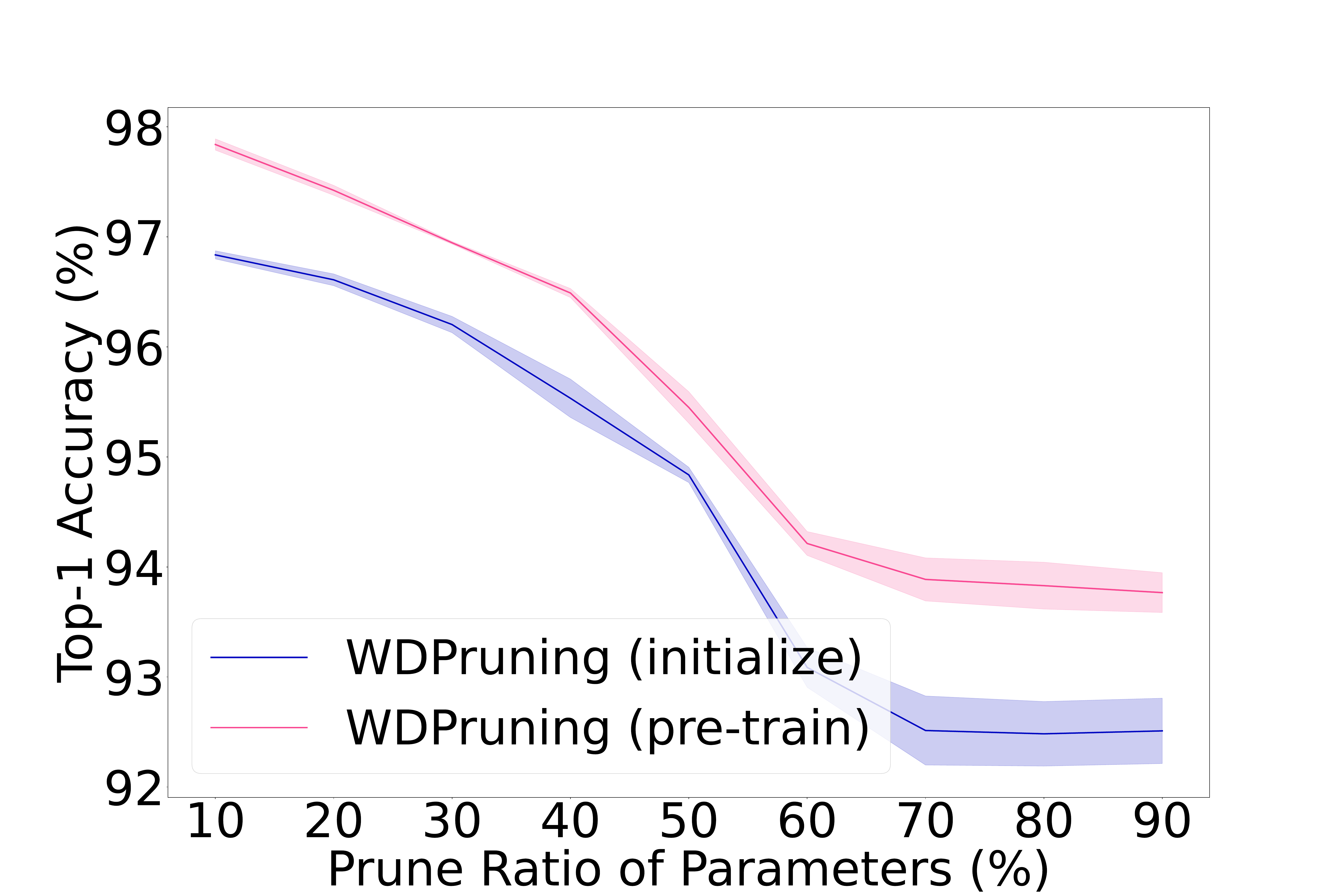}
      \vspace{-0.2cm} 
     \label{Fig:initialized-pretrained-weights-b}
  \end{minipage} 
  }
 \caption{Pruning using randomly initialized vs. pre-trained weights. Shaded regions indicate standard deviation based on three independent runs (best view in color).}
 \label{Fig:initialized-pretrained-weights}
\vspace{-0.8cm}
\end{figure}

\citet{frankle2021pruning} find the subnetworks in CNNs obtained by pruning on randomly initialized weights (such as SNIP \cite{lee2019snip}, GraSP \cite{wang2020picking}, SynFlow \cite{tanaka2020pruning}) are robust to ablation treatments (i.e., randomly shuffling the mask positions within each layer or reinitializing weights while keeping masks unchanged). To find the reason behind such immunity, \citet{singh2021why} use the Wasserstein distance to measure distributional similarity and find that the remaining weights' distribution changes minimally with these ablations, which helps maintain similar performances. In contrast, \citet{su2020sanity} find the subnetworks in CNNs achieved by pruning on pre-trained weights, such as LTH \cite{frankle2019lottery}, are sensitive to these ablations. \citet{qiu2020train} claim that training weights can be decoupled into two dimensions: the locations of weights and their exact values, with the locations of weights holding most of the information encoded by the training. \citet{wolfe2022how} theoretically analyze the impact of pre-training on the performance of a pruned subnetwork in CNNs obtained by greedy forward selection and find that the number of pre-training iterations increases logarithmically with the dataset size. Unlike CNNs, Transformers typically need appropriate self-supervised pre-training to perform well \cite{amos2024never}. Therefore, pruning of Transformers generally targets pre-trained models. We conduct experiments to explore whether pre-trained weights facilitate achieving better subnetworks. Fig.~\ref{Fig:initialized-pretrained-weights} (a) indicates that for the PBT method, GraSP \cite{wang2020picking}, pruning with pre-trained weights does not necessarily improve Top-1 accuracy. However, Fig.~\ref{Fig:initialized-pretrained-weights} (b) shows that for the PAT method, WDPruning \cite{yu2022width}, pre-training may be crucial for obtaining subnetworks with better performance.

\subsection{Global vs. Local Pruning}
The difference between global (\cite{chin2020towards,he2018amc,tiwari2021ChipNet,liu2021group,ouderaa2024llm}) and local (\cite{wang2021neural,wang2023trainability,ding2018auto,luo2017thinet,ma2023llmpruner}) pruning lies in whether structures are removed from a subset or all available structures of a network. 
A major limitation of local pruning is that setting a pre-defined prune ratio for each layer can be complex and lead to sub-optimal sparsity. To simplify, local pruning often uses a consistent prune ratio across layers. In contrast, global pruning automatically generates a varying prune ratio for each layer. However, global pruning poses great challenges, particularly for LLMs, due to significant variations in layer magnitudes. For instance, some outlier features may have magnitudes up to 20 times larger than others \cite{kovaleva2021bert}, leading to incomparability issues. \citet{ma2023llmpruner} notes a marginal advantage of local pruning compared to global pruning in LLMs. Although previous pruning methods for LLMs (\cite{ma2023llmpruner,zhang2023loraprune,frantar2023sparsegpt,sun2024simple}) are primarily local, some global methods (\cite{ouderaa2024llm,bai2024sparsellm,an2024fluctuationbased}) start to emerge. For example, \citet{bai2024sparsellm} argue that local pruning excessively restricts the alignment of input and output in all the intermediate layers, leading to a sub-optimal solution, and propose a global pruning method called SparseLLM to address the drawbacks.

\subsection{Training from Scratch vs. Fine-tuning}
\begin{figure}[t]
\centering
 \subfloat[ResNet-152 on CIFAR-100]{  
  \begin{minipage}{4cm}   \includegraphics[width=4cm,height=2.5cm]{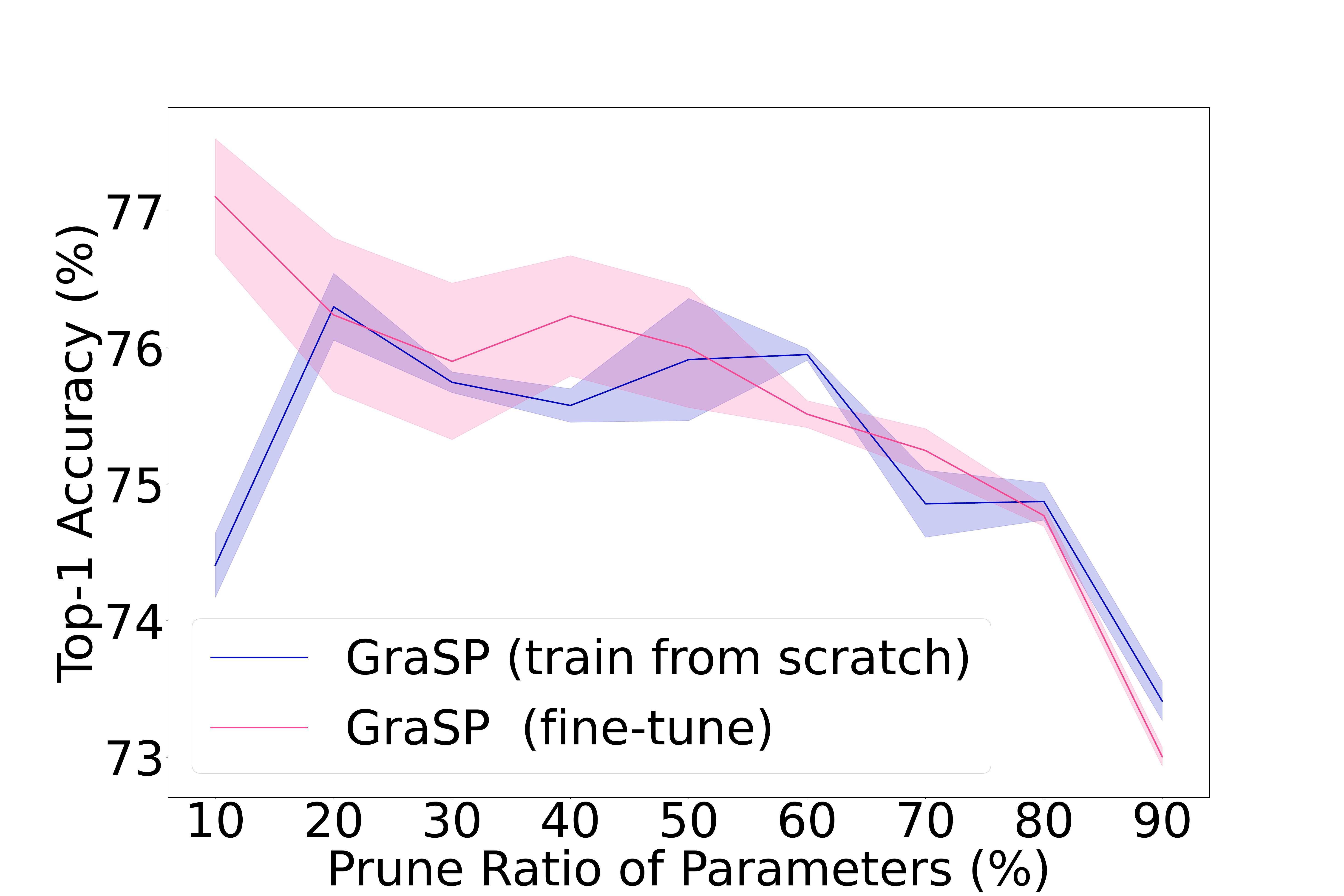}
  \vspace{-0.2cm}
     \label{Fig:Train-from-scratch-finetune-a}
  \end{minipage} 
  }
  \subfloat[DeiT-Tiny on ImageNet]{  
  \begin{minipage}{4cm}
       \includegraphics[width=4cm,height=2.5cm]{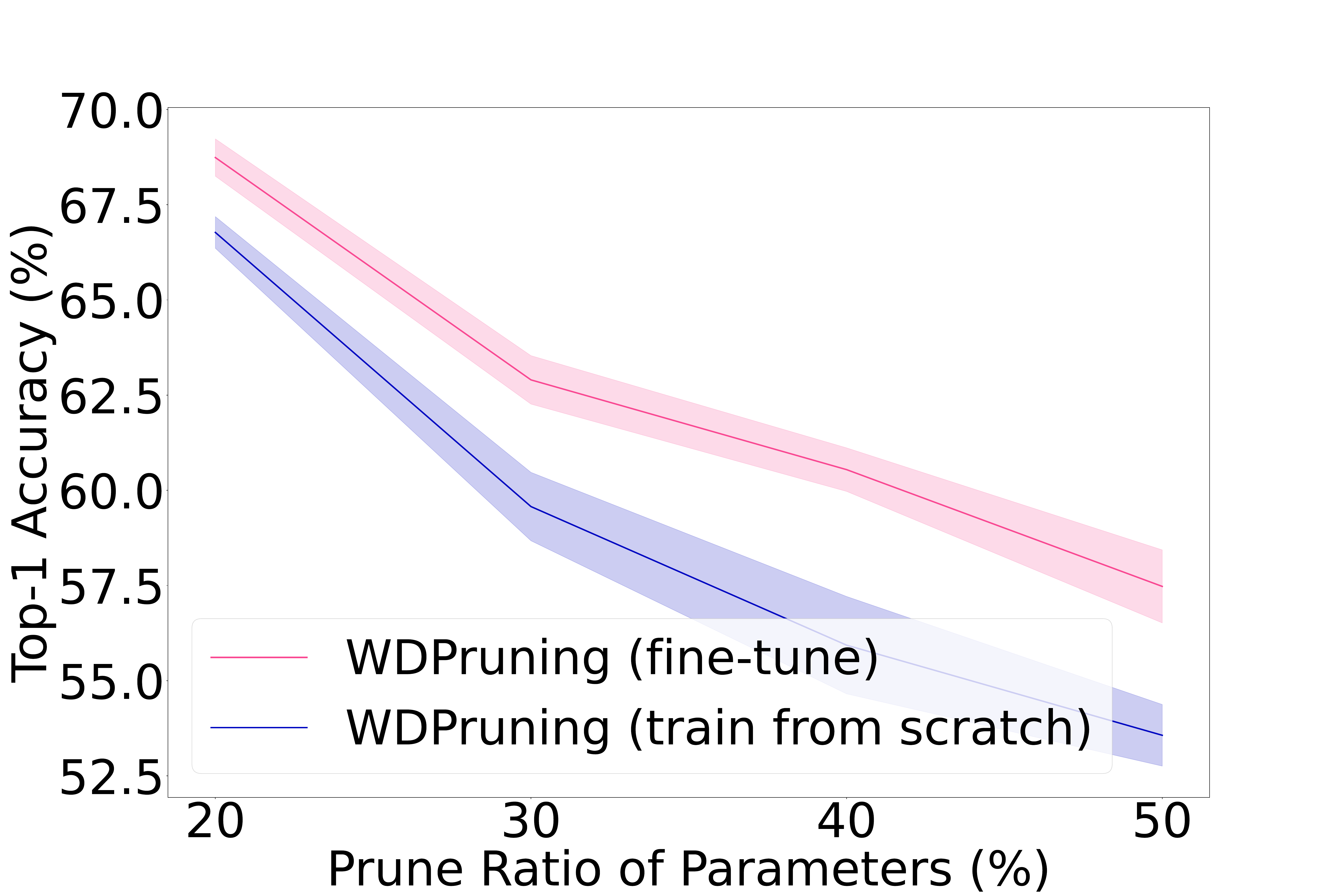}
     \vspace{-0.2cm} 
     \label{Fig:Train-from-scratch-finetune-b}
  \end{minipage} 
  }
 \caption{Training from scratch vs. fine-tuning. Shaded regions indicate standard deviation based on three independent runs (best view in color).}
 \label{Fig:Train-from-scratch-finetune}
\vspace{-0.4cm}
\end{figure}

After pruning, many pruning methods require training the subnetwork for several epochs to regain performance. \citet{le2021network} argue that retraining is essential to recover loss accuracy in pruning. Generally, retraining can be divided into two types: training from scratch or fine-tuning. There has been debate over whether fine-tuning is more effective than training from scratch in recovering accuracy. On the one hand, \citet{liu2019rethinking} find that for ResNet, VGG, and other standard structures on ImageNet, training the subnetworks with new random initialization can achieve better performance than fine-tuning them. On the other hand, \citet{li2017pruning} observe that training a subnetwork from scratch performs worse than fine-tuning it. \citet{liu2021lottery} investigate pruning ResNet20 on CIFAR-10 with ADMM-based \cite{zhang2018systematic} one-shot pruning method and find that pruning \& fine-tuning outperforms LTH (pruning \& training from scratch) over various prune ratios. Additionally, the results in \cite{gao2021network,ye2020good} show that fine-tuning is necessary for better performance on sparse mobile networks than training from scratch. The results in \cite{fang2023structural} reveal that training from scratch requires more steps to achieve convergence, suggesting that starting pruned models from scratch may not be the most cost-effective strategy, given its training cost is comparable to that of pre-trained models. In recent years, some compromise methods (such as weight rewinding \cite{frankle2020linear}) have been proposed. The results in \cite{frankle2020linear,renda2020comparing} show that weight rewinding can achieve higher accuracy than fine-tuning. We prune ResNet-152 and DeiT-Tiny \cite{touvron2021training} on CIFAR-100 and ImageNet with pre-trained weights, respectively. Then we fine-tune the pruned networks or train them from scratch. Fig.~\ref{Fig:Train-from-scratch-finetune} indicates that fine-tuning generally outperforms training from scratch. Notably, on ImageNet, fine-tuning achieves significantly higher accuracy than training from scratch.

\subsection{Original Task vs. Transfer Pruning}
In recent literature, pruning is combined with transfer learning \cite{pan2009survey} that can dramatically improve accuracy and speed up convergence. For ease of distinction, in this survey, original task pruning denotes the pruning pipeline that directly performs on the target task. In contrast, transfer pruning is performed on the source task and then transfers the subnetwork to the target task. Specifically, transfer pruning is divided into two types: dataset transfer and architecture transfer. The former prunes networks on the source dataset and transfers the subnetwork to the target dataset, while the latter prunes on one architecture and transfers the subnetwork to another. 

Some works (\cite{tiwari2021ChipNet,chen2020lottery,chen2021lottery,mehta2019sparse}) study the transferability of sparsity masks on datasets. 
\citet{morcos2019one} observe that for image classification, winning tickets generated on larger datasets (such as with larger training set size and/or more classes) consistently transfer better than those generated with smaller datasets. For example, winning tickets generated on ImageNet and Places365 demonstrate better performance across other smaller target datasets such as CIFAR-10 and CIFAR-100. \citet{iofinova2022how} present a pioneering study of the transfer performance of subnetworks and find that pruning methods with similar Top-1 accuracy on ImageNet \cite{russakovsky2015imagenet} can have surprisingly different Top-1 accuracy when used for transfer learning. For pruning on LLMs, performing pruning directly on the target task typically yields better performance than zero-shot pruning \cite{ouderaa2024llm}. For architecture transfer pruning, Elastic Ticket Transformations (\textbf{ETTs}) \cite{chen2021elastic} transforms winning tickets in one kind of network (such as ResNet \cite{he2016deep}) to another deeper or shallower one from the same model family. 

\subsection{Static vs. Dynamic Pruning}
Static pruning \cite{dong2017more} uses static pruning criteria and removes components. In contrast, dynamic pruning (\cite{tang2021manifold,gao2019dynamic,hua2019channel,li2021dynamic}) exploits input-specific pruning criteria, preserves the entire network structures, and accelerates the networks by dynamically skipping unimportant components. However, dynamic pruning generally does not perform run-time fine-tuning or retraining. The difference between static and dynamic pruning is mainly reflected in the pruning criteria and the pruned model. The advantages and disadvantages of static and dynamic pruning are shown in Table~\ref{Table6-comparison-static-dynamic}.

\subsection{Layer-wise Weight Density Analysis}
Some works (\cite{ma2021sanity,sanh2020movement,liu2021group,zhang2022graph}) study the distributions of layer-wise weight density in a subnetwork, showing that different layers can have very different weight densities. The differences arise from the joint action of networks' structural characteristics and pruning methods. \citet{zhang2022are} empirically divide the layers into either ``ambient'' or ``critical''. Ambient layers are not sensitive to weight changes, while critical layers are. Thus, ambient layers should be heavily pruned, resulting in lower weight densities. 

Pruning methods can also result in different weight densities. \citet{gong2024fast} categorize the sparsity allocation formed by pruning methods into uniform and non-uniform. Uniform sparsity methods (\cite{frantar2023sparsegpt,sun2024simple,ma2023llmpruner}) allocate the same prune ratio for all the layers, whereas non-uniform methods (\cite{yin2024outlier,an2024fluctuationbased,ouderaa2024llm,bai2024sparsellm}) assign varying sparsity rates to different layers. For example, in CNNs, some pruning methods tend to assign more weights to earlier layers than to later ones. \citet{ma2021sanity} investigate the layer-wise keep-ratios of subnetworks obtained by GraSP \cite{wang2020picking}, SNIP \cite{lee2019snip}, and LTH \cite{frankle2019lottery} on VGG and ResNet, observing a common trend of declining layer-wise keep-ratios, except for some special layers (such as the downsampling layers in ResNet). In contrast, \citet{liu2021group} find their pruned networks keep higher percentages of channels in the deeper layers than those in the lower layers for image classification. 
For pruning Transformers, the results in \cite{sanh2020movement} show that global magnitude pruning tends to prune Transformer layers uniformly, while global first-order methods heavily prune the deeper layers.
\citet{yang2023global} discover a unique less-more-less distribution among stacked ViT blocks. LLM surgeon \cite{ouderaa2024llm} prunes relatively more in the first layers and less in the middle layers. 
Some works (\cite{men2024shortgpt,yuan2024lift,kim2024shortened,yang2024laco}) reveal that some layers are not essential and can be entirely removed or merged.

\begin{table}[t]
  \caption{Advantages and disadvantages of static and dynamic pruning.}
  \label{Table6-comparison-static-dynamic}
  \vspace{-0.1cm}
  \begin{tabular}{l|cc}
    \Xhline{0.3ex}  
     Type & Advantages & Disadvantages\\
    \hline
    Static Pruning &  model size reduced & fixed subnetwork\\
    \hline
    Dynamic Pruning & flexible subnetwork & model size unreduced \\
   \Xhline{0.3ex}  
\end{tabular}
\vspace{-0.4cm}
\end{table}

\subsection{Pruning with Different Levels of Supervision}
In descending order of supervision level during neural network pruning, pruning can be divided into supervised, semi-supervised, self-supervised, and unsupervised pruning \cite{kalyan2018unsupervised}. Self-supervised learning can be divided into two classes: generative and contrastive learning \cite{liu2021self}. Similar to supervised learning, supervised pruning works on fully labeled datasets, and most current pruning methods fall into this category. However, supervised pruning suffers from similar bottlenecks as supervised learning, such as the expensive manual labeling. As a promising alternative, semi-supervised, self-supervised, and unsupervised pruning have drawn massive attention. \footnote{The differences between supervised, semi-supervised, unsupervised, self-supervised learning refer to \cite{liu2021self}.} 
 
For example, \citet{caron2020pruning} observe different results in self-supervision pruning compared to supervision pruning, where winning tickets initialization only introduces a slight performance improvement compared to random re-initialization. \citet{contrastivepruning} claim that unsupervised pruning usually fails to preserve the accuracy of the original model. Notably, label supervision for network pruning and training can be independent. For example, \citet{chen2021lottery} use supervised pruning method \textbf{IMP} (Iterative Magnitude Pruning) to explore the subnetworks of self-supervised pre-trained models (simCLR \cite{chen2020simple} and MoCo \cite{chen2020improved}) on ImageNet. Similarly, \citet{lai2021parp} exploit the supervised pruning method IMP to prune self-supervised speech recognition models.

\section{Fusion of Pruning and other Compression Techniques}
\label{fusion}
In this section, we review the fusion of neural network pruning with other network compression techniques, such as quantization \cite{ma2023llmpruner}, tensor decomposition \cite{li2021towards}, knowledge distillation \cite{hinton2015distilling}, and network architecture search \cite{li2021npas}. On the one hand, fusion provides more choices for network compression. On the other hand, combined compression techniques can complement each other to further improve the performance and prune ratio.  

\textbf{Pruning \& Quantization:}
Quantization \cite{tung2018clipq} is a compression technique that reduces the number of bits used to represent the network weights and/or activations, significantly reducing the model size and memory footprint with only a minor performance drop. To obtain more compact models and achieve model acceleration, \citet{han2015deep} pioneer pruning the redundant network connections and quantizing the weights. CLIP-Q \cite{tung2018clipq} jointly performs pruning and quantization during the fine-tuning stage. MPTs \cite{diffenderfer2021multi} integrates pruning and quantizing randomly weighted full-precision neural networks to obtain binary weights and/or activations. EB \cite{you2020drawing} applies 8-bit low-precision training to the stage of searching EB tickets. 

\textbf{Pruning \& Tensor Decomposition:}
Tensor decomposition \cite{lin2018holistic} decomposes convolutions into a sequence of tensors with fewer parameters. In contrast to pruning, it explores the original weights' low-rank structure, while keeping the dimension of the convolutional output unchanged. CC \cite{li2021towards} combines channel pruning and tensor decomposition to compress CNN models by simultaneously learning model sparsity and low rankness. Hinge \cite{li2020hinge} introduces group sparsity to fuse filter pruning and decomposition under the same formulation. \citet{li2023losparse} propose LoSparse to prune Transformers by combining low-rank approximations and pruning.

\textbf{Pruning \& NAS:}
Neural Architecture Search (\textbf{NAS}) provides a mechanism to automatically discover the best architecture for the problem of interest, offering a new approach for pruning to find suitable network depth and width. For example, for CNNs, NPAS \cite{li2021npas} performs a compiler-aware joint network pruning and NAS, determining the filter type (different kernel sizes), the pruning scheme, and the rate for each layer. 
TAS \cite{2019dongnetwork} exploits NAS to search for the depth and width of a network to obtain pruned networks and uses knowledge distillation to train these pruned networks. \citet{klein2023structural} explore structured pruning of fine-tuned Transformers via NAS.

\textbf{Pruning \& Knowledge Distillation:} Knowledge Distillation (\textbf{KD}) \cite{hinton2015distilling} guides the student to effectively inherit knowledge from the teacher and mimic the teacher's output. Some works (\cite{liu2021content,park2022prune}) exploit pruning before KD to boost KD's quality. For example, 
\citet{liu2021content} prune unimportant channels to the contents of interest and focus the distillation on the interest regions. \citet{park2022prune} prune the teacher network first to make it more transferable and then distill it to the student. Some works (\cite{hua2019channel,chen2022knowledge, chen2021long,zhang2022advancing}) use KD to train the pruned networks. The results in \cite{chen2022knowledge} show that the pruned network recovered by KD performs better than it regained by fine-tuning. \citet{zou2022dreaming} propose a data-free deraining model compression method that distills the pruned model to fit the pre-trained model. \cite{wang2023large} introduce a multi-stage compression strategy, AntGMM, to compress large multimodal models by utilizing structured pruning and knowledge distillation.

\textbf{Pruning \& Multi-compression Techniques:}
Some works (\cite{mao2020ladabert,yao2021DetNAS,wang2020gan}) explore the fusion of pruning with more than one compression technique. For example, GS \cite{wang2020gan} combines pruning, quantization, and KD for GANs compression. Joint-DetNAS \cite{yao2021DetNAS} integrates pruning, NAS, and KD for image translation. LadaBERT \cite{mao2020ladabert} merges pruning, matrix factorization, and KD to compress BERTs \cite{michel2019sixteen} for natural language understanding.   

\section{Suggestions and Future Directions}
\label{recommendations}
In this section, we discuss how to choose different pruning methods and provide promising directions for future work.
\subsection{Recommendations on pruning method selection}
After years of research and exploration, there are many off-the-shelf pruning methods. However, no golden standard exists to determine which one is the best. Different suitable pruning methods exist to compact deep neural networks for specific application requirements and hardware/software resources. Here are some general recommendations for choosing an appropriate pruning method. 

(1) If you do not have special hardware (e.g., FPGAs or ASICs) or software (such as sparsity convolutional libraries) but need actual neural network acceleration and compression, structured pruning is more suitable than unstructured pruning because most software frameworks and hardware cannot accelerate sparse matrices' computation. 

(2) If you have sufficient computational resources during the pruning stage, consider using iterative PAT methods that can typically minimize the impact on performance under the same prune ratio. On the other hand, if you have limited computational resources during both the pruning and inference stages, consider using one-shot PBT or one-shot post-training pruning methods, particularly for LLMs.  

(3) If you have enough labeled examples on the target task, consider using supervised pruning methods. However, if only a few examples on the target task are labeled, semi-supervised or transfer pruning methods may be considered. If the examples on the target task are not labeled, consider self-supervised, unsupervised, or transfer pruning methods.

(4) If you have a sufficient memory footprint during pruning for NLP tasks, consider heavily compressing large models rather than lightly compressing smaller models to meet the same budgets. Some results in \cite{li2020train} show that for NLP tasks finding pruned models derived from larger dense networks outperform small dense networks of comparable size to pruned models.

(5) If you have enough memory to store the dense neural network during the inference stage and wish to provide run-time flexible computational cost allocation for different inputs, dynamic pruning methods can be considered where inputs with smaller shapes can allocate less computational cost to perform the task and if the input shape is bigger more computational cost can be allocated. 

(6) If you need to trim down neural networks in multiple dimensions, you can comprehensively consider layerwise pruning (decreasing the model's depth), channel pruning (reducing the model's width), and image resolution pruning (scaling down the model's input resolution) or token pruning (selectively removing tokens from text data). In addition, pruning can be integrated with quantization to further reduce the memory footprint and the neural networks' size.

(7) If you want to achieve a better tradeoff between speed and accuracy, the following settings may help: use a pre-trained model; set an appropriate learning rate (if any) for both the pruning and retraining stages; fine-tune the pruned models for several epochs; integrate pruning with knowledge distillation, NAS, or other compression methods to achieve complementarity; and adversarial training may have some help \cite{gan2022playing}. 

(8) If you need to train a subnetwork to recover performance, \citet{ma2021sanity} show that subnetworks with residual connections achieve higher accuracy using a relatively small learning rate. In contrast, subnetworks without residual connections benefit from a larger learning rate. 

(9) To benefit from large-scale pre-training, adding parameters is more critical than minimizing FLOPs \cite{han2024parameternet}. By incorporating techniques such as dynamic convolution \cite{chen2020dynamic}, models with low FLOPs can increase their capacity without a substantial rise in computational cost, enhancing their performance during extensive pre-training.

\subsection{Future Directions}
We discuss four promising directions for the further development of neural network pruning, namely, (1) theories, (2) techniques, (3) applications, and (4) evaluation. 

\textbf{Theories:}
Despite the existing works, several fundamental questions about pruning still need to be answered. For example, prior works demonstrate that network layers contain irreplaceable information as long as redundant ones. Does a theoretical upper bound of the prune ratio exist for a given network that still maintains the performance of its dense equivalent? In other words, how heavily can a network be pruned theoretically without accuracy loss? It is a tricky question due to the intricate relationships between network layers. Besides, is pruning explainable? A common belief is that deep neural networks are hard to interpret. As such, making pruning explainable is an uphill task. However, the interpretability of pruning is vital for understanding the factors behind pruning (e.g., model structure and weights) and exploring more effective pruning methods.     

\textbf{Techniques:} To obtain better algorithm designs whose architectures are learned in an economical, efficient, and effective manner, it is a trend to extend Automated Machine Learning (\textbf{AutoML}) methods and NAS to pruning. Furthermore, pruning is also beginning to combine with various learning contexts, such as lifelong learning \cite{chen2021long}, continual learning \cite{yan2022learning}, contrast learning \cite{corti2022studying}, and federated learning \cite{jiang2022model}, etc. In addition, the rising energy consumption of networks requires more attention to energy-aware pruning. However, preliminary efforts mainly focus on reducing computation and memory costs, which may not necessarily reduce the most energy consumption. Moreover, incorporating pruning into hardware to help deploy pruned networks is also an emerging trend. For example, \citet{sui2023hardware} propose a hardware-friendly pruning method and deploy the pruned models on an FPGA platform.   

\textbf{Applications:} 
Pruning has begun to draw attention to more complex applications such as visual question answering, natural language understanding, speech recognition, and content generation than image classification. Foundation models such as GPT-4 \cite{openai2023gpt4} might be a possible way to Artificial General Intelligence (\textbf{AGI}). However, its enormous size hinders its application in many downstream tasks. Fig.~\ref{Fig:pruning-taxonomy} highlights content related to large model pruning. In the future, more pruning methods will enable colossal foundation models to benefit from pruning research, making them more compact and efficient \cite{frantar2023sparsegpt}.

\textbf{Evaluation:} With the emergence of many pruning methods, standardized benchmarks, and metrics are required to provide a fair evaluation. Different pruning techniques, network architectures, tasks, and experimental settings lead to incomparable results and make it hard to compare pruning methods fairly \cite{wang2023why}. ShrinkBench \cite{blalock2020what} takes the first step and provides a benchmark of pruning methods for image classification. As pruning is applied to applications beyond image classification, standardized benchmarks and metrics for other applications are needed.

\section{Conclusion}
\label{conclusion}
As an essential compression technique, deep neural network pruning has attracted increasing research attention with the recent emergence of various pruning methods and applications. This survey conducts a comprehensive review on the following four scopes: 1) universal/specific speedup, with a systematic review of unstructured, structured, and semi-structured pruning; 2) when to prune, including pruning before/during/after training for static pruning and run-time pruning; 3) how to prune, including pruning by criteria and by learning; 4) fusion of pruning with other compression techniques, such as KD and NAS. A comprehensive comparative analysis, including eight pairs of contrast settings for pruning, layer-wise weight density, and different supervision levels, can help researchers to efficiently and effectively grasp the characteristics of different pruning methods. In addition, recommendations on pruning method selection and future research directions are highlighted and discussed. To facilitate future research, real-world miscellaneous applications and commonly used resources of datasets, networks, and evaluation metrics in different applications are summarized in Appendix~D. To help researchers and practitioners keep up with the development of pruning technologies, we continue updating the representative research efforts and open-source codes for pruning at
\url{https://github.com/hrcheng1066/awesome-pruning}.


%


\appendices
\section{Terms and Notations}
\label{terms}
This section presents the commonly used terms in pruning literature. It is worth mentioning that some terms (e.g., compression ratio) have different definitions in prior works. In addition, for better readability, we list the notations used in the main text in Table~\ref{Table-notations-this-paper}.

\begin{itemize}
\item{\textbf{Prune Ratio}}: Prune ratio \cite{liu2019rethinking} denotes the percentage of weights (or filters, neurons, etc.) that are removed from the dense network and is the complement of Keep Ratio \cite{ning2020dsa}. In general, it can be determined in two ways: pre-defined or learning-decided.   
\item{\textbf{Compression Ratio}}: Compression ratio in \cite{tanaka2020pruning,renda2020comparing} is defined as the ratio of the original number of weights to the preserved number of weights, but in \cite{qian2021probabilistic} it is defined as the ratio of the preserved number of weights to the original number of weights. For example, if 10\% of the weights are preserved, the compression ratio in \cite{renda2020comparing} is 10, but it is 10\% in \cite{qian2021probabilistic}.
\item{\textbf{Sparsity Ratio}}: Sparsity ratio denotes the portion of zero weights (or channels, filters, neurons, etc.) in networks after pruning \cite{chen2021elastic,he2018amc}. It is equivalent to compression ratio in \cite{qian2021probabilistic}.
\item{\textbf{Speedup Ratio}}:  Speedup ratio is defined as the ratio of the pruned number of FLOPs in \cite{dong2017more}, or MACs in \cite{fang2023depgraph} to the original number of FLOPs or MACs, respectively. In \cite{wang2018exploring}, the speedup ratio is calculated by dividing the pruned number of filters in one layer by the original number of filters in that layer.
\item{\textbf{One-shot Pruning}}: One-shot pruning, also called single-shot pruning in \cite{lee2019snip}, scores only once and then prunes the network to the target prune ratio \cite{frankle2019lottery,frankle2021pruning}. 
\item{\textbf{Iterative Pruning}}: Iterative pruning \cite{han2015learning}, also called greedy pruning or oracle pruning in \cite{he2020learning}, repeatedly performs the score-prune-retrain circle for multiple rounds, and each round is one iteration.
\item{\textbf{Local Pruning}}: Local pruning prunes a network by subdividing all weights (or filter, channels, etc.,) into subsets (e.g., layers) and then removing a percentage of each subset \cite{nonnenmacher2022sosp}.
\item{\textbf{Global Pruning}}: In contrast to local pruning, global pruning removes structures from all available structures of a network until a target prune ratio is reached \cite{nonnenmacher2022sosp}. 
\item{\textbf{Dynamic Pruning}}: Dynamic pruning depends on specific inputs \cite{tang2021manifold}, wherein different subnetworks are generated for each input sample. 
\item{\textbf{Static Pruning}}: In contrast to dynamic pruning, the pruned model is shared by different samples for
static pruning \cite{tang2021manifold}. In other words, the model capacities are fixed for different inputs.   
\item{\textbf{Lottery Ticket Hypothesis}}: Lottery Ticket Hypothesis (\textbf{LTH}) \cite{frankle2019lottery} suggests that a randomly-initialized dense network $f(\mathbf{x};W_{0})$ contains a sparse subnetwork $f(\mathbf{x};W_{0}\odot M)$ which can be trainable with the original weights to achieve competitive performance compared to the original networks.
\item{\textbf{Winning Tickets}}: For a randomly initialized network $f(\mathbf{x};W_{0})$, a winning ticket $f(\mathbf{x};W_{0}\odot M)$ is its subnetwork that once be trained for $T$ epochs (i.e., $f(\mathbf{x};W_{t}\odot M$) will match the performance of the trained network $f(\mathbf{x};W_{t})$ under a non-trivial prune ratio \cite{frankle2019lottery}.
\item{\textbf{Layer Collapse}}: Layer collapse is a phenomenon mentioned in \cite{tanaka2020pruning}, which occurs when all weights in a layer in a network are removed, rendering the network untrainable. \citet{hayou2021robust} provide a formal definition (i.e., ill-conditioned NN) to this problem. 
\item{\textbf{Weight Rewinding}}: Weight rewinding \cite{frankle2020linear} rewinds the weights of the subnetwork to the values in an earlier epoch in training $W_{t}$, where $t<<T$. 
\item{\textbf{Learning Rate Rewinding}}: Learning rate rewinding, proposed in \cite{renda2020comparing}, trains the remaining weights from the final values using the learning rate schedule for a specified number of epochs.
\item{\textbf{FLOPs}}: Float Point Operations (FLOPs) is a commonly used metric (\cite{liu2021group,dong2017more,he2018soft,he2019filter,liu2017learning,zhao2019variational,tiwari2021ChipNet}) to evaluate acceleration of the pruned models theoretically. Some works \cite{he2018soft,dong2017more} introduce the estimation methods to compute FLOPs of neural networks. 
\item{\textbf{MACs}}: In addition to FLOPs, Multiply-Accumulate Operations (MACs) is another popular proxy for evaluating the computational consumption of a network. \citet{nonnenmacher2022sosp} introduce a MACs estimation method.
\item{\textbf{Fine-tuning}}: In the context of pruning, fine-tuning continues to train the preserved weights using the final weight values after pruning \cite{renda2020comparing,blalock2020what}.
\item{\textbf{Training from Scratch}}: Training from scratch is a particular case of weight rewinding, where the weights of the subnetwork are rewinded to their original values $W_{0}$.
\end{itemize}

\begin{table}[t]
  \caption{Notations and descriptions.}
  \label{Table-notations-this-paper}
    \vspace{-0.1cm}
  \centering
    \begin{tabular}{l|c}
    \Xhline{0.3ex}
    Notation & Description \\
    \hline 
    $\mathbf{x}$ & input data or activation/feature\\
    $\mathbf{x}_{i}$ & the $i$-th input data or activation/feature\\
    $\mathbf{y}$ & output data for $\mathbf{x}$ \\
    $\mathbf{y}_{i}$ & the output data for $\mathbf{x}_{i}$ \\
    $\mathcal{D}$ & dataset \\
    $f$ & a network function \\
    $N$ & the number of samples in a dataset \\
    $W$ & the model weights \\
    $w_{i}$ & the $i$-th weight of a model \\
    $w_{ij}$ & a weight that connects input $i$ to output $j$ \\
    $W_{t}$ & the weights after training $t$ epochs \\
    $\mathcal{F}_{i,j}$ & the $j$-th filter of the $i$-th layer \\
    $c_{i}$ & the $i$-th channel of a network \\    
    $\ell$ & standard loss function, e.g., cross-entropy loss\\
    $\mathcal{L}$ & target loss function \\
    $\mathcal{L}_{v}$ & the validation loss function \\
    $L$ & total number of layers in a network \\
    $\odot$ & element-wise multiplication \\
    $\lambda$ & a balance factor\\
    $\boldsymbol{\gamma}$ & a scaling factor vector\\
    $M$ & the masks of weights (or filters, channels, etc.)\\
    $m_{i}$ & the $i$-th mask of $M$\\
    $\mathcal{R}(\cdot)$ & regularization term \\
    $\Delta \mathcal{L}$ & loss change \\
    $c_{out}^{(i)}$ & the number of filters at layer $i$ \\
    $\mathcal{A}$ & the set of every layer's keep ratio \\
    \Xhline{0.3ex}
  \end{tabular}
  \vspace{-0.4cm}
\end{table}

\section{Experimental settings}
\label{experimental-settings}
We use CIFAR-10/100 \cite{krizhevsky2009learning} or ImageNet ILSVRC-2012 \cite{russakovsky2015imagenet} to evaluate Top-1 accuracy of the pruning methods on VGG-16 \cite{simonyan2015very}, ResNet-32/152 \cite{he2016deep}, or DeiT-Tiny \cite{touvron2021training} for image classification. CIFAR-10 and CIFAR-100 datasets contain 50K training and 10K test images for 10 and 100 classes, respectively. ImageNet includes over 1.28 million training and 50K validation images for 1000 classes. For VGG-16 and ResNet-32 on CIFAR-10/100, we conduct the experiments on one NVIDIA A100 GPU (40 GB) using SGD with 0.9 momentum \cite{sutskever2013on}, a weight decay of $10^{-4}$, and train the pruned networks for 160 epochs with a batch size of 128. The initial learning rate is 0.1, reduced at epochs 60 and 120, as in \cite{tanaka2020pruning}. For DeiT-Tiny on ImageNet, we use four NVIDIA A100 GPUs (40 GB) with AdamW (0.9 momentum) and a cosine learning rate decay strategy, training the pruned networks for 100 epochs with a batch size 256 of per GPU and an initial learning rate of $5\times 10^{-4}$, as in \cite{yu2022width}. The code for SNIP \cite{lee2019snip} and GraSP \cite{wang2020picking} can be found at \url{https://github.com/JingtongSu/sanity-checking-pruning}. The code for SynFlow \cite{tanaka2020pruning} and WDPruning \cite{yu2022width} is available at \url{https://github.com/ganguli-lab/Synaptic-Flow} and \url{https://github.com/andyrull/width-and-Depth-pruning-for-Vision-Transformer}, respectively.

In addition, we conduct experiments for LLaMA-7B \cite{touvron2023llama} on one NVIDIA A100 GPU (40 GB) to assess the zero-shot ability of pruned LLMs on WikiText2~\cite{merity2016pointer} and PTB~\cite{marcus1993building} for language generation using perplexity (PPL)\footnote{https://huggingface.co/spaces/evaluate-metric/perplexity}. Besides, we follow LLaMA to implement zero-shot task classification and multiple-choice on four common sense reasoning datasets: BoolQ~\cite{clark2019boolq}, PIQA~\cite{bisk2020piqa}, HellaSwag~\cite{zellers2019hellaswag}, and WinoGrande~\cite{sakaguchi2021winogrande}. The code for LLM-Pruner \cite{ma2023llmpruner}, SparseGPT \cite{frantar2023sparsegpt}, and Wanda \cite{sun2024simple} is available at \url{https://github.com/horseee/LLM-Pruner}, \url{https://github.com/IST-DASLab/sparsegpt}, and \url{https://github.com/locuslab/wanda}, respectively.

For the experiments in Section 7.2 of the main text, we choose a blocksize of 128 for SparseGPT's iterative pruning. In Section 7.4, we use pre-trained DeiT-Tiny on ImageNet as the initialized weights and then retrain DeiT-Tiny on CIFAR-10 to obtain the pre-trained weights. Both a randomly initialized ResNet-152 and a ResNet-152 pre-trained on CIFAR-100 are pruned using GraSP \cite{wang2020picking}. In Section 7.5, ResNet-152 pre-trained on CIFAR-100 is pruned using GraSP, while DeiT-Tiny pre-trained on ImageNet is pruned using WDPruning \cite{yu2022width}. The pruned networks are subsequently fine-tuned or trained from scratch for the same number of epochs.

\section{More Comparison Results}
\label{more-pruning-results}
In this Section, we provide additional comparison results of pruning methods under contrast settings, such as unstructured vs. structured, to expand on the related content discussed in Section 7 in the main text. For instance, Table~\ref{tab:summary-pruning-cnns-classification} - Table~\ref{tab:summary-pruning-llms}\footnote{The results in Table~\ref{tab:summary-pruning-cnns-classification} - Table~\ref{tab:summary-pruning-llms} focus on reflecting the pruning outcomes under a pair of contrast settings, rather than comparing the specific pruning methods themselves.} offer a comprehensive comparison of different pruning methods applied to CNNs and Transformer-based models of varying sizes (small, medium, or large).

\subsection{\small More Results for Unstructured vs. Structured Pruning}
\label{more-results-unstruct-struct}
Table~\ref{tab:unstructured-and-structured-llm-more} compares the results of applying the same scoring method (LLM-surgeon~\cite{ouderaa2024llm}) to both unstructured and structured pruning across various models and five prune ratios. The results, sourced from \cite{ouderaa2024llm}, indicate that unstructured pruning consistently outperforms structured pruning at the same prune ratio. Additionally, Table~\ref{tab:summary-pruning-cnns-classification} - Table~\ref{tab:summary-pruning-llms} show that unstructured pruning typically achieves higher prune ratios, such as over 70\% or 80\%, while structured pruning generally remains below 50\%. In terms of performance, unstructured pruning generally results in better outcomes at similar prune ratios. For example, under similar settings, the unstructured pruning method, Jackpot \cite{zhang2023lottery} in Table~\ref{tab:summary-pruning-cnns-classification}, results in a Top-1 accuracy loss of 0.44 on ImageNet at an 80\% prune ratio, whereas the structured method, SCOP \cite{tang2020scop}, incurs a 0.89 loss at a 51.80\% prune ratio. The advantages and disadvantages of unstructured and structured pruning are illustrated in Table~\ref{Tab:unstruc-struct-comparison}. 

\begin{table}[ht]
  \centering
  \caption{Perplexity of unstructured and structured pruning using LLM-surgeon \cite{ouderaa2024llm} on LLMs with WikiText2~\cite{merity2016pointer} (lower is better). \textbf{Bold}/\underline{Underline} mark the best/second best performance, respectively, among the compared entities.}
  \label{tab:unstructured-and-structured-llm-more}
  \scalebox{0.95}{
  \begin{tabular}{l|c|c|c|c|c|c}
  \Xhline{0.3ex}      
  \multirow{2}{*}{Method} & \multirow{2}{*}{Model} & \multicolumn{5}{c}{Prune Ratio (\%)} \\
  \cline{3-7}
  &  & 10.00 & 20.00 & 30.00 & 40.00 & 50.00 \\
  \hline  
  unstruct & \multirow{2}{*}{OPT-125M}& \textbf{27.69} & \textbf{27.83} & \textbf{28.35} &  \textbf{28.98} & \textbf{30.30} \\
  struct & & \underline{28.01} & \underline{28.73} & \underline{31.82} & \underline{38.47} & \underline{49.78} \\
  \hline
  unstruct & \multirow{3}{*}{OPT-1.3B}& \textbf{14.62} & \textbf{14.66} & \textbf{14.81} &  \textbf{14.91} & \textbf{15.47} \\
  struct & & \underline{14.70} & \underline{15.12} & \underline{16.24} & \underline{18.45} & \underline{22.95} \\
  \hline
  unstruct & \multirow{3}{*}{OPT-2.7B}& \textbf{12.01} & \textbf{12.14} & \textbf{12.25} &  \textbf{12.28} & \textbf{12.68} \\
  struct & & \underline{12.02} & \underline{12.27} & \underline{12.92} & \underline{14.23} & \underline{17.15} \\
  \hline
  unstruct & \multirow{3}{*}{OPT-6.7B}& \underline{10.86} & \textbf{10.87} & \textbf{10.82} & \textbf{10.83} & \textbf{10.97} \\
  struct & & \textbf{10.77} & \underline{11.02} & \underline{11.64} & \underline{12.58} & \underline{14.90} \\  
  \hline 
  unstruct  & \multirow{2}{*}{LLaMA-2-7B} & \textbf{5.13} & \textbf{5.20} & \textbf{5.36} & \textbf{5.66} & \textbf{6.08} \\
  struct &  & \underline{5.25} & \underline{6.18} & \underline{7.83} & \underline{10.39} & \underline{15.38} \\
  \Xhline{0.3ex}      
\end{tabular}
}
\vspace{-0.1cm}
\end{table}

\begin{table}[ht]
  \caption{Advantages and disadvantages of unstructured and structured pruning.}
  \centering
  \label{Tab:unstruc-struct-comparison}
  \vspace{-0.1cm}
  \begin{tabular}{l|c|c}
     \Xhline{0.3ex}      
     & Unstructured & Structured \\
     \hline 
     High sparsity with  & \multirow{2}{*}{\CheckmarkBold} &  \multirow{2}{*}{hard} \\ 
     minor accuracy drop & & \\
     \hline
     Speedup w/o specific hardware & \multirow{2}{*}{hard} & \multirow{2}{*}{\CheckmarkBold}\\
     (e.g., FPGAs or ASICs) & & \\
     \hline
     Speedup w/o specific software & \multirow{2}{*}{hard} & \multirow{2}{*}{\CheckmarkBold}\\
     (e.g., sparsity CNNs libraries) & & \\
     \hline
     Really compressed with & \multirow{2}{*}{hard} & \multirow{2}{*}{\CheckmarkBold}\\
     significant acceleration & & \\
     \hline
     Structure coupling & \XSolidBrush & \CheckmarkBold\\
    \Xhline{0.3ex}      
\end{tabular}
\vspace{-0.1cm}
\end{table}

\subsection{More Results for One-shot vs. Iterative Pruning}
\label{more-results-one-shot-iterative}
In Table~\ref{tab:summary-pruning-cnns-classification}, pruning 90\% on ResNet-50 with ImageNet by using SNIP \cite{lee2019snip} results in a Top-1 accuracy loss of 14.10, and pruning 95\% leads to a loss of 31.30. In contrast, iterative-SNIP \cite{jorge2021progressive} shows a lower Top-1 accuracy loss of 11.90 at 90\% pruning and 30.90 at 95\% pruning. Similarly, iterative pruning using LoRAPruner \cite{zhang2023loraprune} in Table~\ref{tab:summary-pruning-llms} demonstrates superior performance compared to one-shot pruning with LLM-Pruner \cite{ma2023llmpruner}, under a similar setting with equivalent prune ratios on LLaMA-7B. However, the influence of specific functions on performance may overshadow the advantages of iterative pruning over one-shot pruning. For instance, one-shot pruning SAViT \cite{zheng2022savit} in Table~\ref{tab:summary-pruning-vits} shows a Top-1 accuracy drop of 1.48 when 30.77\% of FLOPs on DeiT-Tiny are pruned, outperforming iterative pruning SPViT \cite{he2024pruning} under similar settings, which exhibits a Top-1 accuracy drop of 1.50 with only 23.08\% of FLOPs pruned.

\subsection{More Results for Data-free vs. Data-driven Pruning}
\label{more-results-data-free-data-driven}
Consistent with the main text, the effectiveness of pruning in PBT methods is not strictly dependent on data usage. For example, when pruning ResNet-50 on ImageNet, data-free pruning NTK-SAP \cite{wang2023ntksap}, as shown in Table~\ref{tab:summary-pruning-cnns-classification}, outperforms data-driven pruning methods SNIP \cite{lee2019snip} and Grasp \cite{wang2020picking}, with a Top-1 accuracy loss of 15.41 at a 95.60\% prune ratio, compared to losses of 35.51 and 16.47 for SNIP and Grasp, respectively. 
However, for PAT methods, data usage appears crucial for maintaining performance in pruned models. For instance, under similar settings, data-free pruning DFPC \cite{narshana2023dfpc} shows a Top-1 accuracy drop of 1.20 when 49.49\% of FLOPs are pruned, whereas data-driven methods GFP \cite{liu2021group} and PGMPF \cite{cai2022prior} show smaller drops of 0.37 at a higher 50.11\% of FLOPs removed and 0.90 at a higher 53.50\% of FLOPs pruned, respectively.
Most pruning methods for Transformer-based models are data-driven, typically requiring target or calibration data during the pruning process. For instance, the pruning methods for LLMs in Table~\ref{tab:summary-pruning-llms} are all data-driven methods. 

\subsection{\scalebox{0.80}{More Results for Pruning on 
Initialized vs. Pre-trained Weights}}
\label{more-results-initialized-pretrained}
\begin{figure}[t]
\centering
 \subfloat[VGG-16 on CIFAR-10]{  
  \begin{minipage}{4cm}
       \includegraphics[width=4cm,height=2.5cm]{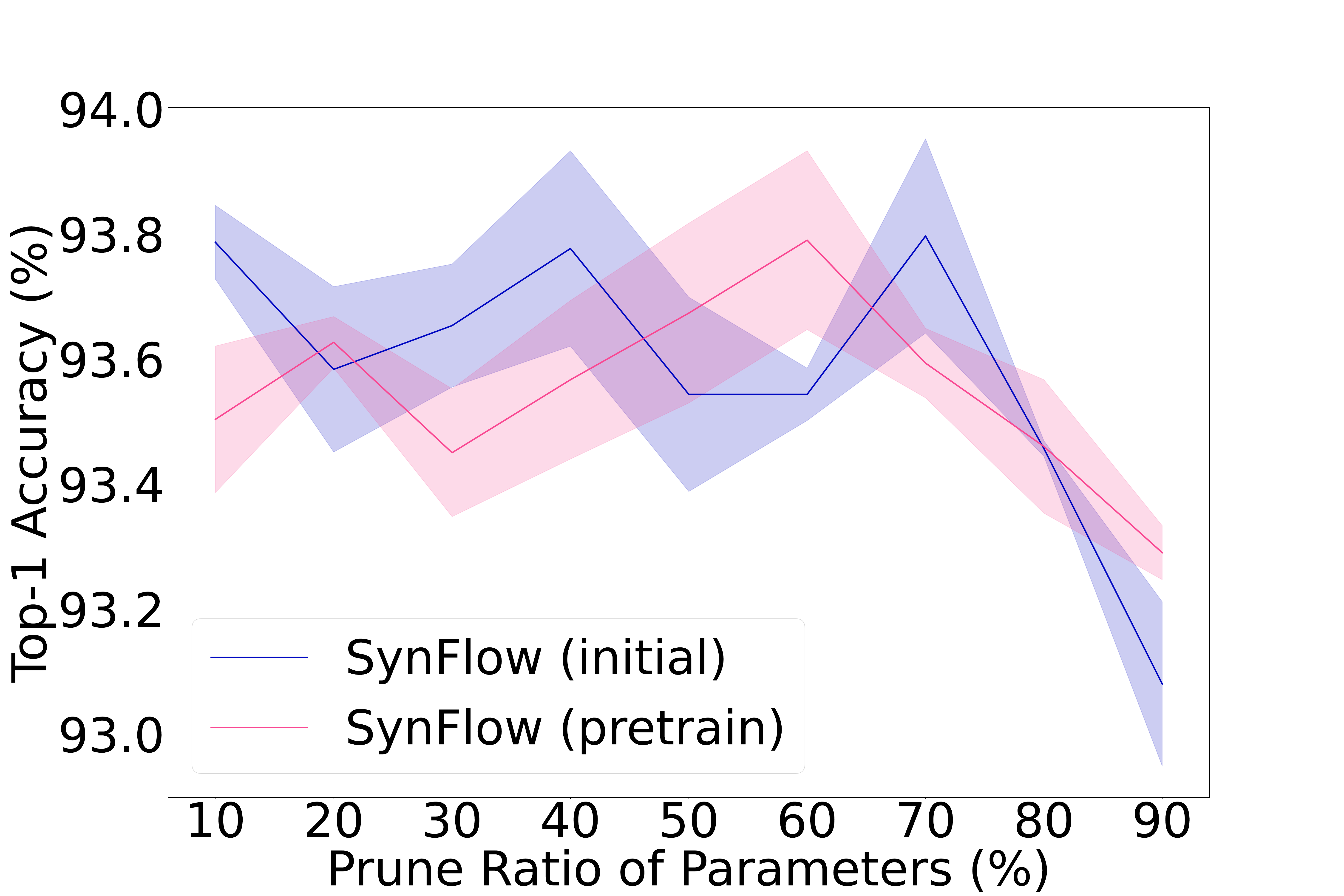}
     \label{Fig:appendix-initialized-pretrained-weights-a}
  \end{minipage} 
  }
  \subfloat[ResNet-32 on CIFAR-10]{  
  \begin{minipage}{4cm}
       \includegraphics[width=4cm,height=2.5cm]{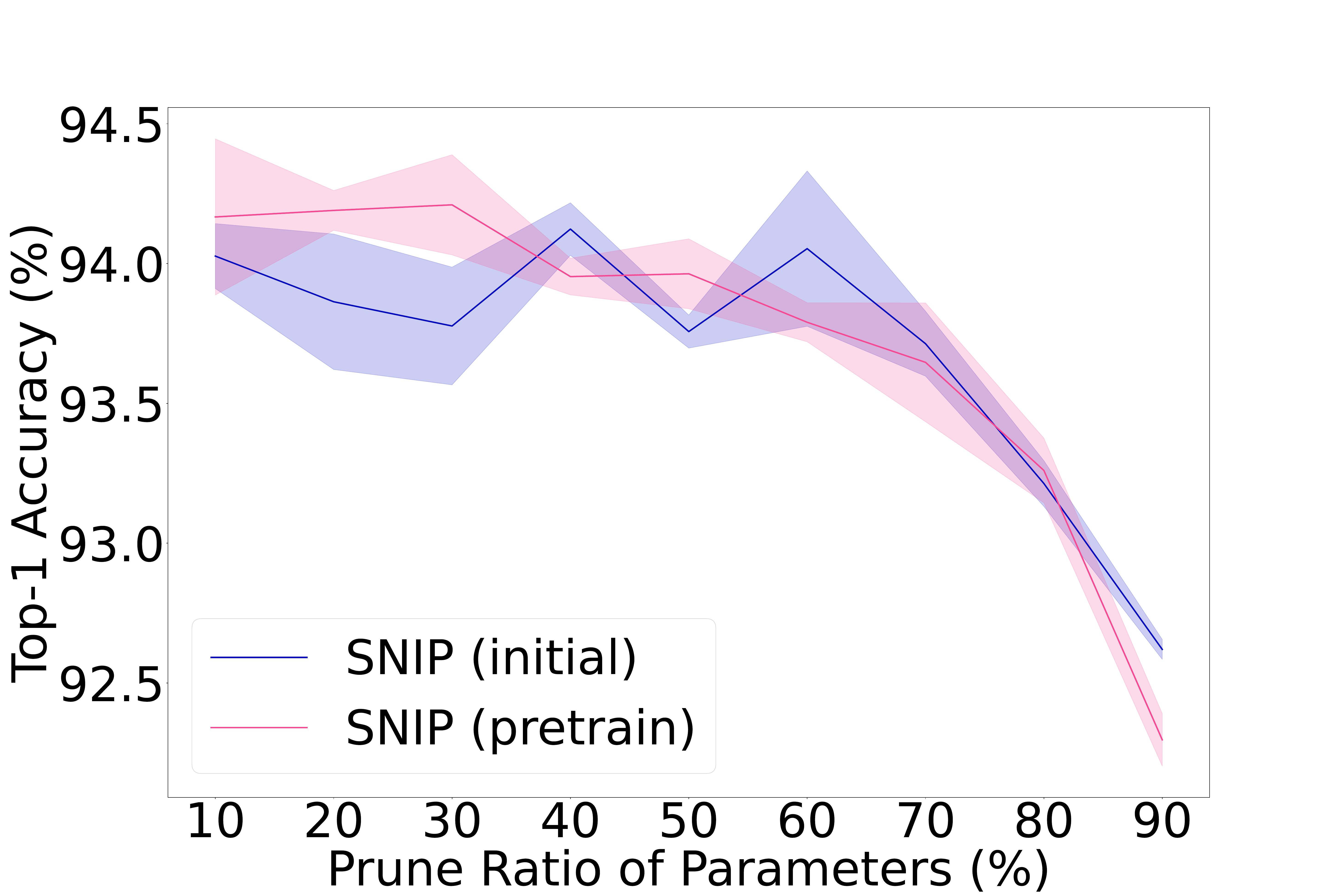}
     \label{Fig:appendix-initialized-pretrained-weights-b}
  \end{minipage} 
  }  
 \caption{PBT methods on initialized vs. pre-trained weights. Shaded regions indicate standard deviation based on three independent runs (best view in color).}
 \label{Fig:appendix-initialized-pretrained-weights}
\vspace{-0.3cm}
\end{figure}

We prune a randomly initialized VGG-16 and a VGG-16 pre-trained on CIFAR-10 using SynFlow to obtain the pruning results on initialized and pre-trained weights, respectively. A similar approach is applied to ResNet-32 as well. The results in Fig.~\ref{Fig:appendix-initialized-pretrained-weights} show that for PBT methods, SynFlow \cite{tanaka2020pruning} and SNIP \cite{lee2019snip}, pruning on the pre-trained weights does not guarantee improved Top-1 accuracy. \citet{cai2022prior} compare the results of pruning ResNet-34 on ImageNet from scratch and with pre-trained weights, and the results indicate that pre-trained weights are generally crucial for non-PBT methods to find effective subnetworks. Table~\ref{tab:summary-pruning-cnns-classification} - Table~\ref{tab:summary-pruning-llms} show that most pruning methods, especially those for Transformer-based models, are based on pre-trained rather than randomly initialized weights.

\subsection{More Results for Global vs. Local Pruning}
\label{more-results-global-local}
Tables~\ref{tab:summary-pruning-cnns-classification} - Table~\ref{tab:summary-pruning-llms} show that more pruning methods use global than local pruning. At an equivalent pruning ratio, global pruning methods, such as Bonsai \cite{dery2024everybody} in Table~\ref{tab:summary-pruning-llms}, can outperform local pruning methods like LLM-pruner \cite{ma2023llmpruner} and LoRAPruner \cite{zhang2023loraprune}. For instance, with a 50\% parameter pruning ratio for LLaMA-7B, Bonsai achieves a result of 10.92 on WikiText2 and 67.22 and 61.64 on BoolQ and WinoGrande, respectively, which are better by a large margin than the results of LLM-pruner (16.41, 60.28, 53.43) and LoRAPruner (11.60, 61.88, 55.01). However, the results can still vary significantly with similar settings for different global pruning methods. For example, as shown in Table 4, when CP \cite{he2017channel} has a FLOPs pruning ratio of 50.11\%, the Top-1 accuracy loss of the pruned ResNet-50 on ImageNet is 1.07, whereas GFP \cite{liu2021group} only has a loss of 0.37. Even with global pruning, achieving a globally optimal result is challenging.

\subsection{\scalebox{0.90}{More Results for Training from Scratch vs. Fine-tuning}}
Fine-tuning a pruned network or training it from scratch are two methods for restoring the performance of pruned models. Among the pruning methods listed in Table~\ref{tab:summary-pruning-cnns-classification} - Table~\ref{tab:summary-pruning-llms}, the majority opt for the former, such as ThiNet \cite{luo2017thinet} and GFP \cite{liu2021group} in Table~\ref{tab:summary-pruning-cnns-classification}, and SCOP \cite{tang2020scop} and PLATON \cite{zhang2022platon} in Table~\ref{tab:summary-pruning-vits}. A few methods, like DMCP \cite{guoO2020dmcp} in Table~\ref{tab:summary-pruning-cnns-classification}, choose the latter. CP \cite{he2017channel} in Table~\ref{tab:summary-pruning-cnns-classification} conducts comparative experiments between the two methods. Results indicate that, under the same pruning ratio (50\%), fine-tuning incurs a loss of 1.40 in Top-1 accuracy for ResNet-50 on ImageNet, while training from scratch results in a loss of 4.00.

\begin{table*}[t]
  \centering
  \caption{A summary of representative pruning methods for CNNs in image classification. ``U/S'' denotes unstructured and structured pruning, respectively. “Ratio” and ``FLOPs'' refer to reduced percentages of parameters and FLOPs, respectively. ``Original'' and ``Pruned'' represent Top-1 accuracy of the original and pruned networks, respectively. Entries marked with an asterisk (*) denote Top-5 accuracy. ``$\dagger$'' denotes MACs. If a second citation is present, it indicates the reported results. ``-'' means the corresponding result is not reported.}
  \small
  \label{tab:summary-pruning-cnns-classification}
  \scalebox{0.69}{
  \begin{tabular}{l|c|c|c|c|c|c|c|c|c|c|c|c}
  \Xhline{0.3ex}      
  \multirow{2}{*}{Method} & \multirow{2}{*}{U/S} & One-shot & Local & Data-driven & Pre-trained & Post-training & \multirow{2}{*}{Model} & Ratio & FLOPs & \multicolumn{3}{|c}{ImageNet} \\
  \cline{11-13}
  & & (Y/N) & (Y/N) & (Y/N) & (Y/N) & (Y/N) & & (\%) & (\%) &  Original (\%) & Pruned (\%) & $\Delta \downarrow$ \\
  \hline
  SNIP (2019) \cite{lee2019snip} & U & Y & N & Y & N & - & ResNet-50 & 80.00  & - & 76.15 & 69.67 & 6.48\\
  SNIP (2019) \cite{lee2019snip} \cite{wang2023ntksap} & U & Y & N & Y & N & - & ResNet-50 & 89.26  & - & 76.20 & 60.98 & 15.22\\   
  SNIP (2019) \cite{lee2019snip} \cite{wang2023ntksap} & U & Y & N & Y & N & - & ResNet-50 & 95.60  & - & 76.20 & 40.69 & 35.51\\
  SNIP (2019) \cite{lee2019snip} \cite{jorge2021progressive} & U & Y & N & Y & N & - & ResNet-50 & 90.00  & - & 75.60 & 61.50 & 14.10\\  
  SNIP (2019) \cite{lee2019snip} \cite{jorge2021progressive} & U & Y & N & Y & N & - & ResNet-50 & 95.00  & - & 75.60 & 44.30 & 31.30\\
  GSM (2019) \cite{ding2019global} & U & N & N & Y & Y & N & ResNet-50 & 75.00  & - & 75.72 & 75.33 & 0.39\\
  GSM (2019) \cite{ding2019global} & U & N & N & Y & Y & N & ResNet-50 & 80.00  & - & 75.72 & 74.30 & 1.42\\
  Iterative-SNIP (2021) \cite{jorge2021progressive} & U & N & N & Y & N & - & ResNet-50 & 90.00  & - & 75.60 & 63.70 & 11.90\\  
  Iterative-SNIP (2021) \cite{jorge2021progressive} & U & N & N & Y & N & - & ResNet-50 & 95.00  & - & 75.60 & 54.70 & 30.90\\    
  GraSP (2020) \cite{wang2020picking} & U & Y & N & Y & N & - & ResNet-50 & 80.00  & - & 76.15 & 72.06 & 4.09\\
  GraSP (2020) \cite{wang2020picking} \cite{wang2023ntksap} & U & Y & N & Y & N & - & ResNet-50 & 89.26 & - & 76.20 & 67.74 & 8.46\\
  GraSP (2020) \cite{wang2020picking} \cite{wang2023ntksap} & U & Y & N & Y & N & - & ResNet-50 & 95.60 & - & 76.20 & 59.73 & 16.47\\
  SynFlow (2020) \cite{tanaka2020pruning} \cite{wang2023ntksap} & U & N & N & N & N & - & ResNet-50 & 89.26 & - & 76.20 & 66.81 & 9.39\\  
  SynFlow (2020) \cite{tanaka2020pruning} \cite{wang2023ntksap} & U & N & N & N & N & - & ResNet-50 & 95.60 & - & 76.20 & 58.88 & 17.32\\
  RigL (2020) \cite{evci2020rigging} & U & Y & Y & Y & N & - & ResNet-50 & 80.00  & - & 76.15 & 74.60 & 1.55\\
  FORCE (2021) \cite{jorge2021progressive} & U & N & N & Y & N & - & ResNet-50 & 90.00  & - & 75.60 & 64.90 & 10.70\\
  FORCE (2021) \cite{jorge2021progressive} & U & N & N & Y & N & - & ResNet-50 & 95.00  & - & 75.60 & 59.00 & 16.60\\  
  ProsPr (2022) \cite{alizadeh2022prospect} & U & Y & N & Y & N & - & ResNet-50 & 90.00  & - & 75.60 & 66.86 & 8.74\\
  ProsPr (2022) \cite{alizadeh2022prospect} & U & Y & N & Y & N & - & ResNet-50 & 95.00  & - & 75.60 & 59.62 & 15.98\\ 
  NTK-SAP (2023) \cite{wang2023ntksap} & U & N & N & N & N & - & ResNet-50 & 89.26  & - & 76.20 & 68.28 & 7.92\\
  NTK-SAP (2023) \cite{wang2023ntksap} & U & N & N & N & N & - & ResNet-50 & 95.60  & - & 76.20 & 60.79 & 15.41\\ 
  Jackpot (2023) \cite{zhang2023lottery} & U & Y & Y & Y & Y & N & ResNet-50 & 80.00  & - & 76.15 & 75.71 & 0.44\\
  Jackpot (2023) \cite{zhang2023lottery} & U & Y & Y & Y & Y & N & ResNet-50 & 90.00  & - & 76.15 & 73.04 & 3.11\\
  FCPTS (2024) \cite{gong2024fast} & U & Y & Y & Y & Y & N & ResNet-50 & 50.00 & - & 77.89 & 77.43 & 0.46 \\  
  \hline
  ThiNet (2017) \cite{luo2017thinet} & S & N & N & Y & Y & Y & ResNet-50 & 33.72  & 36.79 & 75.30 & 74.03 & 1.27\\
  ThiNet (2017) \cite{luo2017thinet} & S & N & N & Y & Y & Y & ResNet-50 &  51.56  & 55.83 & 75.30 & 72.03 & 3.27\\
  ThiNet (2017) \cite{luo2017thinet} & S & N & N & Y & Y & Y & ResNet-50 &  66.12  & 71.50 & 75.30 & 68.17 & 7.13\\
  CP (2017) \cite{he2017channel} \cite{liu2021group} & S & N & N & Y & Y & Y & ResNet-50 & -  & 50.11 & 76.13 & 75.06 & 1.07\\
  CP (2017)-train-from-scratch \cite{he2017channel} & S & N & N & Y & Y & N & ResNet-50 & - & 50.00 & $92.20^{*}$ & $88.20^{*}$ & 4.00\\  
  CP (2017)-finetuned \cite{he2017channel} & S & N & N & Y & Y & Y & ResNet-50 & - & 50.00 & $92.20^{*}$ & $90.80^{*}$ & 1.40\\  
  SFP (2018) \cite{he2018soft} & S & Y & Y & Y & N & - & ResNet-50 & -  & 41.80 & 76.15 & 74.61 & 1.54\\ 
  SSS (2018) \cite{huang2018data} & S & N & N & Y & N & - & ResNet-50 &  0.78  & 15.06 & 76.12 & 75.44 & 0.68\\  
  SSS (2018) \cite{huang2018data} & S & N & N & Y & N & - & ResNet-50 &  27.06 & 31.08 & 76.12 & 74.18 & 1.94\\  
  IE (2019) \cite{molchanov2019importance} & S & N & N & Y & Y & Y &  ResNet-50 & - & 20.03 & 76.18 & 76.43 & -0.25 \\
  IE (2019) \cite{molchanov2019importance} & S & N & N & Y & Y & Y & ResNet-50 & - & 44.97 & 76.18 & 74.50 & 1.68 \\
  IE (2019) \cite{molchanov2019importance} & S & N & N & Y & Y & Y & ResNet-50 & - & 67.23 & 76.18 & 71.69 & 4.49 \\
  GBN (2019) \cite{you2019gate} & S & N & N & Y & Y & Y & ResNet-50 & -  & 40.57 & 75.85 & 76.19 & -0.34 \\ 
  Meta (2019) \cite{liu2019metapruning} & S & Y & N & Y & Y & N & ResNet-50 & - & 26.83 & 76.60 & 76.20 & 0.40 \\
  FPGM-prune-from-scratch (2019) \cite{he2019filter} & S & Y & Y & Y & N & - & ResNet-50 & - & 42.20 & 76.15 & 75.03 & 1.12 \\    
  FPGM-finetuned (2019) \cite{he2019filter} & S & Y & Y & Y & Y & Y & ResNet-50 & - & 42.20 & 76.15 & 75.59 & 0.56 \\    
  FPGM-prune-from-scratch (2019) \cite{he2019filter} & S & Y & Y & Y & N & - & ResNet-50 & - & 53.50 & 76.15 & 74.13 & 2.02 \\    
  FPGM-finetuned (2019) \cite{he2019filter} & S & Y & Y & Y & Y & Y & ResNet-50 & - & 53.50 & 76.15 & 74.83 & 1.32 \\
  PFS (2020) \cite{wang2020pruning} & S & N & N & Y & N & - & ResNet-50 & 16.74 & 26.83 & 77.20 & 76.70 & 0.50\\ 
  PFS (2020) \cite{wang2020pruning} & S & N & N & Y & N & - & ResNet-50 & 71.16 & 51.22 & 77.20 & 75.60 & 1.60\\  
  HRank (2020) \cite{lin2020hrank} & S & Y & Y & Y & Y & Y & ResNet-50 & 36.70 & 43.70 & 76.15 & 74.98 & 1.17 \\
  SCOP-A (2020) \cite{tang2020scop} & S & Y & Y & Y & Y & Y & ResNet-50 & 42.80 & 45.30 & 76.15 & 75.95 & 0.20 \\
  SCOP-B (2020) \cite{tang2020scop} & S & Y & Y & Y & Y & Y & ResNet-50 & 51.80 & 54.60 & 76.15 & 75.26 & 0.89 \\
  AutoPruner (2020) \cite{luo2020autopruner} & S & N & N & Y & Y & Y & ResNet-50 & - &  51.10 & 76.15 & 74.76 & 1.39 \\
  AutoPruner (2020) \cite{luo2020autopruner} & S & N & N & Y & Y & Y & ResNet-50 & - &  66.01 & 76.15 & 73.05 & 3.10 \\
  LFPC (2020) \cite{he2020learning} & S & Y & Y & Y & Y & Y & ResNet-50 & - & 25.17 & 76.79 & 76.95 & -0.16 \\
  DSA (2020) \cite{ning2020dsa} & S & Y & Y & Y & N & - & ResNet-50 & - & 40.00 & 76.02 & 75.10 & 0.92 \\
  DSA (2020) \cite{ning2020dsa} & S & Y & Y & Y & N & - & ResNet-50 & - & 50.00 & 76.02 & 74.69 & 1.33 \\
  DMCP-train from scratch (2020) \cite{guoO2020dmcp} & S & N & Y & Y & N & - & ResNet-50 & - & 31.70 & 76.60 & 77.00 & -0.40 \\  
  DMCP-train-from-scratch (2020) \cite{guoO2020dmcp} & S & N & Y & Y & N & - & ResNet-50 & - & 46.34 & 76.60 & 76.20 & 0.40 \\  
  GFP (2021) \cite{liu2021group} & S & N & N & Y & Y & Y & ResNet-50 & - & 25.17 & 76.79 & 76.95 & -0.16 \\
  GFP (2021) \cite{liu2021group} & S & N & N & Y & Y & Y & ResNet-50 & - & 50.11 & 76.79 & 76.42 & 0.37 \\
  GFP (2021) \cite{liu2021group} & S & N & N & Y & Y & Y & ResNet-50 & - & 75.06 & 76.79 & 73.94 & 2.85 \\ 
  Greg-2 (2021) \cite{wang2021neural} & S & N & Y & Y & Y & Y & ResNet-50 & - & 32.93 & 76.13 & 75.36 & 0.77 \\  
  PGMPF (2022) \cite{cai2022prior} & S & N & Y & Y & Y & Y & ResNet-50 & - & 53.50 & 76.01 & 75.11 & 0.90 \\
  DFPC (2023) \cite{narshana2023dfpc} & S & N & N & N & Y & Y & ResNet-50 & 45.65 & 49.49 & 76.10 & 75.90 & 1.20 \\  
  DFPC (2023) \cite{narshana2023dfpc} & S & N & N & N & Y & Y & ResNet-50 & 62.26 & 71.10 & 76.10 & 73.80 & 2.30 \\
  PDP (2023) \cite{cho2023pdp} & S & N & N & Y & N & - & ResNet-50 & -  & $54.90^{\dagger}$ & 76.10 & 75.90 & 0.20\\
  PPSM (2023) \cite{ma2022differentiable} & S & N & N & Y & Y & Y & ResNet-50 & - & 53.07 & 76.13 & 75.78 & 0.35 \\  
  PPSM (2023) \cite{ma2022differentiable} & S & N & N & Y & Y & Y & ResNet-50 & - & 65.91 & 76.13 & 75.43 & 0.70 \\  
  DepGraph (2023) \cite{fang2023depgraph} & S & Y & N & Y & Y & Y & ResNet-50 & - & $51.82^{\dagger}$ & 76.15 & 75.83 & 0.32 \\
  ATO (2024) \cite{wu2024auto} & S & N & N & Y & N & - & ResNet-50 & - & 55.20 & 76.13 & 76.59 & -0.46 \\
  ATO (2024) \cite{wu2024auto} & S & N & N & Y & N & - & ResNet-50 & - & 61.70 & 76.13 & 76.07 & 0.06 \\ 
  ATO (2024) \cite{wu2024auto} & S & N & N & Y & N & - & ResNet-50 & - & 71.00 & 76.13 & 74.77 & 1.36 \\   
  \hline
  SSS (2018) \cite{huang2018data} & S & N & N & N & Y & N & ResNet-101 & -  & 55.52 & 76.40 & 75.44 & 0.96 \\
  SFP (2018) \cite{he2018soft} & S & Y & Y & Y & N & - & ResNet-101 & -  & 42.20 & 77.37 & 77.03 & 0.34\\
  IE (2019) \cite{molchanov2019importance} & S & N & N & Y & Y & Y & ResNet-101 & 30.20 & 39.74 & 77.37 & 77.35 & 0.02 \\  
  FPGM-finetuned (2019) \cite{he2019filter} & S & Y & Y & Y & Y & Y & ResNet-101 & - & 42.20 & 77.37 & 77.32 & 0.05 \\
  SCOP-A (2020) \cite{tang2020scop} & S & Y & Y & Y & Y & Y & ResNet-101 & 46.80 & 48.60 & 77.37 & 77.75 & -0.48 \\
  SCOP-B (2020) \cite{tang2020scop} & S & Y & Y & Y & Y & Y & ResNet-101 & 57.80 & 60.20 & 77.37 & 77.36 & 0.01 \\
  GFP (2021) \cite{liu2021group} & S & N & N & Y & Y & Y & ResNet-101 & 39.63 & 50.02 & 78.29 & 78.30 & -0.01 \\
  \Xhline{0.3ex}      
\end{tabular}
}
\vspace{-0.3cm}
\end{table*}

\begin{table*}[t]
  \centering
  \caption{A summary of representative pruning methods in object detection. ``O-mAP'' and ``P-mAP'' represent mAP of the original and pruned networks, respectively.  “Ratio” and ``FLOPs'' have the same meaning as in Table~\ref{tab:summary-pruning-cnns-classification}.}
  \small
  \label{tab:summary-pruning-cnns-detection}
  \scalebox{0.68}{
  \begin{tabular}{l|c|c|c|c|c|c|c|c|c|c|c|c|c|c}
  \Xhline{0.3ex}      
  \multirow{2}{*}{Method} & \multirow{2}{*}{U/S} & One-shot & Local & Pre-trained & Post-training & \multirow{2}{*}{Model} & Ratio & FLOPs & \multicolumn{3}{|c|}{PASCAL VOC} & \multicolumn{3}{|c}{COCO}\\
  \cline{10-15}
  & & (Y/N) & (Y/N) & (Y/N) & (Y/N) &  & (\%) & (\%) & O-mAP & P-mAP & $\Delta \downarrow$ & O-mAP & P-mAP & $\Delta \downarrow$ \\
  \hline
  LTH (2021) \cite{girish2021lottery} & U & N & N & Y & N & Faster-RCNN (ResNet-18) & 78.94 & - & 69.74 & 68.47 & 1.27 & - & - & - \\ 
  CP (2017) \cite{he2017channel} & S & N & N & Y & Y & Faster R-CNN (VGG-16) & -  & 50.00 & 68.70 & 68.30 & 0.40 & - & - & -\\
  GFP (2021) \cite{liu2021group} & S & N & N & Y & Y & Faster R-CNN & - & 50.00 & - & - & - & 37.40 & 37.80 & -0.40\\
  GFP (2021) \cite{liu2021group} & S & N & N & Y & Y & Faster R-CNN & - & 75.00 & - & - & - & 37.40 & 36.60 & 0.80\\ 
  SAViT (2022) \cite{zheng2022savit} & S & Y & N & Y & Y & Faster R-CNN (Swin-T) & 57.52 & 69.42 & - & - & - & 45.50 & 45.20 & 0.30\\
  \hline
  FCPTS (2024) \cite{gong2024fast} & U & Y & Y & Y & N & MobileNetV1 SSD & 90.00 & - & 67.70 & 65.10 & 2.60 & - & - & -\\
  FCPTS (2024) \cite{gong2024fast} & U & Y & Y & Y & N & MobileNetV2 SSD-Lite & 90.00 & - & 68.60 & 59.10 & 9.50 & - & - & - \\
  GFP (2021) \cite{liu2021group} & S & N & N & Y & Y & RetinaNet & 31.03 & 50.02 & - & - & - & 36.50 & 36.80 & -0.30\\
  MDC (2022) \cite{hou2022multidimensional} & S & Y & Y & Y & Y & DeiT-Base & - & 60.00 & 81.80 & 81.50 & 0.30 & - & - & -\\
  \Xhline{0.3ex}      
\end{tabular}
}
\vspace{-0.1cm}
\end{table*}

\begin{table*}[t]
  \centering
  \caption{A summary of representative pruning methods for ViTs in image classification.}
  \small
  \label{tab:summary-pruning-vits}
  \scalebox{0.70}{%
  \begin{tabular}{l|c|c|c|c|c|c|c|c|c|c|c|c}
  \Xhline{0.3ex}      
  \multirow{2}{*}{Method} & \multirow{2}{*}{U/S} & One-shot & Local & Data-driven & Pre-trained & Post-training & \multirow{2}{*}{Model} & Ratio & FLOPs & \multicolumn{3}{|c}{ImageNet} \\
  \cline{10-13}
  & & (Y/N) & (Y/N) & (Y/N) & (Y/N) & (Y/N) &  & (\%) & (\%) & Original (\%) & Pruned (\%) & $\Delta \downarrow$\\
  \hline
  SViTE-T (2021) \cite{chen2021chasing} & U & N & Y & Y & N & - & DeiT-Tiny & 29.72 & 25.56 & 72.20 & 71.78 & 0.42 \\
  SViTE-T (2021) \cite{chen2021chasing} & U & N & Y & Y & N & - & DeiT-Tiny & 39.51 & 34.16 & 72.20 & 71.75 & 0.45\\  
  $\textrm{S}^{2}$ViTE-T \cite{chen2021chasing} (2021) & S & N & Y & Y & N & - & Deit-Tiny & 26.40 & 23.69 & 72.20 & 70.12 & 2.08\\
  UVC-T (2022) \cite{yu2022unified} & S & N & N & Y & Y & Y & DeiT-Tiny & - & 50.77 & 72.20 & 71.30 & 0.90\\
  UVC-T (2022)\cite{yu2022unified} & S & N & N & Y & Y & Y & DeiT-Tiny & - & 60.88 & 72.20 & 70.60 & 1.60 \\
  SAViT-T (2022) \cite{zheng2022savit} & S & Y & N & Y & Y & Y & DeiT-Tiny & 26.32 & 30.77 & 72.20 & 70.72 & 1.48\\
  GOHSP-T (2023) \cite{yin2023gohsp} & S & Y & N & Y & Y & Y & DeiT-Tiny & 29.82 & 30.00 & 72.20 & 70.24 & 1.96\\
  X-Pruner-T (2023) \cite{yu2023xpruner} & S & N & N & Y & Y & N & DeiT-Tiny & - & 50.80 & 72.20 & 71.10 & 1.10\\    
  SPViT-T (2024) \cite{he2024pruning} & S & N & N & Y & Y & Y & Deit-Tiny & 15.79 & 23.08 & 72.20 & 70.70 & 1.50\\
  \hline
  SViTE-S (2021)\cite{chen2021chasing} & U & N & Y & Y & N & - & DeiT-Small & 39.82 & 36.73 & 79.90 & 80.26 & -0.36\\
  SCOP-S (2020) \cite{tang2020scop} & S & Y & Y & Y & Y & Y & DeiT-Small & - & 43.48 & 79.80 & 77.50 & 2.30\\
  $\textrm{S}^{2}$ViTE-S (2021) \cite{chen2021chasing} & S & N & Y & Y & N & - & DeiT-Small & 33.94 & 31.63 & 79.90 & 79.22 & 0.68\\
  UVC-S (2022) \cite{yu2022unified} & S & N & N & Y & Y & Y & DeiT-Small & - & 49.59 & 79.80 & 78.82 & 0.98\\ 
  SAViT-S (2022) \cite{zheng2022savit} & S & Y & N & Y & Y & Y & DeiT-Small & 33.48 & 32.61 & 79.85 & 80.11 & -0.26\\   
  NViT-S (2023) \cite{yang2023global} & S & N & N & Y & Y & N & DeiT-Small & 4.55 & 8.70 & 81.20 & 82.19 & -0.99\\
  NViT-S-ASP (2023) \cite{yang2023global} & S & N & N & Y & Y & Y & DeiT-Small & 52.27 & 8.70 & 81.20 & 82.19 & -0.99\\
  GOHSP-S (2023) \cite{yin2023gohsp} & S & Y & N & Y & Y & Y & DeiT-Small & 34.84 & 35.00 & 79.90 & 79.98 & -0.08\\   
  GOHSP-S (2023) \cite{yin2023gohsp} & S & Y & N & Y & Y & Y & DeiT-Small & 49.77 & 39.00 & 79.90 & 79.86 & 0.04\\
  X-Pruner-S (2023) \cite{yu2023xpruner} & S & N & N & Y & Y & N & DeiT-Small & - & 47.90 & 79.80 & 78.93 & 0.87\\  
  SPViT-S (2024) \cite{he2024pruning} & S & N & N & Y & Y & Y & DeiT-Small & 28.26 & 28.05 & 79.90 & 78.30 & 1.60\\
  \hline
  SViTE-B (2021) \cite{chen2021chasing} & U & N & Y & Y & N & - & Deit-Base & 39.95 & 38.30 & 81.80 & 81.56 & 0.24\\
  $\textrm{S}^{2}$ViTE-B (2021) \cite{chen2021chasing} & S & N & Y & Y & N & - & Deit-Base & 34.41 & 33.13 & 81.80 & 82.22 & -0.42\\  
  SCOP-B (2020) \cite{tang2020scop} & S & Y & Y & Y & Y & Y & DeiT-Base & - & 41.70 & 81.80 & 79.70 & 2.10\\
  VTP-B (2021)\cite{zhu2021vision} & S & Y & N & Y & Y & Y & DeiT-Base & 22.11 & 21.59 & 81.80 & 81.30 & 0.50\\
  VTP-B (2021)\cite{zhu2021vision} & S & Y & N & Y & Y & Y & DeiT-Base & 44.44 & 43.18 & 81.80 & 80.70 & 1.10\\
  UVC-B (2022)\cite{yu2022unified} & S & N & N & Y & Y & Y & DeiT-Base & - & 54.50 & 81.80 & 80.57 & 1.23\\
  SAViT-B (2022)\cite{zheng2022savit} & S & Y & N & Y & Y & Y & DeiT-Base & 70.67 & 69.89 & 81.84 & 81.66 & 0.18\\
  CP-ViT-B (2022) \cite{song2022cpvit}-w/o finetune & S & N & Y & Y & Y & N & DeiT-Base & 30 & 22.16 & 81.82 & 80.91 & 0.91\\ 
  CP-ViT-B (2022) \cite{song2022cpvit}-w/o finetune & S & N & Y & Y & Y & N & DeiT-Base & 40 & 30.67 & 81.82 & 80.31 & 1.51\\ 
  CP-ViT-B (2022) \cite{song2022cpvit}-w/ finetune & S & N & Y & Y & Y & Y & DeiT-Base & 30 & 22.62 & 81.82 & 81.66 & 0.16\\ 
  CP-ViT-B (2022) \cite{song2022cpvit}-w/ finetune & S & N & Y & Y & Y & Y & DeiT-Base & 40 & 32.41 & 81.82 & 81.52 & 0.30\\ 
  CP-ViT-B (2022) \cite{song2022cpvit}-w/ finetune & S & N & Y & Y & Y & Y & DeiT-Base & 50 & 41.62 & 81.82 & 81.13 & 0.69\\ 
  WDPruning-B (2022)\cite{yu2022width} & S & N & Y & Y & Y & Y & DeiT-Base & - & 43.40 & 81.80 & 80.76 & 1.04\\
  MDC-B (2022) \cite{hou2022multidimensional} & S & Y & Y & Y & Y & Y & DeiT-Base & - & 54.29 & 81.80 & 82.30 & -0.50\\
  MDC-B (2022) \cite{hou2022multidimensional} & S & Y & Y & Y & Y & Y & DeiT-Base & - & 60.00 & 81.80 & 81.50 & 0.30\\
  X-Pruner-B (2023) \cite{yu2023xpruner} & S & N & N & Y & Y & N & DeiT-Base & - & 51.50 & 81.80 & 81.02 & 0.78\\
  NViT-B (2023)\cite{yang2023global} & S & N & N & Y & Y & N & DeiT-Base & 60.47 & 61.36 & 83.36 & 83.29 & 0.07\\
  NViT-B-ASP (2023)\cite{yang2023global} & S & N & N & Y & Y & Y & DeiT-Base & 80.23 & 61.36 & 83.36 & 83.29 & 0.07\\
  SPViT-B (2024) \cite{he2024pruning} & S & N & N & Y & Y & Y & Deit-Base & 52.00 & 51.85 & 81.80 & 81.50 & 0.30\\  
 \hline
  PLATON-B (2022) \cite{zhang2022platon} & U & N & N & Y & Y & Y & ViT-Base & 60 & - & 83.50 & 82.60 & 0.90 \\
  FCPTS-B (2024) \cite{gong2024fast} & U & Y & Y & Y & Y & N & ViT-Base & 50 & - & 75.68 & 74.90 & 0.78\\  
  FCPTS-B (2024) \cite{gong2024fast} & U & Y & Y & Y & Y & N & ViT-Base & 60 & - & 75.68 & 72.09 & 3.60\\  
  CP-ViT-B (2022) \cite{song2022cpvit}-w/o finetune & S & N & Y & Y & Y & N & ViT-Base & 30 & 23.02 & 77.91 & 76.77 & 1.14\\   
  CP-ViT-B (2022) \cite{song2022cpvit}-w/o finetune & S & N & Y & Y & Y & N & ViT-Base & 40 & 32.34 & 77.91 & 75.09 & 2.82\\
  CP-ViT-B (2022) \cite{song2022cpvit}-w/ finetune & S & N & Y & Y & Y & Y & ViT-Base & 30 & 24.91 & 77.91 & 77.75 & 0.16\\   
  CP-ViT-B (2022) \cite{song2022cpvit}-w/ finetune & S & N & Y & Y & Y & Y & ViT-Base & 40 & 33.62 & 77.91 & 77.36 & 0.55\\    
  CP-ViT-B (2022) \cite{song2022cpvit}-w/ finetune & S & N & Y & Y & Y & Y & ViT-Base & 50 & 46.34 & 77.91 & 76.75 & 1.16\\ 
  DepGraph-B (2023) \cite{fang2023depgraph} & S & Y & N & Y & Y & Y & ViT-Base & - & $40.91^{\dagger}$ & 81.07 & 79.17 & 1.90 \\
  \hline
  DIMAP-T (2023) \cite{he2024data} & U & Y & N & N & Y & Y & Swin-Tiny & - & 32.40 & 81.16 & 81.11 & 0.05\\   
  DIMAP-T (2023) \cite{he2024data} & U & Y & N & N & Y & Y & Swin-Tiny & - & 50.80 & 81.16 & 80.35 & 0.81\\  
  X-Pruner-T (2023) \cite{yu2023xpruner} & S & N & N & Y & Y & N & Swin-Tiny & - & 28.90 & 81.20 & 80.70 & 0.50\\
  DIMAP-S (2023) \cite{he2024data} & U & Y & N & N & Y & Y & Swin-Small & - & 33.20 & 83.19 & 82.99 & 0.20\\   
  DIMAP-S (2023) \cite{he2024data} & U & Y & N & N & Y & Y & Swin-Small & - & 52.30 & 83.19 & 82.63 & 0.56\\  
  WDPruning-S (2022)\cite{yu2022width} & S & N & Y & Y & Y & Y & Swin-Small & - & 27.58 & 83.00 & 81.80 & 1.20\\  
  X-Pruner-S (2023) \cite{yu2023xpruner} & S & N & N & Y & Y & N & Swin-Samll & - & 27.60 & 83.20 & 82.00 & 1.20\\
  DIMAP-B (2023) \cite{he2024data} & U & Y & N & N & Y & Y & Swin-Base & - & 33.50 & 83.48 & 83.43 & 0.05\\   
  DIMAP-B (2023) \cite{he2024data} & U & Y & N & N & Y & Y & Swin-Base & - & 52.70 & 83.48 & 83.28 & 0.20\\   
  \Xhline{0.3ex}      
\end{tabular}
}
\vspace{-0.3cm}
\end{table*}

\begin{table*}[t]
  \centering
  \caption{A summary of representative pruning methods for BERTs.}
  \small
  \label{tab:summary-pruning-berts}
  \scalebox{0.65}{%
  \begin{tabular}{l|c|c|c|c|c|c|c|c|c|c|c|c|c}
  \Xhline{0.3ex}      
  \multirow{2}{*}{Method} & \multirow{2}{*}{U/S} & One-shot & Local & Pre-trained & Post-training & \multirow{2}{*}{Model} & Ratio & FLOPs & \multicolumn{5}{|c}{Dataset} \\
  \cline{10-14}
  & & (Y/N) & (Y/N) & (Y/N) & (Y/N) &  & (\%) & (\%) & QQP & MNLI & QNLI & SST-2 & $\textrm{SQuAD}_{1.1}$\\
  \hline
  LadaBERT (2020) \cite{mao2020ladabert} & U & Y & N & Y & Y & $\textrm{BERT}_{base}$ & 60.00 & - & 71.20/70.70 & 84.60/83.50 & 90.50/89.60 & 93.50/92.80 & - \\ 
  Prune OFA (2021) \cite{zafrir2021prune} & U & N & N & Y & Y & $\textrm{BERT}_{base}$ & 85.00 & - & 91.20/90.69 & 84.06/81.67 & 91.16/89.95 & 92.13/91.34 & 80.80/78.59\\
  PLATON-unstruct (2022) \cite{zhang2022platon} & U & N & N & Y & Y & $\textrm{BERT}_{base}$ & 50.00 & - & - & - & - & - & 80.40/78.50\\
  PLATON-unstruct (2022)
  \cite{zhang2022platon} & U & N & N & Y & Y & $\textrm{BERT}_{base}$ & 60.00 & - & - & - & - & - & 80.40/78.00\\
  PLATON-unstruct (2022)
  \cite{zhang2022platon} & U & N & N & Y & Y & $\textrm{BERT}_{base}$ & 80.00 & - & 91.50/90.70 & 84.60/83.10 & 91.30/90.10 & 92.70/91.30 & -\\  
  CAP-m (2022) \cite{xu2022from} & U & N & Y/N & Y & Y & $\textrm{BERT}_{base}$ & 90.00 & - & 90.90/90.70 & 84.50/81.00 & - & 92.90/- & 80.70/76.50 \\ 
  oBERT (2022) \cite{kurtic2022optimal} & U & N & N & Y & Y & $\textrm{BERT}_{base}$ & 90.00 & - & 91.06/90.99 & 84.54/83.40 & - & 91.25/89.97 & - \\  
  ISP (2023) \cite{jaiswal2023instant} & U & N & N & Y & Y & $\textrm{BERT}_{base}$ & 70.00 & - & - & 82.40/82.71 & 89.10/90.06 & - & - \\
  \hline
  BSP (2020)\cite{li2020efficient} & S & Y & Y & Y & Y & $\textrm{BERT}_{base}$ & 30.00 & - & 91.20/90.70 & 84.60/82.90 & 90.50/88.20 & 93.50/89.30 & - \\ 
  SuperTicket (2021) \cite{liang2021super} & S & Y & N & Y & Y & $\textrm{BERT}_{base}$ & 13.20 & - & 91.30/88.30 & 84.50/84.50 & -/91.30 & - & -\\  
  PLATON-struct (2022)
  \cite{zhang2022platon} & S & N & N & Y & Y & $\textrm{BERT}_{base}$ & 50.00 & - & - & - & - & - & 80.4/77.00\\ 
  PLATON-struct (2022)
  \cite{zhang2022platon} & S & N & N & Y & Y & $\textrm{BERT}_{base}$ & 60.00 & - & - & - & - & - & 80.4/75.60\\ 
  Mask-Tuning (2022)
  \cite{kwon2022fast} & S & N & Y & Y & N & $\textrm{BERT}_{base}$ & - & 40.00 & 91.00/90.38 & 84.53/82.26 & 91.41/90.00 & 93.57/92.47 & -\\
  CAP-f (2022) \cite{xu2022from} & S & N & Y/N & Y & Y & $\textrm{BERT}_{base}$ & 90.00 & - & 90.90/90.20 & 84.50/81.00 & - & 92.90/89.70 & 80.70/70.20 \\ 
  KCM (2023)
  \cite{nova2023gradient} & S & Y & Y & Y & N & $\textrm{BERT}_{base}$ & - & 30.00 & 91.00/90.39 & 84.53/81.18 & 91.41/90.58 & 93.57/92.26 & -\\   
  KCM (2023)
  \cite{nova2023gradient} & S & Y & Y & Y & N & $\textrm{BERT}_{base}$ & - & 40.00 & 91.00/89.15 & 84.53/77.24 & 91.41/87.79 & 93.57/91.11 & -\\ 
  \Xhline{0.3ex}      
\end{tabular}
}
\vspace{-0.3cm}
\end{table*}

\begin{table*}[t]
  \centering
  \caption{A summary of representative pruning methods for LLMs. ``$\dagger$'' indicates the sequence length for inference is 2048, otherwise it is 128. If a second citation is present, it references the reported results. ``-'' means the corresponding result is not reported or N/A.}
  \small
  \label{tab:summary-pruning-llms}
  \scalebox{0.68}{%
  \begin{tabular}{l|c|c|c|c|c|c|cc|cccc|c}
  \Xhline{0.3ex}      
  \multirow{2}{*}{Method} & \multirow{2}{*}{U/S} & One-shot & Local & Post-training & Pre-trained & Ratio & \multicolumn{2}{|c|}{Prediction $\downarrow$} & \multicolumn{4}{|c|}{Common Sense $\uparrow$} & \multirow{2}{*}{MMLU $\uparrow$} \\
   \cline{8-9} \cline{10-13}
  &  & (Y/N) & (Y/N) & (Y/N) & Model & (\%) & WikiText2 & PTB & BoolQ & PIQA & Helleswag & WinoGrande  &  \\
  \hline
  Dense \cite{ma2023llmpruner} & - & - & - & - & LLaMA-7B & 0.00 & 12.62/$5.68^{\dagger}$ & 22.14 & 73.18 & 78.35 & 72.99 & 67.01 & 34.43 \cite{sun2024simple} \\
  Dense \cite{sun2024simple} & - & - & - & - & LLaMA-2-7B & 0.00 & 12.62 & 22.14 & 77.74 & - & 57.17 & 68.90 & 52.08 \\
  Dense \cite{bai2024sparsellm} & - & - & - & - & LLaMA-2-13B & 0.00 & $4.88^{\dagger}$ & $50.94^{\dagger}$ & 77.89 & - & 59.94 & 72.77 & - \\ 
  Dense \cite{song2024sleb} & - & - & - & - & OPT-6.7B & 0.00 & $10.86^{\dagger}$ & - & - & 76.39 & 67.16 & 65.19 & - \\  
  Dense \cite{bai2024sparsellm} & - & - & - & - & OPT-30B & 0.00 & $9.56^{\dagger}$ & $14.04^{\dagger}$ & 70.46 & - & 54.27 & 69.02 & - \\   
  Dense \cite{kim2024shortened} & - & - & - & - & Vicuna-7B & 0.00 & 17.10 & 63.20 & 78.10 & 77.30 & 73.90 & 69.50 & - \\
  \hline
  SparseGPT (2023)~\cite{frantar2023sparsegpt} \cite{sun2024simple} & U & Y & Y & N & LLaMA-7B & 50.00 & $7.22^{\dagger}$ & - & 75.02 & - & 52.37 & 69.85 & 34.43\\
  SparseGPT (2023)~\cite{frantar2023sparsegpt} \cite{yin2024outlier} & U & Y & Y & N & LLaMA-7B & 70.00 & $26.30^{\dagger}$ & - & 64.53 & - & 42.11 & 58.64 & - \\  
  Wanda (2024)~\cite{sun2024simple} & U & Y & Y & N & LLaMA-7B & 50.00 & $7.26^{\dagger}$ & - & 71.22 & - & 51.85 & 66.06 & 33.49\\
  Wanda (2024)~\cite{sun2024simple} \cite{yin2024outlier} & U & Y & Y & N & LLaMA-7B & 70.00 & $85.77^{\dagger}$ & - & 55.11 & - & 31.83 & 51.38 & -\\  
  BESA (2024)~\cite{xu2024besa} & U & Y & Y & N & LLaMA-7B & 50.00 & $6.86^{\dagger}$ & $66.96^{\dagger}$ & 72.17 & 76.66 & 54.31 & 67.64 & -\\
  OWL w. SparseGPT (2024)~\cite{yin2024outlier} & U & Y & Y & N & LLaMA-7B & 70.00 & $19.49^{\dagger}$ & - & 67.13 & - & 48.56 & 62.03 & -\\   
  OWL w. Wanda (2024)~\cite{yin2024outlier} & U & Y & Y & N & LLaMA-7B & 70.00 & $24.55^{\dagger}$ & - & 62.48 & - & 44.79 & 58.72 & -\\     
  LLM-Pruner (2023)~\cite{ma2023llmpruner} & S & Y & Y & Y & LLaMA-7B & 20.00 & 17.58 & 30.11 &  64.62 & 77.20 & 68.80 & 63.14 & - \\
  LLM-Pruner (2023)~\cite{ma2023llmpruner} & S & Y & Y & Y & LLaMA-7B & 50.00 & 38.12/$16.41^{\dagger}$ & 66.35 &  60.28 & 69.31 & 47.06 & 53.43 & -\\
  LoRAPruner (2023)~\cite{zhang2023loraprune} & S & N & Y & Y & LLaMA-7B & 20.00 & 16.80 & 28.75 &  65.62 & 79.31 & 70.00 & 62.76 & -\\
  LoRAPruner (2023)~\cite{zhang2023loraprune} & S & N & Y & Y & LLaMA-7B & 50.00 & 30.12/$11.60^{\dagger}$ & 50.30 &  61.88 & 71.53 & 47.86 & 55.01 & -\\
  Compresso (2023)~\cite{guo2023compresso} & S & N & N & N & LLaMA-7B & 28.57 & - & - & 73.55 & 73.07 & 49.16 & 64.80 & 27.68\\
  Compresso (2023)~\cite{guo2023compresso} & S & N & N & N & LLaMA-7B & 35.71 & - & - & 68.69 & 72.85 & 47.18 & 63.38 & 25.92\\
  LoRAShear (2023)~\cite{chen2023lorashear} & S & N & N & Y & LLaMA-7B & 20.00 & - & - & 72.78 & 76.36 & 69.49 & 67.63 & -\\ 
  LoRAShear (2023)~\cite{chen2023lorashear} & S & N & N & Y & LLaMA-7B & 50.00 & - & - & 63.40 & 72.15 & 49.83 & 56.40 & -\\ 
  Bonsai (2024)~\cite{dery2024everybody} & S & N & N & Y & LLaMA-7B & 50.00 & $10.92^{\dagger}$ & - & 67.22 & - & 43.09 & 61.64 & 28.92\\
  Shortened (2024)~\cite{kim2024shortened} & S & Y & N & Y & LLaMA-7B & 20.00 & 17.70 & 30.70 & 72.70 & 75.70 & 70.04 & 63.60 & -\\
  FLAP (2024)~\cite{an2024fluctuationbased} & S & N & N & N & LLaMA-7B & 20.00 & 14.62 & - & 69.63 & 76.82 & 71.20 & 68.35 & -\\
  SliceGPT (2024)~\cite{ashkboos2024slicegpt} \cite{song2024sleb} & S & Y & N & N & LLaMA-7B & 25.00 & - & - & - & 66.87 & 54.16 & 63.38 & -\\ 
  SliceGPT (2024)~\cite{ashkboos2024slicegpt} \cite{song2024sleb} & S & Y & N & N & LLaMA-7B & 30.00 & - & - & - & 63.55 & 49.62 & 61.33 & -\\  
  SLEB (2024)~\cite{song2024sleb} & S & N & N & N & LLaMA-7B & 10.00 & - & - & - & 76.44 & 70.23 & 63.14 & -\\
  SLEB (2024)~\cite{song2024sleb} & S & N & N & N & LLaMA-7B & 20.00 & $12.94^{\dagger}$ & - & - & 73.07 & 62.47 & 58.96 & -\\ 
  \hline
  SparseGPT (2023) \cite{frantar2023sparsegpt} \cite{sun2024simple} & U & Y & Y & N & LLaMA-2-7B & 50.00 & $6.52^{\dagger}$ & - & 75.02 & - & 52.37 & 69.85 & 38.68\\
  Wanda (2024)~\cite{sun2024simple} & U & Y & Y & N & LLaMA-2-7B & 50.00 & $6.44^{\dagger}$ & - & 75.99 & - & 52.49 & 68.19 & 39.27\\
  BESA (2024)~\cite{xu2024besa} & U & Y & Y & N & LLaMA-2-7B & 50.00 & $6.60^{\dagger}$ & $44.09^{\dagger}$ & 74.83 & 76.66 & 54.60 & 68.59 & -\\ 
  LLM-Pruner (2023)~\cite{ma2023llmpruner} \cite{men2024shortgpt} & S & Y & Y & Y & LLaMA-2-7B & 27.00 & - & - &  55.20 & 71.22 & 56.46 & - & 23.33 \\ 
  SliceGPT-Alpaca (2024)~\cite{ashkboos2024slicegpt} & S & Y & N & N & LLaMA-2-7B & 20.00 & - & - & - & 76.50 & 65.20 & 65.51 & -\\
  SliceGPT-Alpaca (2024)~\cite{ashkboos2024slicegpt} & S & Y & N & N & LLaMA-2-7B & 30.00 & - & - & - & 72.25 & 55.86 & 59.83 & -\\
  SliceGPT-Wiki (2024)~\cite{ashkboos2024slicegpt} & S & Y & N & N & LLaMA-2-7B & 20.00 & $6.64^{\dagger}$ & - & - & 69.42 & 59.04 & 65.11 & -\\ 
  SliceGPT-Wiki (2024)~\cite{ashkboos2024slicegpt} & S & Y & N & N & LLaMA-2-7B & 30.00 & $8.12^{\dagger}$ & - & - & 63.55 & 49.62 & 61.33 & -\\ 
  LLM Surgeon (2024)~\cite{ouderaa2024llm} & S & N & N & N & LLaMA-2-7B & 50.00 & 43.68  & - & 39.60 & 64.36 & 40.29 & 52.57 & -\\
  K-OBD (2024)~\cite{ouderaa2024llm} & S & N & N & N & LLaMA-2-7B & 50.00 & 136.33  & - & 61.56 & 60.66 & 36.84 & 53.04 & -\\
  Sheared (2024)~\cite{xia2024sheared} & S & N & N & Y & LLaMA-2-7B & 61.00 & -  & - & 73.70 & 75.80 & 70.80 & 64.20 & 26.40 \\
  Sheared (2024)~\cite{xia2024sheared} & S & N & N & Y & LLaMA-2-7B & 81.00 & -  & - & 64.00 & 73.40 & 60.70 & 57.90 & 25.70\\
  LaCo (2024)~\cite{yang2024laco} & S & Y & N & N & LLaMA-2-7B & 27.10 & - & - & 64.07 & 69.80 & 55.69 & - & 26.45\\
  ShortGPT (2024)~\cite{men2024shortgpt} & S & Y & N & N & LLaMA-2-7B & 27.10 & - & - & 74.71 & 66.43 & 53.02 & - & 43.96\\
  LLM-Streamline (2024)~\cite{chen2024compressing} & S & Y & N & Y & LLaMA-2-7B & 26.20 & - & - & 65.00 & 70.10 & 59.20 & - & 47.00\\ 
  \hline
  SparseGPT (2023)~\cite{frantar2023sparsegpt} \cite{bai2024sparsellm} & U & Y & Y & N & LLaMA-2-13B & 70.00 & $12.98^{\dagger}$ & $267.63^{\dagger}$ & 70.03 & - & 42.20 & 66.54 & -\\
  Wanda (2024)~\cite{sun2024simple} \cite{bai2024sparsellm} & U & Y & Y & N & LLaMA-2-13B & 50.00 & $23.42^{\dagger}$ & $502.53^{\dagger}$ & - & - & - & - & -\\
  SparseLLM (2024)~\cite{bai2024sparsellm} & U & Y & N & Y & LLaMA-2-13B & 70.00 & $12.95^{\dagger}$ & $277.76^{\dagger}$ & 69.87 & - & 42.50 & 68.64 & -\\
  LLM-Pruner (2023)~\cite{ma2023llmpruner} \cite{men2024shortgpt} & S & Y & Y & Y & LLaMA-2-13B & 24.40 & - & - &  56.42 & 76.66 & 67.76 & - & 25.21 \\ 
  SliceGPT-Alpaca (2024)~\cite{ashkboos2024slicegpt} & S & Y & N & N & LLaMA-2-13B & 20.00 &  & - & - & 77.97 & 69.64 & 68.90 & -\\
  SliceGPT-Alpaca (2024)~\cite{ashkboos2024slicegpt} & S & Y & N & N & LLaMA-2-13B & 30.00 &  & - & - & 74.10 & 60.91 & 65.82 & -\\ 
  SliceGPT-Wiki (2024)~\cite{ashkboos2024slicegpt} & S & Y & N & N & LLaMA-2-13B & 20.00 & $5.81^{\dagger}$ & - & - & 71.87 & 63.04 & 69.38 & -\\ 
  SliceGPT-Wiki (2024)~\cite{ashkboos2024slicegpt} & S & Y & N & N & LLaMA-2-13B & 30.00 & $6.99^{\dagger}$ & - & - & 66.10 & 52.69 & 65.11 & -\\  
  LaCo (2024)~\cite{yang2024laco} & S & Y & N & N & LLaMA-2-13B & 24.60 & - & - & 63.98 & 74.27 & 64.39 & - & 45.93\\
  ShortGPT (2024)~\cite{men2024shortgpt} & S & Y & N & N & LLaMA-2-13B & 24.60 & - & - & 62.48 & 73.45 & 66.64 & - & 54.69\\ 
  \hline
  SparseGPT (2023)~\cite{frantar2023sparsegpt} & U & Y & Y & N & OPT-6.7B & 50.00 & $11.55^{\dagger}$ & $17.44^{\dagger}$ & - & - & - & - & -\\
  SliceGPT-Alpaca (2024)~\cite{ashkboos2024slicegpt} & S & Y & N & N & OPT-6.7B & 20.00 & - & - & - & 74.54 & 62.84 & 62.67 & -\\
  SliceGPT-Alpaca (2024)~\cite{ashkboos2024slicegpt} & S & Y & N & N & OPT-6.7B & 30.00 & - & - & - & 73.34 & 58.93 & 61.80 & -\\ 
  SliceGPT-Wiki (2024)~\cite{ashkboos2024slicegpt} & S & Y & N & N & OPT-6.7B & 20.00 & $11.48^{\dagger}$ & - & - & 72.74 & 61.04 & 61.09 & -\\ 
  SliceGPT-Wiki (2024)~\cite{ashkboos2024slicegpt} & S & Y & N & N & OPT-6.7B & 30.00 & $12.51^{\dagger}$ & - & - & 68.61 & 54.56 & 60.69 & -\\
  SLEB (2024)~\cite{song2024sleb} & S & N & N & N & OPT-6.7B & 10.00 & $11.22^{\dagger}$ & - & - & 76.61 & 66.36 & 64.72 & -\\
  SLEB (2024)~\cite{song2024sleb} & S & N & N & N & OPT-6.7B & 20.00 & $12.94^{\dagger}$ & - & - & 74.92 & 62.13 & 61.33 & -\\ 
  LLM-Streamline (2024)~\cite{chen2024compressing} & S & Y & N & Y & OPT-6.7B & 26.00 & - & - & 63.00 & 73.10 & 52.40 & - & 24.60\\  
  \hline
  SparseGPT (2023)~\cite{frantar2023sparsegpt}  \cite{bai2024sparsellm} & U & Y & Y & N & OPT-30B & 70.00 & $9.58^{\dagger}$ & $14.41^{\dagger}$ & 68.78 & - & 53.83 & 67.64 & -\\
  Wanda (2024)~\cite{sun2024simple} \cite{bai2024sparsellm} & U & Y & Y & N & OPT-30B & 70.00 & $7766.61^{\dagger}$ & $5547.45^{\dagger}$ & - & - & - & - & -\\
  SparseLLM (2024)~\cite{bai2024sparsellm} & U & Y & N & Y & OPT-30B & 70.00 & $9.56^{\dagger}$ & $14.40^{\dagger}$ & 69.11 & - & 53.97 & 68.43 & -\\
  SliceGPT-Alpaca (2024)~\cite{ashkboos2024slicegpt} & S & Y & N & N & OPT-30B & 20.00 & - & - & - & 78.35 & 70.64 & 66.61 & -\\ 
  SliceGPT-Alpaca (2024)~\cite{ashkboos2024slicegpt} & S & Y & N & N & OPT-30B & 30.00 & - & - & - & 76.93 & 68.66 & 64.96 & -\\ 
  SliceGPT-Wiki (2024)~\cite{ashkboos2024slicegpt} & S & Y & N & N & OPT-30B & 20.00 & $9.87^{\dagger}$ & - & - & 76.50 & 70.61 & 66.61 & -\\ 
  SliceGPT-Wiki (2024)~\cite{ashkboos2024slicegpt} & S & Y & N & N & OPT-30B & 30.00 & $10.27^{\dagger}$ & - & - & 74.97 & 68.15 & 65.04 & -\\ 
  SLEB (2024)~\cite{song2024sleb} & S & N & N & N & OPT-30B & 10.00 & $9.57^{\dagger}$ & - & - & 77.64 & 72.32 & 68.75 & -\\
  SLEB (2024)~\cite{song2024sleb} & S & N & N & N & OPT-30B & 20.00 & $10.73^{\dagger}$ & - & - & 76.93 & 70.62 & 67.40 & -\\ 
  \hline
  LLM-Pruner (2023)~\cite{ma2023llmpruner} & S & Y & Y & Y & Vicuna-7B & 20.00 & 19.69 & 78.25 &  63.33 & 76.17 & 65.13 & 60.22 & -\\
  LLM-Pruner (2023)~\cite{ma2023llmpruner} \cite{kim2024shortened} & S & Y & Y & Y & Vicuna-7B & 20.00 & 19.60 & 76.40 &  65.40 & 76.20 & 68.90 & 64.40 & -\\
  LLM-Pruner (2023)~\cite{ma2023llmpruner} \cite{kim2024shortened} & S & Y & Y & Y & Vicuna-7B & 35.00 & 27.60 & 102.00 &  52.00 & 72.40 & 61.60 & 59.90 & -\\
  Shortened (2024)~\cite{kim2024shortened} & S & Y & N & Y & Vicuna-7B & 20.00 & 18.80 & 67.90 & 71.70 & 74.40 & 67.60 & 63.60 & -\\
  Shortened (2024)~\cite{kim2024shortened} & S & Y & N & Y & Vicuna-7B & 35.00 & 26.60 & 89.40 & 65.20 & 70.40 & 56.50 & 56.60 & -\\
  FLAP (2024)~\cite{an2024fluctuationbased} \cite{kim2024shortened} & S & N & N & N & Vicuna-7B & 20.00 & 22.00 & 74.90 & 73.10 & 74.80 & 67.90 & 65.80 & -\\
  FLAP (2024)~\cite{an2024fluctuationbased} \cite{kim2024shortened} & S & N & N & N & Vicuna-7B & 35.00 & 26.60 & 89.40 & 65.20 & 70.40 & 56.50 & 56.50 & -\\ 
  \Xhline{0.3ex}      
\end{tabular}
}
\vspace{-0.3cm}
\end{table*}
                                     
\section{Pruning for Specific Applications}
\label{applications}
\begin{figure}[t]
\centering
  \subfloat[Pipelines \& applications]{
  \begin{minipage}{4.5cm}
      \includegraphics[width=4.5cm,height=3.5cm]{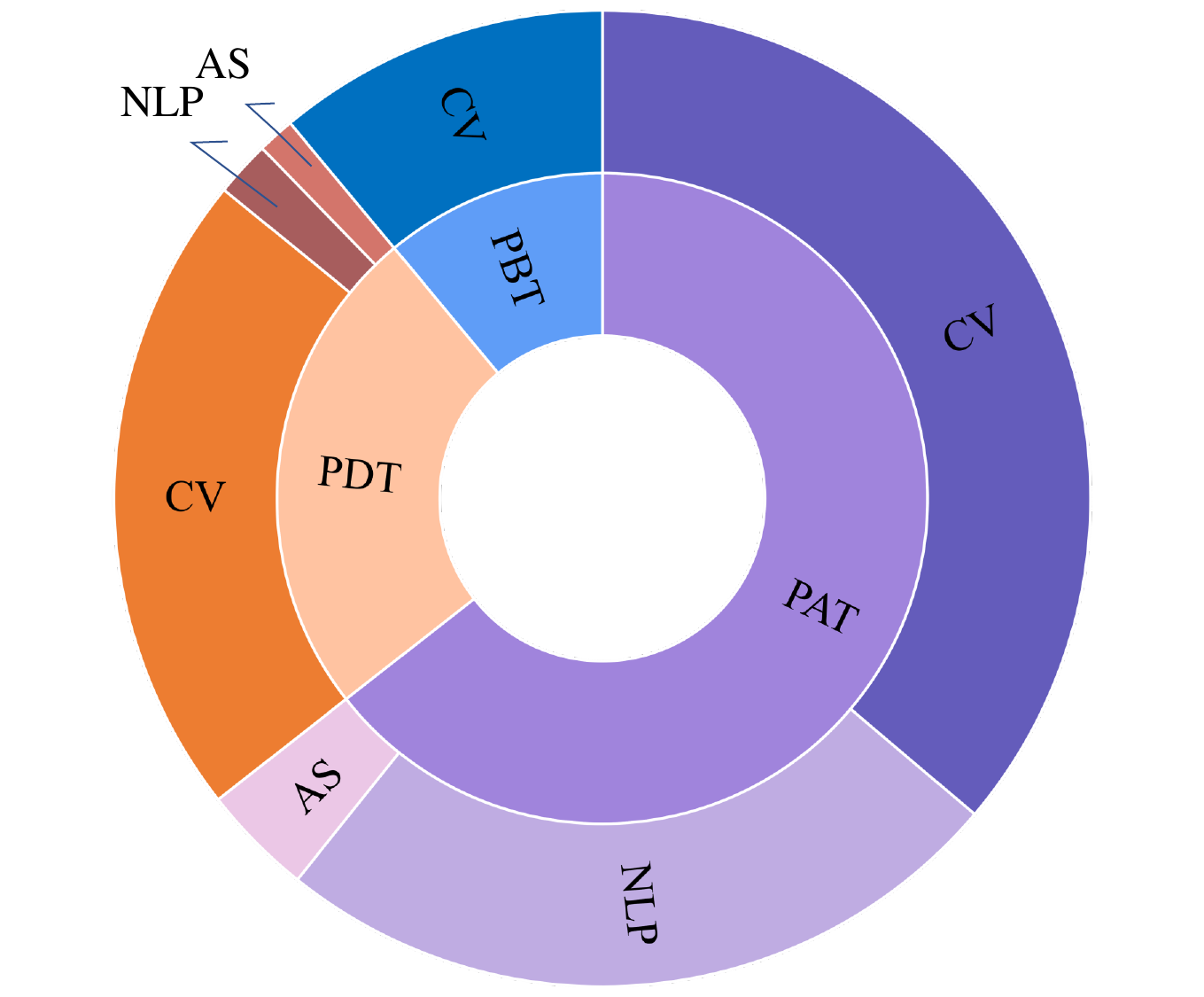}
        \label{Fig:statistics-a}
  \end{minipage}
  }
  \subfloat[Pipelines \& unstruct/struct]{  
  \begin{minipage}{4.5cm}
       \includegraphics[width=4.5cm,height=3.5cm]{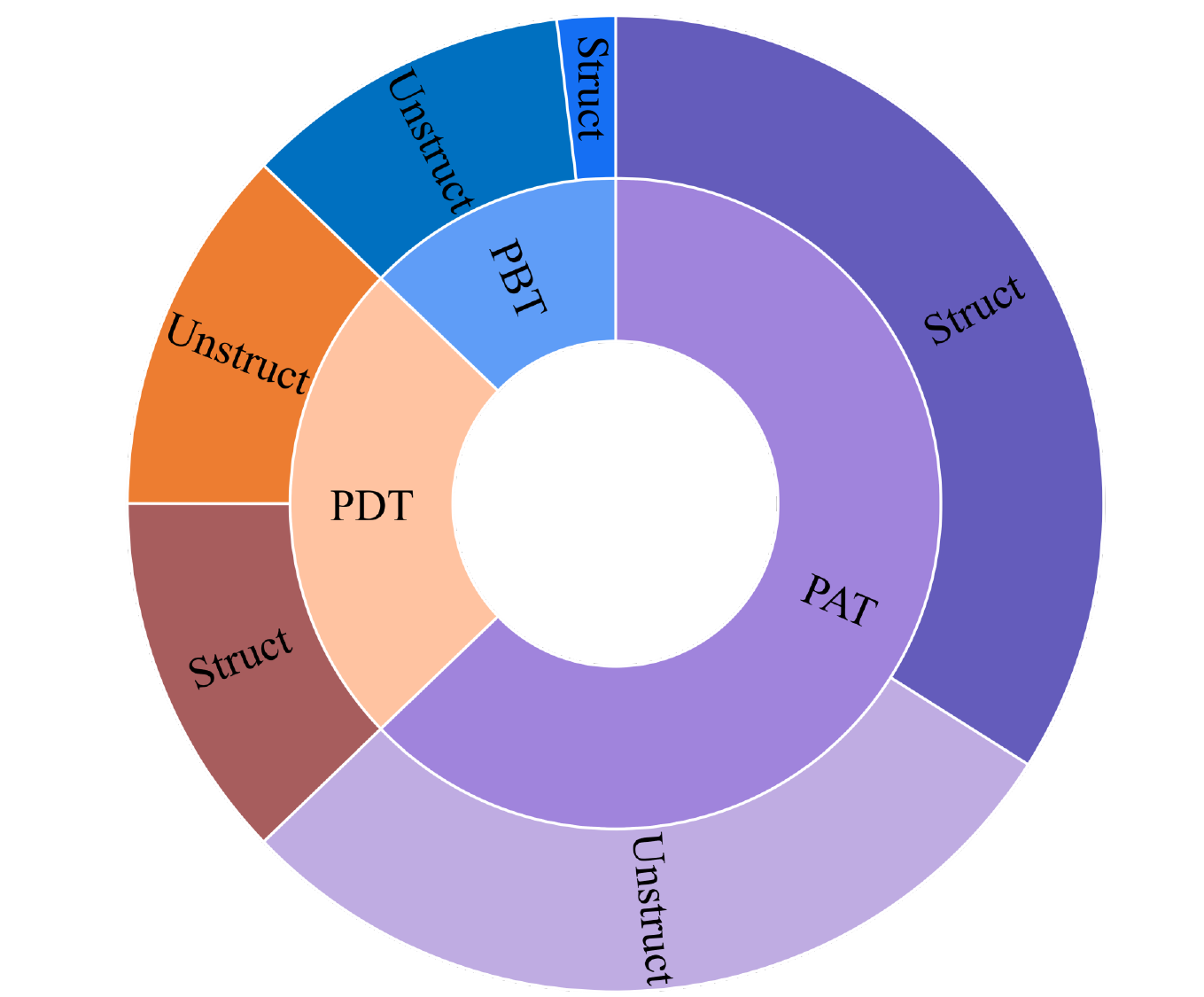}
     \label{Fig:statistics-b}
  \end{minipage} 
  }
 
 \caption{Statistics of pruning literature in our survey.}
 \label{Fig:statistics}
\vspace{-0.3cm}
\end{figure}

This section summarizes the characteristics of mainstream applications involved in pruning. The commonly used networks, datasets, and evaluation metrics of pruning are shown in Table \ref{Table:application-summary}. The statistics of pruning literature across different applications are shown in Fig.~\ref{Fig:statistics}.

\subsection{Image Classification}
\begin{figure}[t]
\centering
\begin{minipage}{8cm}
  \includegraphics[width=8cm,height=4cm]{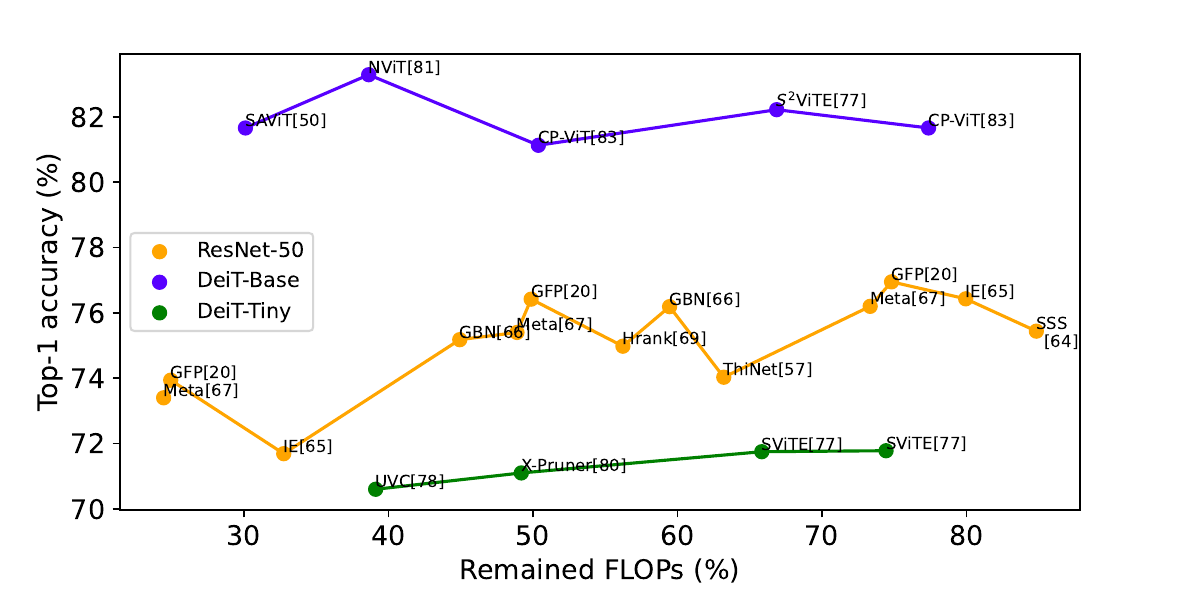}
\end{minipage}
\caption{Top-1 accuracy of representative pruning methods on ImageNet (best view in color and zoom in).}
\label{Fig:resnet-50-imagenet}
\vspace{-0.3cm}
\end{figure}

Image classification refers to classifying images according to their visual content and is the most basic task for pruning. Most pruning works in CV provide experimental results on ImageNet (ILSVRC-2012) \cite{russakovsky2015imagenet}. The representative pruning results on ImageNet are shown in Fig.~\ref{Fig:resnet-50-imagenet}, with results sourced from the corresponding papers. The networks used in papers with the same name may differ in the network structures, leading to different accuracy even if other settings are the same. For example, some works (e.g., \cite{nonnenmacher2022sosp,su2020sanity}) expand the layers' width of ResNet-32 by a factor (such as 2 or 4). Top-1 accuracy of the expanded ResNet-32 is higher than that of the vanilla ResNet-32. Some works (such as \cite{liu2017learning,huang2018data}) use VGG-16 with one FC layer, while others (\cite{lin2019towards,zhao2019variational}) adopt two or three FC layers. 

\subsection{Object Detection}
In comparison with image classification, object detection requires predicting the class and the exact location of each object in an image. Correspondingly, neural networks for object detection have more complex architectures, including backbones and other detection components. In addition, object detection generally requires larger input sizes, making pruning for object detection more challenging than image classification. Only a few works (e.g., \cite{yao2021DetNAS,bonnaerens2022anchor}) study pruning for object detection. For instance, \citet{bonnaerens2022anchor} propose an anchor pruning method. \citet{girish2021lottery} investigate LTH for object detection. \citet{liu2021group} prune one-stage and two-stage object detection models to validate performance. The commonly used accuracy metrics, as shown in Table~\ref{Table:application-summary}, include mAP (mean Average Precision) and COCO mAP that is mAP evaluated at Intersection-Over-Union (\textbf{IoU}) thresholds evenly distributed between 0.5 and 0.95.

\subsection{Image Style Translation}
Image style translation, which means transferring the style from one image to another, is an important application for deploying GANs on mobile devices. Compared with networks for image classification, GANs have very different network structures and outputs, consisting of a generator and a discriminator that output high-dimension images. \citet{shu2019coevolutionary} identify two major differences between compressing models for image classification and GANs. First, the discriminator network does not need to be compact because it will be discarded after training the generative network. Second, it is difficult to quantitatively evaluate the output images by generated GANs. Besides, the training difficulty poses extra challenges for pruning GANs \cite{shu2019coevolutionary}. Some pruning methods (e.g., \cite{shu2019coevolutionary,wang2020gan}) are proposed to reduce GAN's parameters and computational complexities. For example, \citet{chen2021gans} prune CycleGAN to verify matching subnetworks in GANs. 

As shown in Table \ref{Table:application-summary}, FCN-scores \cite{isola2017image2image}, and Fr\'echet Inception Distance (\textbf{FID}) between the feature distributions of real and
generated samples \cite{kalibhat2021winning} are often utilized to evaluate pruning results. FCN-scores include pixel accuracy, class accuracy, and class IoU. The larger FID indicates better transfer results. 

\subsection{Adversarial Robustness}
In safety-critical but computationally resource-constrained applications, neural network pruning faces the challenge of whether sparse models can preserve robustness. While the research community has extensively explored robust training and network pruning independently, only a few recent works (\cite{sehwag2019towards,sehwag2020hydra}) have studied them jointly. For example, \citet{sehwag2019towards} empirically verify that adversarial robustness can be achieved with weight magnitude-based pruning. \citet{sehwag2020hydra} increase the awareness of pruning techniques to the robust training objective and formulates the pruning objective as an empirical risk minimization problem. 

The benign accuracy in Table \ref{Table:application-summary} refers to the percentage of correctly classified original (i.e., non-modified) inputs. Empirical Robust Accuracy (\textbf{ERA}) refers to the percentage of robust test samples under gradient-based attacks. Verifiable Robust Accuracy (\textbf{VRA}) corresponds to the fraction of test samples that are verified to be robust by network verification methods as described in \cite{wong2018scaling}.

\subsection{Other CV Tasks}
In addition to the above tasks, some works explore pruning methods for other CV tasks, such as semantic segmentation (\cite{girish2021lottery}), image deraining (\cite{zou2022dreaming}), human-pose estimation (\cite{wang2018exploring}), head-pose estimation (\cite{aghli2021combining}), image super-resolution (\cite{li2020dhp}), object tracking (\cite{zhang2022pruning}), text-to-image generation (\cite{fang2023structural}), and backdoor attack (\cite{li2022baat}) etc. For example, \citet{wang2018exploring} prune CMU-pose model \cite{cao2017realtime} for human pose estimation. \citet{li2020dhp} prune DnCNN \cite{zhang2017beyond} and U-Net \cite{ronneberger2015unet} for image denoising. 

\subsection{Natural Language Processing}
\citet{see2016compression} explore one of the earliest pruning methods of deep neural networks for NLP, where a magnitude-based method is proposed to prune LSTMs \cite{hochreiter1997neural} for machine translation. \citet{yu2020playing} explore the lottery hypothesis in NLP. \citet{Voita2019analyzing} prune multi-head self-attention of Transformer \cite{vaswani2017attention}. \citet{chen2020lottery} pioneer the study of the lottery ticket hypothesis in pre-trained BERT \cite{devlin2019bert} models and find matching subnetworks. Currently, transformer-based large language models (such as LLaMA \cite{touvron2023llama}, OPT \cite{zhang2022opt}) have become the dominant modeling paradigm in natural language processing.

As shown in Table \ref{Table:application-summary} \footnote{WMT’14/16 refers to https://www.statmt.org/wmt14/translation-task.html}, the GLUE benchmark \footnote{The detailed information of GLUE tasks can refer to Table 1 in \cite{wang2019glue}.} provides nine tasks and their corresponding metrics, including accuracy, Pearson correlation, Matthew's correlation, etc. BiLingual Evaluation Understudy (\textbf{BLEU}) score \cite{papineni2002bleu} is often used for evaluating the accuracy of machine translation. Perplexity is used for language modeling or machine translation to measure the change in performance due to pruning. The lower the Perplexity, the better the pruned model.  

\subsection{Vision-and-Language Tasks}
Vision-and-Language (VL) is one of the most common areas of multimodal research. VL tasks include Visual Question Answering (VQA) \cite{goyal2017making}, image-text retrieval, etc. \citet{gan2022playing} pioneer the investigation of lottery ticket in VL tasks and find ``relaxed'' winning tickets that match 99\% of the full accuracy can be found with a 50\%-70\% prune ratio of parameters in UNITER \cite{chen2020uniter}. Table~\ref{Table:application-summary} lists widely employed models, datasets, and evaluation metrics for pruning on VQA and image-text retrieval. The evaluation metrics TR@1 and IR@1 for image retrieval tasks denote Top-1 text recall and image recall, respectively.

\subsection{Audio and Speech Processing}
Speech recognition is one of the most common tasks in audio and speech processing. Lightweight speech recognition has become an indispensable feature on mobile devices. \citet{nrang2017exploring} prune Deep Speech 2 \cite{amodei2016deep}, a recurrent neural network architecture, to validate their gradual pruning scheme. \citet{ding2022audio} extend the lottery ticket hypothesis to speech recognition models and investigate the existence of winning tickets. PARP \cite{lai2021parp} prunes pre-trained wav2vec 2.0 \cite{wav2vec2} and XLSR-53 \cite{xlsr53} for self-supervised speech recognition.

As shown in Table \ref{Table:application-summary}, the Word Error Rate (\textbf{WER}) is the standard metric for measuring the accuracy of speech recognition, which is defined as $\textrm{WER} = (S+I+D)/(S+I+C)$, where $S$, $I$, $D$, and $C$ denote the number of substitutions, insertions, deletions, and correct words, respectively. The lower the value, the better the accuracy of the speech recognition model. The Character Error Rate (\textbf{CER}) has the exact definition of WER, except that CER counts characters while WER counts words.

\begin{table*}[t]
  \caption{Summary of commonly used pruning evaluation metrics for various applications.}
  \centering
  \label{Table:application-summary}
    \scalebox{0.84}{%
  \begin{tabular}{l|c|cccc|c}
    \Xhline{0.3ex}   
    Task & Type & Dataset & Model & Performance & Efficiency & Example work\\
    \hline
    Image & \multirow{24}{*}{CV} & CIFAR-10/100 \cite{krizhevsky2009learning}, & ResNet-32/50/56 \cite{he2016deep}, & Top-1/Top-5 accuracy & FLOPs, & \cite{frankle2019lottery}, \cite{lin2020hrank},\\
    Classification & & ImageNet ILSVRC-2012 \cite{russakovsky2015imagenet} & VGG-16/19 \cite{simonyan2015very}, &  & MACs, & \cite{nonnenmacher2022sosp}, \cite{liu2021group},\\
    &   &  & ViT \cite{dosovitskiy2021image}/DeiT \cite{touvron2021training} /Swin \cite{liu2021swin} &  & $\lVert W \rVert_{0}$ & \cite{evci2022gradeint}, \cite{tanaka2020pruning}\\
    & & & & & &\\
    Object & & COCO \cite{lin2014coco}, & RetinaNet \cite{lin2017focal}, & mAP, & FLOPs, & \cite{liu2021group}, \cite{bonnaerens2022anchor}, \\
    Detection & & PASCAL VOC 2007/2010 \cite{everingham2008pascal}, & Faster R-CNN \cite{fasterrcnn}, & COCO mAP & $\lVert W \rVert_{0}$ & \cite{girish2021lottery} \\
    &  &  & SSD \cite{liu2016ssd} & & \\
    & & & & & & \\
    Generative tasks & & CIFAR-10 \cite{krizhevsky2009learning} & DDPMs \cite{ho2020denoising} & FID \cite{heusel2017gans} & MACs & \cite{fang2023structural} \\
    &  & CelebA-HQ (64$\times$64) \cite{liu2015deep} & LDMs \cite{rombach2022high} & SSIM \cite{wang2004image} & & \\
    & & LSUN Church (256$\times$256)) \cite{yu2016lsun} & & & & \\ 
    & & & & & &\\
    Image Style &  & horse2zebra \cite{zhu2017unpaired}, & CycleGAN \cite{zhu2017unpaired} &  FCN-scores \cite{isola2017image2image}, & FLOPs, & \cite{shu2019coevolutionary}, \cite{wang2020gan} \\
    Translation &  & summer2winter \cite{zhu2017unpaired}, & & FID \cite{heusel2017gans} & $\lVert W \rVert_{0}$ & \\
    &  & Cityscapes \cite{cityscape} & & &  & \\
    & & & & & &\\
    Adversarial & & CIFAR-10 \cite{krizhevsky2009learning}, & VGG-16 \cite{simonyan2015very}, & Top-1/Top-5 accuracy, & $\lVert W \rVert_{0}$ & \cite{sehwag2020hydra,sehwag2019towards}\\
    Robustness &  & ImageNet ILSVRC-2012 \cite{russakovsky2015imagenet} & ResNet-18/50 \cite{he2016deep}, & Benign accuracy, &  &  \\  
    &  &  & Wide-ResNet-28-5 & ERA/VRA &  & \\
    & & & & & \\
    Image Denoising & & U-Net \cite{ronneberger2015unet} & SIDD \cite{abdelhamed2018a} & PSNR and SSIM \cite{wang2004iamge} & $\lVert W \rVert_{0}$ & \cite{shi2023memory} \\
    & & & & & \\
    Human-pose &  & COCO \cite{lin2014coco} & CMU-pose \cite{cao2017realtime} & mAP & & \cite{wang2018exploring} \\
    Estimation & & & & & \\
    \hline
    Machine & \multirow{9}{*}{NLP} & WMT’14/16, & Transformer-based \cite{vaswani2017attention}, & BLEU \cite{papineni2002bleu}, & FLOPs, & \cite{yu2020playing}, \cite{see2016compression}, \\
    Translation &  & OpenSubtitles2018 \cite{lison2018opensubtitles} & LSTMs \cite{hochreiter1997neural} & & $\lVert W \rVert_{0}$ & \cite{zhu2017to}, \cite{Voita2019analyzing}\\
    & & & & & & \\
    Natural Language &  & SST-2 \cite{socher2013recursive}, QNLI \cite{rajpurkar2016squad} & BERTs \cite{devlin2019bert}, & Accuracy, & $\lVert W \rVert_{0}$ & \cite{chen2020lottery}, \cite{sanh2020movement}, \\
    Understanding &  & SQuAD \cite{rajpurkar2016squad}, QQP \cite{shankar2017first}, & LLaMA~\cite{touvron2023llama} & Pearson/Matthew cor., &  & \cite{ma2023llmpruner,zhang2023loraprune} \\
    &  &  MNLI \cite{williams2018broad} & OPT \cite{zhang2022opt} & F1 & & \cite{frantar2023sparsegpt,ashkboos2024slicegpt} \\
    & & & & &  & \\ 
    Language &  & Wikitext-2 \cite{merity2017pointer}, & LSTMs \cite{hochreiter1997neural}, & Perplexity & $\lVert W \rVert_{0}$ & \cite{yu2020playing}, \cite{zhu2017to} \\
    Modeling &  & Penn Tree Bank \cite{marcus1993building} & LLaMA~\cite{touvron2023llama} & & & \cite{ma2023llmpruner},\cite{xia2024sheared} \\
    \hline
    \multirow{3}{*}{Speech Recognition} & \multirow{3}{*}{ASP} & TED-LIUM \cite{rousseau2012ted}, & wav2vec 2.0 \cite{wav2vec2}, & WER, & $\lVert W \rVert_{0}$ & \cite{nrang2017exploring}, \cite{ding2022audio}, \\
     &  & Common Voice \cite{ardila2020common}, & LSTMs \cite{hochreiter1997neural}, & CER & & \cite{lai2021parp,jiang2023accurate} \\
    & & LibriSpeech \cite{panayotov2015librispeech} & Conformer \cite{gulati2020conformer} &  & & \\
    \hline
    \multirow{2}{*}{Image-Text Retrieval} & \multirow{4}{*}{VL}& COCO \cite{lin2014coco} & CLIP \cite{radford2021learning}, BLIP \cite{li2022blip} & TR@1, IR@1 & FLOPs & \cite{shi2023upop,sung2024ecoflap} \\
    & & Flickr30K \cite{young2014from} & & & $\lVert W \rVert_{0}$ & \\
    & & & & & & \\
    Visual Question Answering &  & VQAv2 \cite{goyal2017making} & CLIP \cite{radford2021learning}, BLIP \cite{li2022blip} & Accuracy & FLOPs & \cite{shi2023upop,sung2024ecoflap} \\
    \Xhline{0.3ex}   
\end{tabular}
}
\end{table*}



\ifCLASSOPTIONcaptionsoff
  \newpage
\fi



{\footnotesize
\bibliographystyle{IEEEtranN}
\bibliography{ref_survey}
}

\end{document}